\documentclass[sigconf, nonacm]{acmart}

\usepackage{amsmath,amsfonts,amsthm}
\usepackage{graphicx}
\usepackage{textcomp}
\usepackage{xcolor}
\usepackage{booktabs}       
\usepackage{nicefrac}       
\usepackage{microtype}      
\usepackage{xspace}
\usepackage{listings}
\usepackage{algorithm}
\usepackage[noend]{algpseudocode}
\usepackage{varwidth}
\usepackage{balance}
\usepackage{multirow}
\usepackage{bm}
\usepackage{dcolumn}
\usepackage{url}
\usepackage{cleveref}
\usepackage[tableposition=top]{caption}
\usepackage{tabularx}
\usepackage{lipsum}   
\usepackage{subcaption}  
\usepackage{hyperref}

\newcolumntype{L}{>{\raggedright\arraybackslash}X}

\newcommand{\method}{\textsc{Uldp-FL}\xspace}

\newtheorem{definition}{Definition}
\newtheorem{theorem}{Theorem}
\newtheorem{lemma}{Lemma}
\newtheorem{proposition}{Proposition}
\newtheorem{assumption}{Assumption}
\newtheorem{remark}{Remark}

\newcounter{protocol}
\newenvironment{protocol}[2]
  {\refstepcounter{protocol}\label{#1}
   \par\addvspace{\topsep}
   \noindent
   \tabularx{\linewidth}{@{} X @{}}
    \hline
    \textbf{Protocol \theprotocol} #2 \\
    \hline}
  { \\
    \hline
   \endtabularx
   \par\addvspace{\topsep}}

\newcommand\vldbdoi{}
\newcommand\vldbpages{}
\newcommand\vldbvolume{}
\newcommand\vldbissue{}
\newcommand\vldbyear{}
\newcommand\vldbauthors{\authors}
\newcommand\vldbtitle{\shorttitle} 
\newcommand\vldbavailabilityurl{https://github.com/FumiyukiKato/uldp-fl}
\newcommand\vldbpagestyle{empty} 

\begin{document}
\title{\method: Federated Learning with Across-Silo User-Level Differential Privacy}

\author{Fumiyuki Kato}
\affiliation{%
  \institution{Kyoto University}
  \country{}
}
\email{fumiyuki@db.soc.i.kyoto-u.ac.jp}

\author{Li Xiong}
\affiliation{%
  \institution{Emory University}
  \country{}
}
\email{lxiong@emory.edu}

\author{Shun Takagi}
\affiliation{%
  \institution{Kyoto University}
  \country{}
}
\email{takagi.shun.45a@st.kyoto-u.ac.jp}

\author{Yang Cao}
\affiliation{%
  \institution{Hokkaido University}
  \country{}
}
\email{yang@ist.hokudai.ac.jp}

\author{Masatoshi Yoshikawa}
\affiliation{%
  \institution{Osaka Seikei University}
  \country{}
}
\email{yoshikawa-mas@osaka-seikei.ac.jp}

\begin{abstract}
Differentially Private Federated Learning (DP-FL) has garnered attention as a collaborative machine learning approach that ensures formal privacy. 
Most DP-FL approaches ensure DP at the record-level within each silo for cross-silo FL.  
However, a single user's data may extend across multiple silos, and the desired user-level DP guarantee for such a setting remains unknown.
In this study, we present \method, a novel FL framework designed to guarantee user-level DP in cross-silo FL where a single user's data may belong to multiple silos.
Our proposed algorithm directly ensures user-level DP through per-user weighted clipping, departing from group-privacy approaches.
We provide a theoretical analysis of the algorithm's privacy and utility.
Additionally, we enhance the utility of the proposed algorithm with an enhanced weighting strategy based on user record distribution and design a novel private protocol that ensures no additional information is revealed to the silos and the server.
Experiments on real-world datasets show substantial improvements in our methods in privacy-utility trade-offs under user-level DP compared to baseline methods.
To the best of our knowledge, our work is the first FL framework that effectively provides user-level DP in the general cross-silo FL setting.
\end{abstract}

\maketitle

\pagestyle{\vldbpagestyle}
\begingroup\small\noindent\raggedright\textbf{PVLDB Reference Format:}\\
\vldbauthors. \vldbtitle. PVLDB, \vldbvolume(\vldbissue): \vldbpages, \vldbyear.\\
\href{https://doi.org/\vldbdoi}{doi:\vldbdoi}
\endgroup
\begingroup
\renewcommand\thefootnote{}\footnote{\noindent
This work is licensed under the Creative Commons BY-NC-ND 4.0 International License. Visit \url{https://creativecommons.org/licenses/by-nc-nd/4.0/} to view a copy of this license. For any use beyond those covered by this license, obtain permission by emailing \href{mailto:info@vldb.org}{info@vldb.org}. Copyright is held by the owner/author(s). Publication rights licensed to the VLDB Endowment. \\
\raggedright Proceedings of the VLDB Endowment, Vol. \vldbvolume, No. \vldbissue\ %
ISSN 2150-8097. \\
\href{https://doi.org/\vldbdoi}{doi:\vldbdoi} \\
}\addtocounter{footnote}{-1}\endgroup

\ifdefempty{\vldbavailabilityurl}{}{
\vspace{.3cm}
\begingroup\small\noindent\raggedright\textbf{PVLDB Artifact Availability:}\\
The source code, data, and/or other artifacts have been made available at \url{\vldbavailabilityurl}.
\endgroup
}

\section{Introduction}
\label{sec:intro}
Federated Learning (FL) \cite{mcmahan2016federated} is a collaborative machine learning (ML) scheme in which multiple parties train a single global model without sharing training data.
FL has attracted industry attention \cite{paulik2021federated, ramaswamy2019federated} as concerns about the privacy of training data have become more serious, as exemplified by GDPR \cite{gdpr}.
It should be noted that FL itself does not provide rigorous privacy protection for the trained model \cite{nasr2019comprehensive, zhao2020idlg}.  \textit{Differentially Private FL} (DP-FL) \cite{geyer2017differentially, kairouz2021distributed} further guarantees a formal privacy for trained models based on differential privacy (DP) \cite{dwork2006differential}.

\begin{figure}[t]
    \centering
    \includegraphics[width=1.0\hsize]{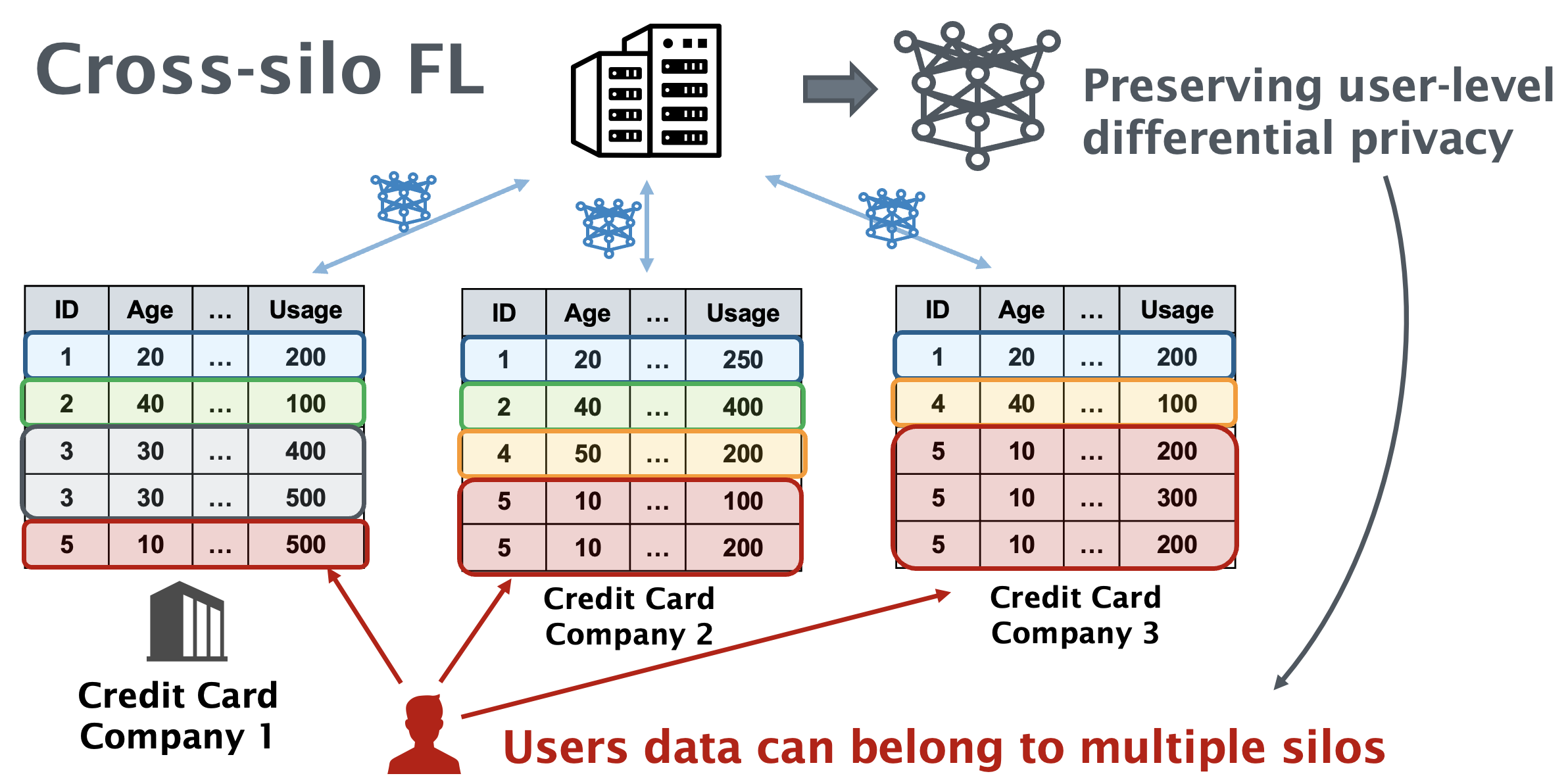}
    \caption{In cross-silo FL,  records belonging to the same user can exist across silos, e.g., a user can use several credit card companies. In this study, we investigate how to train models satisfying \textit{user-level} DP in this setting.}
    \label{fig:problem}
\end{figure}

Although DP is the de facto standard in the field of statistical privacy protection, it has a theoretical limitation.
The standard DP definition takes a single record as a unit of privacy.
This can easily break down in realistic settings where one user may provide multiple records, theoretically deteriorating the privacy loss bound of DP.
Furthermore, the necessity of this can be intuitively understood from the following example. 
User-level DP effectively protects privacy by restricting the influence of multiple records from a single user (e.g., multiple transactions in credit card history or one patient is diagnosed with the same disease at multiple hospitals), preventing the disclosure of his distinctive patterns, i.e., features and/or trends. 
This contrasts with record-level DP, which falls short in addressing the cumulative privacy risks associated with aggregating these records.
To address this, the notion of \textit{user-level DP} has been studied \cite{liu2020learning, levy2021learning, wilson2020differentially, amin2019bounding}.
In user-level DP, all records belonging to a single user are considered as a unit of privacy, which is a stricter definition than standard DP.
We distinguish user-level DP from \textit{group-privacy} \cite{dwork2014algorithmic}, which considers any $k$ records as privacy units.
User-level DP has also been studied in the FL context \cite{erlingsson2020encode, geyer2017differentially, mcmahan2018general, mcmahan2017learning, kairouz2021distributed}.
However, these studies focus on the cross-device FL setting, where one user's data belongs to a single device only.

Cross-silo FL \cite{ogier2022flamby, liu2022privacy, lowy2021private, lowy2023private} is a practical variant of FL in which a relatively small number of silos (e.g., hospitals or credit card companies) participate in training rounds.
In cross-silo FL, unlike in cross-device FL, a single user can have multiple records across silos, as shown in Figure \ref{fig:problem}.
Existing cross-silo DP-FL studies \cite{liu2022privacy, lowy2021private, lowy2023private} have focused on record-level DP for each silo; user-level DP across silos has not been studied.
Therefore, an important research question arises: \textit{How do we design an FL framework that guarantees user-level DP across silos in cross-silo FL?}

A baseline solution for guaranteeing user-level DP is to bound user contributions (number of records per user) as in \cite{wilson2020differentially, amin2019bounding} and then utilize group-privacy property of DP \cite{dwork2014algorithmic}.
Group-privacy simply extends the indistinguishability of the record-level DP to multiple records. 
We can convert any DP algorithm to group-privacy version of DP (see Lemma \ref{lemma:normaldp_group_privacy}, \ref{lemma:rdp_group_privacy} later), which we formally define as Group DP (GDP).
However, this approach can be impractical due to the super-linear privacy bound degradation of conversion to GDP and the need to appropriately limit the maximum number of user records (group size) in a distributed environment.
In particular, the former issue is a fundamental limitation for DP and highlights the need to develop algorithms that directly satisfy user-level DP without requiring conversion to GDP.

In this study, we present \method, a novel cross-silo FL framework, designed to directly ensure user-level DP through per-user weighted clipping and an effective weighting strategy. 
Additionally, we propose a novel private weighted aggregation protocol for implementing the weighting algorithm. 
This protocol ensures the private information of each silo is protected from both the server and other silos. 
The limitation of previous methods that relied solely on the Paillier cryptosystem for private weighted summation \cite{9476969} is that the raw data is visible to the party with the secret key.
We overcome this limitation by using a combination of several cryptographic techniques such as Paillier, Secure Aggregation \cite{bonawitz2017practical}, and Multiplicative blinding \cite{damgaard2007efficient}.
To the best of our knowledge, our work is the first FL framework that effectively provides user-level DP across silos in the general cross-silo FL setting (as in Figure \ref{fig:problem}).

The contributions of this work are summarized as follows:
\begin{itemize}
    
    
    \item We propose the \method framework and design baseline algorithms capable of achieving user-level DP across silos. The baseline algorithms limit the maximum number of records per user and use group-privacy property of DP.
    
    \item We then propose ULDP-AVG/SGD algorithms that directly satisfy user-level DP by implementing user-level weighted clipping within each silo, effectively bounding user-level sensitivity even when a single user may have unbounded number of records across silos. We provide theoretical analysis, including privacy bound and convergence analysis.

    \item We evaluate our proposed method and baseline approaches through comprehensive experiments on various real-world datasets. The results underscore that our proposed method yields superior trade-offs between privacy and utility compared to the baseline approaches.

    \item We further design an effective method by refining the weighting strategy for individual user-level clipping bounds. Since this approach may lead to additional privacy leaks, we develop a novel private weighted aggregation protocol employing several cryptographic techniques. We evaluate the extra computational overhead of the proposed private protocol using real-world benchmark scenarios.
\end{itemize}


\section{Background \& Preliminaries}
\label{sec:preliminaries}

\subsection{Cross-silo Federated learning}
\label{sec:preliminaries_crossfl}

In this work, we consider the following cross-silo FL scenario.
We have a central aggregation server and a set of silos $S$ participating in all rounds.
In each round, the server aggregates models from all silos and then redistributes the aggregated models.
Each silo $s \in S$ optimizes a local model $f_s$, which is the expectation of a loss function $F(x; \xi)$ that may be non-convex, where $x \in \mathbb{R}^d$ denotes the model parameters and $\xi$ denotes the data sample, and the expectation is taken over local data distribution $\mathcal{D}_{s}$.
In cross-silo FL, we optimize this global model parameter cooperatively across all silos.
Formally, the overarching goal in FL can be formulated as follows:
{\small
\begin{equation}
\label{eq:fl}
\begin{aligned}
\min_{x} \, \left\{ f(x) := \frac{1}{|S|}\sum_{s\in S}{f_s(x)} \right\},\, f_s(x) := \mathbb{E}_{\xi \sim \mathcal{D}_{s}}F(x; \xi).
\end{aligned}
\end{equation}
}

\noindent
Additionally, in our work, we have user set $U$ across all datasets across silos, where each record  belongs to one user $u \in U$, and each user may have multiple records in one silo and across multiple silos.
Each silo $s$ has local objectives for each user $u$, $f_{s, u} := \mathbb{E}_{\xi \sim \mathcal{D}_{s,u}}F(x; \xi)$, where $\mathcal{D}_{s,u}$ is the data distribution of $s$ and $u$.
In round $t \in [T]$ in FL, the global model parameter is denoted as $x_t$.

Note that this modeling is clearly different from cross-device FL in that there is no constraint that one user should belong to one device. 
Records from one user can belong to multiple silos.
For example, the same customer may use several credit card companies, etc. 
Additionally, all silos participate in all training rounds, unlike the probabilistic participation in cross-device FL \cite{mcmahan2017learning}, and the number of silos $|S|$ is small, around 2 to 100 \cite{kairouz2021advances}.

\subsection{Differential Privacy}
\label{sec:preliminaries_dp}

\noindent \textbf{Differential privacy.} DP \cite{dwork2006differential} is a rigorous mathematical privacy definition that quantitatively evaluates the degree of privacy protection for outputs.

\begin{definition}[$(\epsilon, \delta)$-DP]
A randomized mechanism $\mathcal{M}:\mathcal{D}\rightarrow\mathcal{Z}$ satisfies $(\epsilon, \delta)$-DP if, for any two input databases $D, D' \in \mathcal{D}$ s.t. $D'$ differs from $D$ in at most one record and any subset of outputs $Z \subseteq \mathcal{Z}$, it holds that
\begin{equation}
  \begin{aligned}
    \Pr[\mathcal{M}(D)\in Z] \leq \exp(\epsilon) \Pr[\mathcal{M}(D')\in Z] + \delta.
  \end{aligned}
\label{eq:dp}
\end{equation}
\end{definition}
\noindent
We call databases $D$ and $D'$ as \textit{neighboring} databases.
The maximum difference of the output for any neighboring database is referred to as \textit{sensitivity}.
We label the original definition as \textit{record-level} DP because the neighboring databases differ in only one record.

\smallskip
\noindent
\textbf{R\'{e}nyi differential privacy.}
R\'{e}nyi DP (RDP) \cite{mironov2017renyi} is a variant of approximate DP based on R\'{e}nyi divergence.
RDP is preferable because it is easy to use for \textit{Gaussian mechanism} \cite{mironov2017renyi} and has a tighter bound than the standard composition theorems.
The following lemmas give the bounds of the RDP for a typical mechanism and are used to further convert it to an original DP bound.

\begin{definition}[$(\alpha, \rho)$-RDP \cite{mironov2017renyi}]
  \label{def:rdp}
  Given a real number $\alpha \in (1, \infty)$ and privacy parameter $\rho \ge 0$, a randomized mechanism $\mathcal{M}$ satisfies $(\alpha, \rho)$-RDP if for any two neighboring datasets $D, D' \in \mathcal{D}$ s.t. $D'$ differs from $D$ in at most one record, we have that $D_{\alpha}(\mathcal{M}(D)||\mathcal{M}(D')) \le \rho$ where $D_{\alpha}(\mathcal{M}(D)||\mathcal{M}(D'))$ is the R\'{e}nyi
  divergence between $\mathcal{M}(D)$ and $\mathcal{M}(D')$ and is given by
  {\small
  \begin{equation}
  \nonumber
    D_{\alpha}(\mathcal{M}(D)||\mathcal{M}(D')) := \frac{1}{\alpha -1}\log{\mathbb{E}\left[\left(\frac{\mathcal{M}(D)}{\mathcal{M}(D')}\right)^{\alpha}\right]} \le \rho,
  \end{equation}
  }
  where the expectation is taken over the output of $\mathcal{M}(D)$.
\end{definition}

\begin{lemma}[RDP composition \cite{mironov2017renyi}]
  \label{lemma:rdp_composition}
  If $\mathcal{M}_{1}$ satisfies $(\alpha, \rho_1)$-RDP and $\mathcal{M}_{2}$ satisfies $(\alpha, \rho_2)$, then their
  composition $\mathcal{M}_{1} \circ \mathcal{M}_{2}$ satisfies $(\alpha, \rho_1 + \rho_2)$-RDP.
\end{lemma}
  
\begin{lemma}[RDP to DP conversion \cite{balle2020hypothesis}]
  \label{lemma:rdp_conversion}
  If $\mathcal{M}$ satisfies $(\alpha, \rho)$-RDP, then it also satisfies $(\rho', \delta)$-DP for any $0 < \delta < 1$ such that
  {\small
  \begin{equation}
    \nonumber
    \rho' = \rho + \log{\frac{\alpha-1}{\alpha}} - \frac{\log{\delta} + \log{\alpha}}{\alpha-1}.
  \end{equation}
  }
\end{lemma}

\begin{lemma}[RDP Gaussian mechanism \cite{mironov2017renyi}]
  \label{lemma:rdp_gaussian}
  If $f: D \rightarrow \mathbb{R}^d$ has $\ell2$-sensitivity $\Delta_{f}$, then the Gaussian mechanism $G_{f}(\cdot) := f(\cdot) + \mathcal{N}(0, I\sigma^2\Delta_{f}^2)$ is $(\alpha, \alpha / 2\sigma^2)$-RDP for any $\alpha > 1$.
\end{lemma}

\begin{lemma}[RDP for sub-sampled Gaussian mechanism \cite{wang2019subsampled}]
  \label{lemma:rdp_subsampled}
  Let $\alpha \in \mathbb{N} $ with $\alpha \ge 2$ and $0 < q < 1$ be a ratio of sub-sampling operation $Samp_q$.
  Let $G'_{f}(\cdot) := G_{f} \circ Samp_q (\cdot)$ be a sub-sampled Gaussian mechanism.
  Then, $G'_{f}$ is $(\alpha, \rho'(\alpha, \sigma))$-RDP where
  {\small
  \begin{equation}
  \begin{aligned}
  \nonumber
  \rho'(\alpha, \sigma) \le \frac{1}{\alpha - 1}\log\biggl(1 + 2q^2 \binom{\alpha}{2} &\min{\{2(e^{1/\sigma^2} - 1), e^{1/\sigma^2}\}} \\
  &+ \sum_{j=3}^{\alpha}{2q^j \binom{\alpha}{j} e^{j(j - 1)/2\sigma^2}} \biggl).
  \end{aligned}
  \end{equation}
  }
\end{lemma}
\noindent
In general, we can compute tighter numerical bounds along with the closed-form upper bounds described above \cite{wang2019subsampled, mironov2019r}.
In particular, RDP computation frameworks such as Opacus \cite{yousefpour2021opacus} use \cite{mironov2019r} analysis with Poisson sampling on records.

\smallskip
\noindent \textbf{Group differential privacy.} To extend privacy guarantees to multiple records, group-privacy \cite{dwork2014algorithmic} has been explored as a solution.
We refer to the group-privacy version of DP as Group DP (GDP).
\begin{definition}[$(k,\epsilon, \delta)$-GDP]
\label{def:group_dp}
A randomized mechanism $\mathcal{M}:\mathcal{D}\rightarrow\mathcal{Z}$ satisfies $(k,\epsilon, \delta)$-GDP if, for any two input databases $D, D' \in \mathcal{D}$, s.t. $D'$ differs from $D$ in at most $k$ records and any subset of outputs $Z \subseteq \mathcal{Z}$, Eq. (\ref{eq:dp}) holds.
\end{definition}
\noindent
GDP is a versatile privacy definition, as it can be applied to existing DP mechanisms without modification.

To convert DP to GDP, it is known that any $(\epsilon, 0)$-DP mechanism satisfies $(k,k\epsilon, 0)$-GDP \cite{dwork2014algorithmic}.
However, in the case of any $\delta > 0$, $\delta$ increases super-linearly \cite{gautam2020lec5}, leading to a much larger $\epsilon$.

\begin{lemma}[Group privacy conversion (record-level DP to GDP) \cite{gautam2020lec5}]
\label{lemma:normaldp_group_privacy}
If $f$ is $(\epsilon, \delta)$-DP, for any two input databases $D, D' \in \mathcal{D}$ s.t. $D'$ differs from $D$ in at most $k$ records and any subset of outputs $Z \subseteq \mathcal{Z}$, it holds that
\begin{equation}
    \nonumber
  \Pr[f(D)\in Z] \leq \exp(k\epsilon) \Pr[f(D')\in Z] + ke^{(k-1)\epsilon}\delta.
\end{equation}
\end{lemma}
\noindent 
It means when $f$ is $(\epsilon, \delta)$-DP, $f$ satisfies $(k, k\epsilon, k\exp^{(k-1)\epsilon}\delta)$-GDP.

Also, we can compute GDP using group-privacy property of R\'{e}nyi DP \cite{mironov2017renyi}.
First, we calculate the RDP of the algorithm, then convert it to group version of RDP, and subsequently to GDP.

\begin{lemma}[Group-privacy of RDP (record-level DP to GDP) \cite{mironov2017renyi}]
    \label{lemma:rdp_group_privacy}
    If $f: D \rightarrow \mathbb{R}^d$ is $(\alpha, \rho)$-RDP, $g: D' \rightarrow D$ is $k$-stable and $\alpha \ge 2^{k+1}$, then $f \circ g$ is $(\alpha /2^{k}, 3^{k} \rho)$-RDP.
\end{lemma}
\noindent
Here, group-privacy property is defined using a notion of $k$-stable transformation \cite{mcsherry2009privacy}.
$g: D' \rightarrow D$ is $k$-stable if $g(A)$ and $g(B)$ are neighboring in $D$ implies that there exists a sequence of length $c+1$ so that $D_0 = A,...,D_c=B$ and all $(D_i, D_{i+1})$ are neighboring in $D'$.
This privacy notion corresponds to $(k,\cdot,\cdot)$-GDP in Definition \ref{def:group_dp}.

\begin{figure}[t]
    \centering
    \includegraphics[width=0.9\hsize]{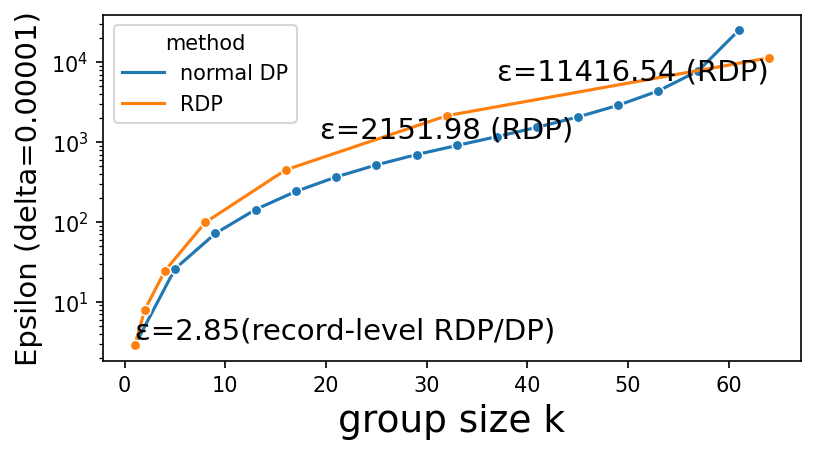}
    \caption{Group-privacy conversion results.}
    \label{fig:groupdp_comparison_sigma5}
\end{figure}

Here, to highlight the significant privacy degradation of GDP, we conduct a pre-experiment and show the converted privacy bounds for increasing group sizes.
Figure \ref{fig:groupdp_comparison_sigma5} illustrates a numerical comparison of the group-privacy conversion from DP to GDP with normal DP (Lemma \ref{lemma:normaldp_group_privacy}) and RDP (Lemma \ref{lemma:rdp_group_privacy}).
To compute the final GDP privacy bounds, we repeatedly run the Gaussian mechanism with $\sigma=5.0$ and a sampling rate of $0.01$ for $10^{5}$ iterations, emulating a typical DP-SGD \cite{abadi2016deep} (which is the common mechanism for DP ML) setup.
We use a fixed $\delta=10^{-5}$ and vary the group size $k=1, 2, 4, 8, 16, 32, 64$.
To compute GDP, RDP for sub-sampled Gaussian mechanisms is calculated according to \cite{wang2019subsampled}.
We then compute GDP using group-privacy of RDP by Lemma \ref{lemma:rdp_group_privacy}.
For normal DP, the computed RDP is converted to normal DP by Lemma \ref{lemma:rdp_conversion}, and then to GDP by Lemma \ref{lemma:normaldp_group_privacy}.

When converting from normal DP to GDP, computing the final $\epsilon$ at a fixed $\delta$ is challenging.
In Lemma \ref{lemma:rdp_conversion}, the output $\epsilon$ (denoted as $\epsilon_{l\ref{lemma:rdp_conversion}}$) depends on the input $\delta$ (denoted as $\delta_{l\ref{lemma:rdp_conversion}}$), and the final $\delta$ (denoted as $\delta_{l\ref{lemma:normaldp_group_privacy}}$) output by Lemma \ref{lemma:normaldp_group_privacy} depends on both $\epsilon_{l\ref{lemma:rdp_conversion}}$ and $\delta_{l\ref{lemma:rdp_conversion}}$.
Therefore, we repeatedly select $\delta_{l\ref{lemma:rdp_conversion}}$ in a binary search manner, compute $\epsilon_{l\ref{lemma:rdp_conversion}}$ and $\delta_{l\ref{lemma:rdp_conversion}}$, and finally report the $\epsilon$ when the difference between $\delta_{l\ref{lemma:rdp_conversion}}$ and $\delta=10^{-5}$ is sufficiently small (accuracy by $10^{-8}$) as the $\epsilon$ of GDP\footnote{The implementation is in the function \texttt{get\_normal\_group\_privacy\_spent()} in \url{https://github.com/FumiyukiKato/uldp-fl/blob/main/src/noise_utils.py}}.
Note that this method does not guarantee achieving the optimal $\epsilon$ for the given $\delta$, but it finds a reasonable $\epsilon$.

In the figure, we plot various group sizes, $k$, on the x-axis and $\epsilon$ of GDP at a fixed $\delta=10^{-5}$ on the y-axis.
Significantly, the results indicate that as the group size $k$ increases, $\epsilon$ grows rapidly, highlighting a considerable degradation in the privacy bound of GDP.
For instance, with $\epsilon=2.85$ at record-level ($k=1$), the value reaches 2100 for only $k=32$, and 11400 at $k=64$.
While there might be some looseness in the group-privacy conversion of RDP compared to normal DP for some small group sizes, the difference is relatively minor (roughly three times at most).
The drastic change in normal DP with large group size is due to the numerical instability in our conversion procedure\footnotemark[\value{footnote}].
RDP's conversion is easier to compute with a fixed $\delta$.
Hence, we utilize RDP's conversion in the experiments.


\subsection{Differentially Private FL}
DP has been applied to the FL paradigm, with the goal of ensuring that the trained model satisfies DP.
A popular DP variant in the context of cross-device FL is user-level DP (also known as client-level DP) \cite{geyer2017differentially, agarwal2021skellam, kairouz2021distributed}.
Informally, this definition ensures indistinguishability for device participation and has demonstrated a favorable privacy-utility trade-off if the number of clients are sufficient, even with large-scale models \cite{mcmahan2017learning}.
These studies often employ secure aggregation \cite{bonawitz2017practical, bell2020secure} to mitigate the need for trust in other parties during FL model training.
This is achieved by allowing the server and other silos to only access appropriately perturbed models after aggregation, often referred to as Distributed DP \cite{kairouz2021distributed, cheu2019distributed}.
In particular, \textit{shuffling}-based variants have recently gained attention \cite{cheu2019distributed, liew2022network,takagi2023bounded} and are being deployed in FL \cite{girgis2021shuffled}, which also provides user-level DP.
All of these studies assume that a single device holds all records for a single user, i.e., cross-device FL.
However, in a cross-silo setting, this definition does not extend meaningful privacy protection to individual users when they possess multiple records across silos.

Another DP definition in cross-silo FL offers record-level DP within each silo \cite{liu2022privacy, lowy2021private, lowy2023private}, referred to as \textit{Silo-specific sample-level} or \textit{Inter-silo record-level DP}.
These studies suggest that record-level DP can guarantee user-level DP through group-privacy.
However, they cannot account for settings where a single user can have records across multiple silos.
As far as we know, no method exists for training models that satisfy user-level DP in cross-silo FL where a single user's records can extend across multiple silos.
More fine-grained comparison between DP variants in FL can be seen in the Appendix.

\section{\method Framework}

\label{sec:problem_setting}

\subsection{Trust model and Assumptions}
We assume that all (two or more) silos and an aggregation server are \textit{semi-honest} (or \textit{honest-but-curious}).
This is a typical assumption in prior works \cite{bonawitz2017practical, so2023securing}.
In our study, aggregation is performed using secure aggregation to ensure that the server only gains access to the model after aggregation \cite{kairouz2021distributed}.
All communications between the server and silos are encrypted with SSL/TLS, and third parties with the ability to snoop on communications cannot access any information except for the final trained model.
We assume that there is no collusion, which is reasonable given that silos are socially separate institutions (such as different hospitals or companies). 
Additionally, in our scenario, we assume that record linkage \cite{vatsalan2017privacy} across silos has already been completed, resulting in shared common user IDs.
Both the server and the silos are aware of the total number of users $|U|$ and the number of silos $|S|$.
When user (device)-level sub-sampling is employed for DP amplification, only the server is permitted to know the sub-sampling results for each round \cite{kairouz2021distributed, agarwal2021skellam}.
Note that all these assumptions do not affect the privacy guarantees of the final model released to external users.

\subsection{Privacy definition}

In contrast to GDP, which offers indistinguishability for any $k$ records, user-level DP \cite{levy2021learning, geyer2017differentially} provides a more reasonable user-level indistinguishability regardless of the number of the records.
While \cite{geyer2017differentially} focuses solely on a cross-device FL context, we re-establish user-level DP (ULDP) in the cross-silo setting as follows:
\begin{definition}[$(\epsilon, \delta)$-User-Level DP (ULDP)]
\label{def:asuldp}
A randomized mechanism $\mathcal{M}:\mathcal{D}\rightarrow\mathcal{Z}$ satisfies $(\epsilon, \delta)$-ULDP if, for any two input databases across silos $D, D' \in \mathcal{D}$, s.t. $D'$ differs from $D$ in at most one user's records, and any $Z \subseteq \mathcal{Z}$, Eq. (\ref{eq:dp}) holds.
\end{definition}
\noindent
The fundamental difference of user-level DP from record-level DP lies in the definition of the neighboring databases.
The user-level neighboring database inherently defines \textit{user-level sensitivity}. 
Additionally, it is important to emphasize that the input database $D$ represents the comprehensive database spanning across silos.

If the number of records per user in the database is less than or equal to $k$, it is clear that GDP is a generalization of ULDP, and the following proposition holds.
\begin{proposition}
\label{prop:gdp_uldp}
    If a randomized mechanism $\mathcal{M}$ is $(k, \epsilon, \delta)$-GDP with input database $D$ in which any user has at most $k$ records, the mechanism $\mathcal{M}$ with input database $D$ also satisfies $(\epsilon, \delta)$-ULDP.
\end{proposition}
\noindent
One drawback of GDP is the challenge of determining the appropriate value for $k$.
Setting $k$ to the maximum number of records associated with any individual user could lead to introducing excessive noise to achieve the desired privacy protection level.
On the other hand, if a smaller $k$ is chosen, the data of users with more than $k$ records must be excluded from the dataset, potentially introducing bias and compromising model utility.
In this context, while several studies have analyzed the theoretical utility for a given $k$ \cite{liu2020learning, levy2021learning} and theoretical considerations for determining $k$ have been partially explored in \cite{amin2019bounding}, it still remains an open problem.
In contrast, ULDP does not necessitate the determination of $k$. 
Instead, it requires designing a specific ULDP algorithm.

\subsection{Baseline methods: ULDP-NAIVE/GROUP}
\label{sec:baseline_methods}

\begin{table}[h]
\caption{\textcolor{black}{Notation Table for Algorithms.}}
\label{table:notation}
\centering
\begin{tabularx}{\columnwidth}{cL} 
\hline
\textbf{\textcolor{black}{Symbol}} & \textbf{\textcolor{black}{Description}} \\
\hline
\hline
\textcolor{black}{$\eta_{l}$} & \textcolor{black}{Local learning rate.} \\
\textcolor{black}{$\eta_{g}$} & \textcolor{black}{Global learning rate.} \\
\textcolor{black}{$\sigma$} & \textcolor{black}{Noise parameter. (Noise multiplier.)} \\
\textcolor{black}{$C$} & \textcolor{black}{Clipping bound.} \\
\textcolor{black}{$T$} & \textcolor{black}{Total number of rounds.} \\
\textcolor{black}{$Q$} & \textcolor{black}{Number of local epochs.} \\
\textcolor{black}{$w_{s,u}$} & \textcolor{black}{Weight for user $u$ in silo $s$.} \\
\textcolor{black}{$\mathbf{W}$} & \textcolor{black}{Matrix of weights for users and silos. $\mathbf{W} = (\mathbf{w}_1,...,\mathbf{w}_{|S|})$: matrix with weight for user $u$ and silo $s$, and $\forall u \in U$, $w_{s,u} \in \mathbf{w}_s$ and $\sum_{s \in S}{w_{s,u}}=1$} \\
\textcolor{black}{$x_{0}$} & \textcolor{black}{Initial model.} \\
\textcolor{black}{$x_{t}$} & \textcolor{black}{Model at round $t$.} \\
\textcolor{black}{$\Delta^{s}_{t}$} & \textcolor{black}{Model update from silo $s$ at round $t$.} \\
\textcolor{black}{$g^{s,u}_{t,q}$} & \textcolor{black}{Stochastic gradient for user $u$ in silo $s$ at round $t$ and epoch $q$.} \\
\textcolor{black}{$\Delta^{s,u}_{t}$} & \textcolor{black}{Model update for user $u$ in silo $s$ at round $t$.} \\
\textcolor{black}{$\Tilde{\Delta}^{s,u}_{t}$} & \textcolor{black}{Clipped and weighted model update for user $u$ in silo $s$ at round $t$.} \\
\textcolor{black}{$g^{s,u}_{t}$} & \textcolor{black}{Stochastic gradient for user $u$ in silo $s$ at round $t$.} \\
\textcolor{black}{$\Tilde{g}^{s,u}_{t}$} & \textcolor{black}{Clipped and weighted gradient for user $u$ in silo $s$ at round $t$.} \\
\textcolor{black}{$\mathcal{N}(0, b)$} & \textcolor{black}{Gaussian noise vector with mean 0 and variance $b$.} \\
\textcolor{black}{$q$} & \textcolor{black}{User-level sub-sampling probability.} \\
\hline
\end{tabularx}
\end{table}

Table \ref{table:notation} summarizes symbols used in later algorithms in the paper.
Please note that all omitted proofs for the following theorems can be found in the Appendix.

\smallskip
\noindent\textbf{ULDP-NAIVE.} We begin by describing two baseline methods.
The first method is ULDP-NAIVE (described in Algorithm \ref{alg:naive}), a straightforward approach using substantial noise.
It works similarly to DP-FedAVG \cite{mcmahan2017learning}, where each silo locally optimizes with multiple epochs, computes the model update (delta), clips by $C$, and adds Gaussian noise. The original DP-FedAVG adds Gaussian noise  with variance $\sigma^2 C^2$.
In ULDP-NAIVE, since a single user may contribute to the model delta of all silos, the sensitivity across silos is $C|S|$ for the aggregated model delta, hence it needs to scale up the noise as $\sigma^2 C^2|S|$ (Line 14) such that the aggregated result from $|S|$ silos satisfies required DP.
Compared to DP-FedAVG, which focuses on cross-device FL, the number of model delta samples (number of silos as opposed to the number of devices) in our setting is very small, resulting in larger variance.
Thus, ULDP-NAIVE satisfies ULDP but at a significant sacrifice in utility.
The aggregation is performed using secure aggregation and is assumed to be so in the following algorithms.

\begin{theorem}
\label{theo:uldp_naive}
For any $0 < \delta < 1$ and $\alpha > 1$, given noise multiplier $\sigma$, ULDP-NAIVE satisfies $(\epsilon=\frac{T\alpha}{2\sigma^2} + \log{((\alpha-1)/\alpha)} - (\log{\delta} + \log{\alpha})/(\alpha-1), \delta)$-ULDP after $T$ rounds.
\textnormal{(The actual $\epsilon$ is numerically calculated by selecting the optimal $\alpha$ so that $\epsilon$ is minimized.)}
\end{theorem}

\begin{algorithm}[t]
    \caption{ULDP-NAIVE}
    \label{alg:naive}
    \begin{algorithmic}[1]
    \renewcommand{\algorithmicrequire}{\textbf{Input:}}
    \renewcommand{\algorithmicensure}{\textbf{Output:}}
    
    \Require $\eta_{l}$, $\eta_{g}$: local and global learning rates, $\sigma$: noise parameter, $C$: clipping bound, $T$: \#rounds, $Q$: \#local epochs
    
    \Procedure{Server}{}
        \State Initialize model $x_{0}$
        \For{each round $t=0, 1, \ldots, T-1$}
            \For{each silo $s \in S$}
                \State $\Delta^{s}_{t} \leftarrow$ \textsc{Client}$(x_{t}, C, \sigma, \eta_l)$
            \EndFor
        \State $x_{t+1} \leftarrow x_{t} + \eta_{g} \frac{1}{|S|}\sum_{s\in S}{\Delta^{s}_{t}}$
        \EndFor
    \EndProcedure
    
    \Procedure{Client}{$x_{t}, C, \sigma, \eta_l$}
        \State $x_{s} \leftarrow x_{t}$
        \For{epoch $q=0, 1, \ldots, Q-1$}
            \State Compute stochastic gradients $g^{(s)}_{t,q}$ \Comment{$\mathbb{E}[g^{(s)}_{t,q}] = \nabla f_s(x_{s})$}
            \State $x_{s} \leftarrow x_{s} - \eta_{l} g^{(s)}_{t,q}$
        \EndFor
        \State $\Delta_{t} \leftarrow x_{t} - x_{s}$
        \State $\Tilde{\Delta}_{t} \leftarrow \Delta_{t} \cdot \min{\left(1, \frac{C}{\lVert\Delta_{t}\rVert_{2}}\right)}$ \Comment{clipping with $C$}
        \State $\Delta'_{t} \leftarrow \Tilde{\Delta}_{t} + \mathcal{N}(0, I\sigma^2 C^2 |S|)$ \Comment{based on user-level sensitivity}
        \State \textbf{return} $\Delta'_{t}$
    \EndProcedure
    
    \end{algorithmic}
\end{algorithm}

\noindent\textbf{ULDP-GROUP-$k$.} 
We introduce a second baseline, ULDP-GROUP-$k$, utilizing group DP (described in Algorithm \ref{alg:baseline}), which limits each user's records to a given $k$ while satisfying $(k, \epsilon, \delta)$-GDP.
As Proposition \ref{prop:gdp_uldp} implies, this ensures $(\epsilon, \delta)$-ULDP.
The algorithm achieves GDP by firstly performing \textsc{DP-SGD} \cite{abadi2016deep} (Line 9) and converting from record-level DP within each silo.
The core principle of the algorithm is similar to that of \cite{liu2022privacy}. 
Before executing \textsc{DP-SGD}, it is essential to constrain the number of records per user to $k$ (Line 8).
This is accomplished by employing flags, denoted as $\mathbf{B}$, which indicate the records to be used for training (i.e., $b^{s}{u,i}=1$), with a total of $k$ records for each user across all silos (i.e., $\forall{u}, \sum_{s,i}{b^{s}_{u,i}} \le k$).
These flags must be consistent across all rounds.
We disregard the privacy concerns in generating these flags as this is a baseline method.

\begin{algorithm}[t]
\caption{ULDP-GROUP-${k}$}
\label{alg:baseline}
\begin{algorithmic}[1]
\renewcommand{\algorithmicrequire}{\textbf{Input:}}
\renewcommand{\algorithmicensure}{\textbf{Output:}}

\Require $\eta_{l}$, $\eta_{g}$: local and global learning rates, $\sigma$: noise parameter, $D_{s}$: training dataset of silo $s$, $C$: clipping bound, $T$: \#rounds, $Q$: \#local epochs, $k$: group size, $\gamma$: sampling rate, $\mathbf{B}$: flags for limit contribution s.t. for each matrix $\mathbf{b}^{s} \in \mathbf{B}$ if $b^{s}_{u, i} = 1$ the user $u$'s $i$-th record in silo $s$ is used, otherwise the record is excluded

\Procedure{Server}{}
    \State Initialize model $x_{0}$
    \For{each round $t=0, 1, \ldots, T-1$}
        \For{each silo $s \in S$}
            \State $\Delta^{s}_{t} \leftarrow$ \textsc{Client}$(x_{t}, C, \sigma, \eta_l, \gamma, \mathbf{b}^{s})$
        \EndFor
    \State $\theta_{t+1} \leftarrow \theta_{t} + \eta_{g} \frac{1}{|S|}\sum_{s\in S}{\Delta^{s}_{t}}$
    \EndFor
\EndProcedure

\Procedure{Client}{$x_{t}, C, \sigma, \eta_l, \gamma, \mathbf{b}^{s}$}
    \State $D'_{s} \leftarrow $ filter $D_{s}$ by $\mathbf{b}^{s}$
    \State $x^{Q}_t \leftarrow $\textsc{DP-SGD}$(\theta_{t}, D'_{s}, C, \sigma, \eta_l, \gamma, Q)$ \Comment{Algorithm 1 in \cite{abadi2016deep}}
    \State $\Delta_{t+1} \leftarrow x^{Q}_t - x_{t}$
    \State \textbf{return} $\Delta_{t+1}$
\EndProcedure

\end{algorithmic}
\end{algorithm}

\begin{theorem}
\label{theo:uldp_group}
If flags $\mathbf{B}$ is given privately, for any $0 < \delta < 1$, any integer $k$ to the power of 2 and $\alpha > 2^{k+1}$, ULDP-GROUP-${k}$ satisfies $(3^{k}\rho + \log{((\frac{\alpha}{2^k}-1)/\frac{\alpha}{2^k})} - (\log{\delta} + \log{\frac{\alpha}{2^k}})/(\frac{\alpha}{2^k}-1), \delta)$-ULDP where $\rho = \max_{s\in S} \rho_s$ s.t. for each silo $s\in S$, DP-SGD of local subroutine satisfies $(\alpha, \rho_s)$-RDP. 
\end{theorem}


While ULDP-GROUP shares algorithmic similarities with existing record-level DP cross-silo FL frameworks \cite{liu2022privacy}, it presents weaknesses from several perspectives:
(1) Significant degradation of privacy bounds due to the group-privacy conversion (DP to GDP).
(2) The challenge of determining an appropriate group size $k$ \cite{amin2019bounding}, which requires substantial insights into data distribution across silos and might breach the trust model. The determination of the flags $\mathbf{B}$ can also be problematic.
(3) The use of group-privacy to guarantee ULDP necessitates removing records from the training dataset, potentially introducing bias and causing utility degradation \cite{amin2019bounding, epasto2020smoothly}.
Our next proposed method aims to address these challenges.

\subsection{Advanced methods: ULDP-AVG/SGD}
\begin{algorithm}[t]
\caption{ULDP-AVG / ULDP-SGD}
\label{alg:uldpavg}
\begin{algorithmic}[1]
\renewcommand{\algorithmicrequire}{\textbf{Input:}}
\renewcommand{\algorithmicensure}{\textbf{Output:}}

\Require $\eta_{l}$, $\eta_{g}$: local and global learning rates, $\sigma$: noise parameter, $C$: clipping bound, $T$: total round, $Q$: \#local epochs, $\mathbf{W} = (\mathbf{w}_1,...,\mathbf{w}_{|S|})$: matrix with weight for user $u$ and silo $s$, and $\forall u \in U$, $w_{s,u} \in \mathbf{w}_s$ and $\sum_{s \in S}{w_{s,u}}=1$

\Procedure{Server}{}
    \State Initialize model $x_{0}$
    \For{each round $t=0, 1, \ldots, T-1$}
        \For{each silo $s \in S$}
            \State $\Delta^{s}_{t} \leftarrow$ \textsc{Client}$(x_{t}, \mathbf{w}_{s}, C, \sigma, \eta_{l})$
        \EndFor
    \State $x_{t+1} \leftarrow x_{t} + \eta_{g} \frac{1}{|U||S|}\sum_{s\in S}{\Delta^{s}_{t}}$
    \EndFor
\EndProcedure

\State \textbf{/* Client algorithm for ULDP-AVG */}
\Procedure{Client}{$x_{t}, \mathbf{w}_{s}, C, \sigma, \eta_{l}$} \Comment{For ULDP-AVG}
    \For{user $u \in U$} \Comment{per-user training with $\mathcal{D}_{s,u}$}
        \State $x^{s,u}_{t} \leftarrow x_{t}$
        \For{epoch $q=0, 1, \ldots, Q-1$}
            \State Compute stochastic gradients $g^{s,u}_{t,q}$ \\ \Comment{$\mathbb{E}[g^{s,u}_{t,q}] = \nabla f_{s,u}(x^{s,u}_{t})$}
            \State $x^{s,u}_{t} \leftarrow x^{s,u}_{t} - \eta_{l} g^{s,u}_{t,q}$
        \EndFor
        \State $\Delta^{s,u}_{t} \leftarrow x^{s,u}_{t} - x_{t}$
        \State $\Tilde{\Delta}^{s,u}_{t} \leftarrow w_{s,u} \cdot \Delta^{s,u}_{t} \cdot \min{\left(1, \frac{C}{\lVert\Delta^{s,u}_{t}\rVert_{2}}\right)}$
    \EndFor
    \State $\Delta^{s}_{t} \leftarrow \sum_{u\in U}{\Tilde{\Delta}^{s,u}_{t}} + \mathcal{N}(0, I\sigma^2 C^2 / |S|)$
    \State \textbf{return} $\Delta^{s}_{t}$
\EndProcedure

\State \textbf{/* Client algorithm for ULDP-SGD */}
\Procedure{Client}{$x_{t}, \mathbf{w}_{s}, C, \sigma$} \Comment{For ULDP-SGD}
    \For{user $u \in U$}
        \State Compute stochastic gradients $g^{s,u}_{t}$
        \State $\Tilde{g}^{s,u}_{t} \leftarrow w_{s,u} \cdot g^{s,u}_{t} \cdot \min{\left(1, \frac{C}{\lVert g^{s,u}_{t}\rVert_{2}}\right)}$
    \EndFor
    \State $g^{s}_{t} \leftarrow \sum_{u \in U}{\Tilde{g}^{s,u}_{t}} + \mathcal{N}(0, I\sigma^2 C^2 / |S|)$
    \State \textbf{return} $g^{s}_{t}$
\EndProcedure

\end{algorithmic}
\end{algorithm}

To directly satisfy ULDP without using group-privacy, we design ULDP-AVG and ULDP-SGD (described in Algorithm \ref{alg:uldpavg}) .
These are the same as the relationship between (DP-)FedAVG and (DP-)FedSGD \cite{mcmahan2017learning}.
In most cases, FedAVG is better in communication-cost and privacy-utility trade-offs.
FedSGD might be preferable only when we have fast networks.
In the following analysis, we focus on ULDP-AVG since it essentially generalizes ULDP-SGD which has only a single SGD step and shares the gradients.

Intuitively, ULDP-AVG limits each user's contribution to the global model by training the model for each user in each silo and performing per-user per-silo clipping across all silos with globally prepared clipping weights.
In each round, ULDP-AVG computes model delta using a per-user dataset in each silo to achieve ULDP: selecting a user (Line 9), training local model with $Q$ epochs using only the selected user's data (Lines 11-14), calculating model delta (Line 15) and clipping the delta (Line 16).
These clipped deltas $\Delta^{s,u}_t$ are then weighted by $w_{s,u}$ (Line 16) and summed for all users (Line 16).
As long as the weights $w_{s,u}$ satisfy constraints $\forall u \in U$, $w_{s,u} > 0$ and $\sum_{s \in S}{w_{s,u}}=1$, each user's contribution, or \textit{sensitivity}, to the delta aggregation $\sum_{s\in S}{\Delta^{s}_{t}}$ is limited to $C$ at most.
This allows ULDP-AVG to provide user-level privacy.
We will discuss better ways to determine $\mathbf{W}$ later, but a simple way is to set $w_{s,u} = 1/|S|$.
Compared to DP-FedAVG, ULDP-AVG increases computational cost due to per-user local training iteration but keeps communication costs the same, which is likely acceptable in the cross-silo FL setting.

\begin{theorem}
\label{theo:uldp_avg}
For any $0 < \delta < 1$ and $\alpha > 1$, given noise multiplier $\sigma$, ULDP-AVG satisfies $(\epsilon=\frac{T\alpha}{2\sigma^2} + \log{((\alpha-1)/\alpha)} - (\log{\delta} + \log{\alpha})/(\alpha-1), \delta)$-ULDP after $T$ rounds.
\end{theorem}

    
\begin{algorithm}[t]
    \small
    \caption{ULDP-AVG \textcolor{red}{with user-level sub-sampling}}
    \label{alg:avg_w_sampling}
    \begin{algorithmic}[1]
    \renewcommand{\algorithmicrequire}{\textbf{Input:}}
    \renewcommand{\algorithmicensure}{\textbf{Output:}}
    
    \Require $\eta_{l}$, $\eta_{g}$, $\sigma$, $C$, $T$, $Q$, $\mathbf{W}$, \textcolor{red}{$q$: user-level sub-sampling probability}

    \Procedure{Server}{}
        \State Initialize model $x_{0}$
        \For{each round $t=0, 1, \ldots, T-1$}
            \textcolor{red}{
                \State $U_t \leftarrow$ Poisson sampling from $U$ with probability $q$
                \For{each silo $u \in U_t$}
                    \For{each silo $s \in S$}
                        \State \textcolor{red}{$w_{s,u} \leftarrow 0$} \Comment{set 0 if user is not sampled.}
                    \EndFor
                \EndFor
            }
            \For{each silo $s \in S$}
                \State $\Delta^{s}_{t} \leftarrow$ \textsc{Client}$(x_{t}, \mathbf{w}_{s}, C, \sigma, \eta_{l})$ \Comment{same as ULDP-AVG}
            \EndFor
        \State $x_{t+1} \leftarrow x_{t} + \eta_{g} \frac{1}{\textcolor{red}{q}|U||S|}\sum_{s\in S}{\Delta^{s}_{t}}$
        \EndFor
    \EndProcedure
    
    \end{algorithmic}
\end{algorithm}

\begin{remark}
{\normalfont
    For further privacy amplification, we introduce user-level sub-sampling, which can make RDP smaller according to sub-sampled amplification theorem (Lemma \ref{lemma:rdp_subsampled}) \cite{wang2019subsampled}.
    User-level sub-sampling must be done globally across silos.
    This sub-sampling can be implemented in the central server by controlling the weight $\mathbf{W}$ for each round, i.e., all users not sub-sampled are set to 0 as shown in Algorithm \ref{alg:avg_w_sampling}.
    This may violate privacy against the server but does not affect the DP when the final model is provided externally as discussed in C.3 of \cite{agarwal2021skellam}.
    Our following experimental results demonstrate the effectiveness of user-level sub-sampling.
}
\end{remark}

\begin{figure}[t]
    \includegraphics[width=0.95\linewidth]{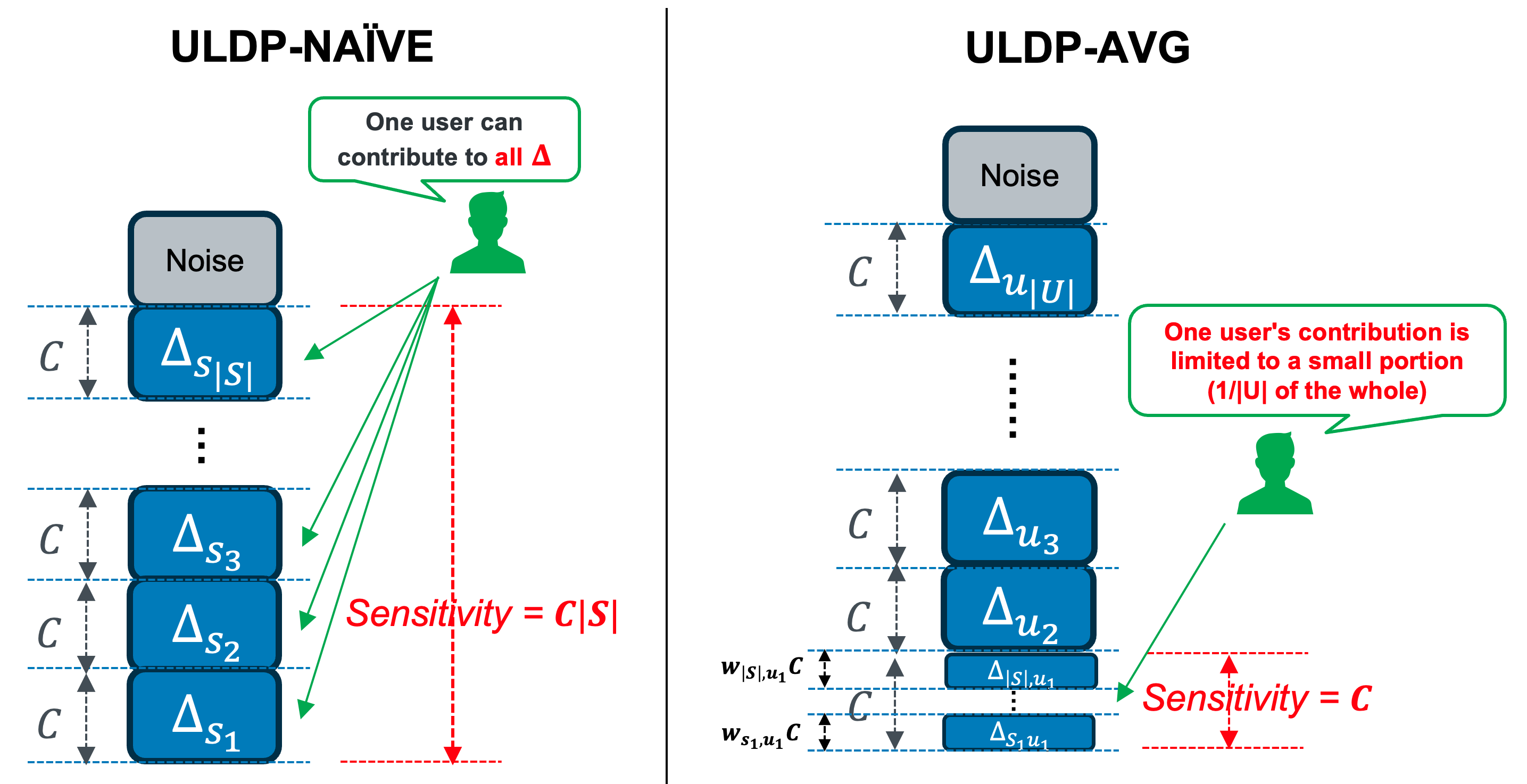}
    \caption{
    An intuitive illustration of the difference between ULDP-NAIVE and ULDP-AVG.
    In ULDP-NAIVE, every user can contribute to whole model deltas. In ULDP-AVG, one user's contribution is limited to a small portion, i.e.,  $1/|U|$ of the whole model delta, which reduces user-level sensitivity. 
    }
    \label{fig:intuitive_diff}
\end{figure}

\noindent
\textbf{Comparison to baselines.} Compared to ULDP-GROUP, ULDP-AVG satisfies ULDP without group-privacy, thus avoiding the large privacy bound caused by group-privacy conversion, the need to choose a group size $k$, and removing records.
ULDP-AVG can be used for an arbitrary number of records per user.
Also, we illustrate the intuitive difference between ULDP-NAIVE and ULDP-AVG in Figure \ref{fig:intuitive_diff}.
Fundamentally, per-user clipping can be viewed as cross-user FL (instead of cross-silo FL), ensuring that each user contributes only to their user-specific portion of the aggregated model updates (i.e., $\sum_{s\in S}{\Tilde{\Delta}^{s,u}_{t}}$) instead of the entire aggregated update (i.e., $\sum_{s\in S}{\Delta^{s}_{t}}$), thereby reducing sensitivity (as illustrated in Figure \ref{fig:intuitive_diff}).
The user contributes only $1/|U|$ of the entire aggregated model update, which is especially effective when $|U|$ is large, as in cross-silo FL (i.e., $|S|\ll|U|$).
Moreover, computing the model delta at the user level leads to lower Gaussian noise variances due to large $|U|$, while it also introduces new biases.
The overhead due to such biases can also be seen in the convergence analysis, motivating a better weighting strategy to reduce this overhead.

\smallskip
\noindent
\textbf{Convergence analysis (sketch)\footnote{Please refer to the Appendix \ref{appendix:convergence} for a rigorous result where we theoretically analyze the convergence of ULDP-AVG to compare with existing methods.}.}
Here, we present a high-level and intuitive summary of the convergence characteristics of ULDP-AVG, compared to existing methods.
ULDP-AVG partially recaptures the standard convergence bounds of FedAVG by considering user-silo pairs as participants and by setting the global and local learning rates under specific conditions.
However, compared to FedAVG, an additional noise term for DP and a bias due to user-silo granularity are introduced, which hinder convergence.
The former, a noise term, is also present in DP-FedAVG and can diminishes as the number of users increases.
The latter bias term can be minimized through the strategic weighting of weights in ULDP-AVG. Thus, a more creative use of weights, as will be explained in the next section, could result in improved convergence.

\section{Enhanced Weighting and Private weighting protocol}
\label{sec:private_method}
\subsection{The weighting protocol}
In considering the bias observed in the convergence analysis on ULDP-AVG (described in detail in Remark \ref{remark:grad_var} in Appendix, we employed uniform clipping weights in the ULDP-AVG algorithm (Algorithm \ref{alg:uldpavg} Line 16), i.e., for any $s \in S$ and $u \in U$, we set $w_{s,u}=1/|S|$, as a simple solution without privacy violation.
Now we propose an enhanced weighting strategy that aims to reduce the bias.
We set a weight $w^{opt}_{s,u}$ for $C_{s,u}$ according to the number of records for user $u$ in silo $s$, following the heuristic that a gradient computed from a large number of records yields a better estimation that is closely aligned with the average.
This results in smaller bias.
That is, let $n_{s,u}$ be the number of records for user $u$ in silo $s$, we set the weight as follows:
{
    \begin{equation}
    \label{eq:optimal_weight}
    \begin{aligned}
    w^{opt}_{s,u} := \frac{n_{s,u}}{\sum_{s\in |S|}{n_{s,u}}}.
    \end{aligned}
    \end{equation}
}
\unskip We empirically demonstrate the effectiveness of this strategy later.

\smallskip
\noindent
\textbf{Private weighting protocol.}
Given the above weighting strategy that relies on number of user records per user per silo, the crucial question arises: \textit{how can this be implemented without violating privacy?}
A central server could aggregate histograms encompassing the user population (number of records per user) within each silo's dataset.
Subsequently, the server could compute the appropriate weights for each silo and distribute these weights back to the respective silos.
However, it raises significant privacy concerns, since the histograms are directly shared with the server.
Moreover, when the server broadcasts the weights back to the silos, it enables an estimation of the entire histogram of users across all the silos, posing a similar privacy risk against other silos.
In essence, the privacy protection is necessary in both directions. 
This is challenging.
Additive homomorphic encryption, such as Paillier cryptosystem, is often used for this situation \cite{9476969}, but it is impossible to securely compute inverse values (for the weight in Eq. (\ref{eq:optimal_weight})). 
Note that unlimited records per user make DP impractical for protection.

To address this privacy issue, we design a novel private weighting protocol to securely aggregate the user histograms and also to securely perform local training and model aggregation.
The protocol leverages well-established cryptographic techniques, including secure aggregation \cite{bonawitz2017practical, so2023securing}, the Paillier cryptosystem \cite{9476969}, and multiplicative blinding \cite{damgaard2007efficient}.
Intuitively, the protocol employs multiplicative blinding to hide user histograms against the server while allowing the server to compute inverses of blinded histograms to compute the weights (Eq. (\ref{eq:optimal_weight})).
Subsequently, the server employs the Paillier encryption to conceal the inverses of blinded histograms because the silo knows the blinded masks.
This enables the server and silos to compute private weighted sum aggregation with its additive homomorphic property.

\begin{figure}
\begin{protocol}{proto:1}{Private Weighting Protocol}
{\small
\textit{Inputs:} Silo $s \in S$ that holds an dataset with $n_{s,u}$ records for each user $u \in U$. $\mathcal{A}$ is central aggregation server. $N_{\text{max}}$ is upper bound on the number of records per user, e.g., 2000. $P$ is precision parameter, e.g, $10^{-10}$. $\lambda$ is security parameter, e.g., $3072$-bit security.
\begin{enumerate}
  \item \textbf{Setup.}
  \begin{enumerate}
    \item
    $\mathcal{A}$ generates Paillier keypairs ($PK$, $SK$) with the given security parameter $\lambda$ and sends the public key $PK$ to all silos. All silos $s$ generate DH keypairs ($pk_{s}$, $sk_{s}$) with the same parameter $\lambda$ and  transmit their respective public key $pk_{s}$ to $\mathcal{A}$. Both $\mathcal{A}$ and all silos compute $C_{\text{LCM}}$, which is the least common multiple of all integers up to $N_{\text{max}}$. The modulus $n$ included in $PK$ is used for the finite field $\mathbb{F}_n$ by $\mathcal{A}$ and all silos. 
    
    \item
    After receiving all $pk_{s}$, $\mathcal{A}$ broadcasts all DH public keys $pk_{s}$ to all $s$.
    All $s$ compute shared keys $sk_{s,s'}$ from $sk_{s}$ and received public keys $pk_{s'}$ for all $s' \in S$.

    \item 
    Silo $0$ ($\in S$) generates a random seed $R$ and encrypts $R$ using $sk_{0,s'}$ to obtain $Enc(R)$ and sends $Enc(R)$ to $s'$ via $\mathcal{A}$ for all $s'$.
    All $s \in S \setminus 0$ receive and decrypt $Enc(R)$ with $sk_{s,0}$ and get $R$ as a shared random seed.

    \item
    All $s$ generate multiplicative blind masks $r_{u} \in \mathbb{F}_n$ with the same $R$ and compute blinded histogram as $B(n_{s,u}) \equiv r_{u} n_{s,u}\, (\mathrm{mod}\, n)$ for all $u \in U$.

    \item
    All $s$ generate pair-wise additive masks $r^{u}_{s,s'} \in \mathbb{F}_n$ employing $sk_{s,s'}$ for all $s'$ and $u$, with $r^{u}_{s,s'}=r^{u}_{s',s}$.
    Subsequently, they calculate the doubly blinded histogram as $B'(n_{s,u}) \equiv B(n_{s,u}) + \sum_{s<s'}{r^{u}_{s,s'}} - \sum_{s>s'}{r^{u}_{s,s'}}\, (\mathrm{mod}\, n)$.
    All $s$ send $B'(n_{s,u})$ to $\mathcal{A}$.
    $\mathcal{A}$ aggregates these contributions to compute $B(N_{u}) \equiv \sum_{s\in S}{B'(n_{s,u})}\, (\mathrm{mod}\, n)$ for each $u$, denoting $N_u = \sum_{s\in S}{n_{s,u}}$.

    \item
    $\mathcal{A}$ computes the inverse of $B(N_{u})$ as $B_{\text{inv}}(N_{u}) = B(N_{u})^{-1}$ for each $u$. This is the multiplicative inverse on $\mathbb{F}_n$, which is efficiently computed by the Extended Euclidean algorithm.

  \end{enumerate}

  \item \textbf{Weighting for each training round $t$.}
  \begin{enumerate}
    \item
    $\mathcal{A}$ encrypts $B_{\text{inv}}(N_{u})$ using Paillier's public key $PK$, resulting in $Enc_{\text{p}}(B_{\text{inv}}(N_{u}))$ for all $u$.
    If user-level sub-sampling is required, the server performs Poisson sampling with a given probability $q$ for each user before the encryption.
    For non-selected users, $B_{\text{inv}}(N_{u})$ is set to 0.
    If we require user-level sub-sampling, we perform Poisson sampling with given probability $q$ on the server for each user before the Paillier's encryption and set $B_{\text{inv}}(N_{u}) = 0$ for all users not selected.
    Subsequently, $\mathcal{A}$ broadcasts all $Enc_{\text{p}}(B_{\text{inv}}(N_{u}))$ to all silos.

    \item
    In each $s$, following the approach of ULDP-AVG, the clipped model delta $\Tilde{\Delta}^{s,u}_{t}$ is computed for each user $u$.
    The weighted clipped model delta is then calculated as
    {
        \begin{align}
            Enc_{\text{p}}&(\Tilde{\Delta}^{s,u}_{t}) = \notag \\
            \;\;\;\;\;\;\;\;\;\;\;\;\;\;\textsc{Encode}&(\Tilde{\Delta}^{s,u}_{t}, P, n) n_{s,u} r_{u} C_{\text{LCM}} Enc_{\text{p}}(B_{\text{inv}}(N_{u})) \nonumber.
        \end{align}
    }
    \unskip Let the Gaussian noise be $z^{s}_{t}$, we then compute $z'_{s} = \textsc{Encode}(z^{s}_{t}, P, n) C_{\text{LCM}}$.
    Note that we need to approximate real number $\Tilde{\Delta}^{s,u}_{t}$ and $z^{s}_{t}$ on a finite field using \textsc{Encode} (described in Algorithm \ref{alg:encode_decode}). Lastly, we compute the summation $Enc_{\text{p}}(\Delta^{s}_{t}) = \sum_{u\in U} {Enc_{\text{p}}(\Tilde{\Delta}^{s,u}_{t})} + z'_{s}$.

    \item
    In each $s$, random pair-wise additive masks are generated, and secure aggregation is performed on $Enc_{\text{p}}(\Delta^{s}_{t})$ mirroring the steps in 1.(f). Then, $\mathcal{A}$ gets $\sum_{s\in S}{Enc_{\text{p}}(\Delta^{s}_{t})}$. 
    $\mathcal{A}$ decrypts it with Paillier’s secret key $SK$ and decodes it by \textsc{Decode}$(\sum_{s\in S}{\Delta^{s}_{t}}, P, C_{\text{LCM}}, n)$ and recovers the aggregated value.

    \item
    Steps 2.(a) through 2.(c) are repeated for each training round.
  \end{enumerate}
\end{enumerate}
}
\end{protocol}
\end{figure}

\begin{algorithm}[t]
    \caption{\textsc{Encode} and \textsc{Decode}}
    \label{alg:encode_decode}
    \begin{algorithmic}[1]
    
    \Procedure{Encode}{$x, P, n$}       \Comment{e.g., $P=10^{-10}$}
        \State /* to turn floating point into fixed point */
        \State $x \leftarrow x/P$ \Comment{compute as floating point}
        \State $x \leftarrow x$ as integer
        \State $x \leftarrow x\, (\mathrm{mod}\, n)$  \Comment{to map integer $\mathbb{Z}$ into finite field $\mathbb{F}_n$}
        \State \textbf{return} $x$
    \EndProcedure

    \Procedure{Decode}{$x, P, C_{\text{LCM}}, n$}
        \State /* to map finite field $\mathbb{F}_n$ number into integer $\mathbb{Z}$ */
        \If{$x > n // 2$} \Comment{$//$ means integer division}
            \State $x \leftarrow x - n$
        \Else
            \State $x \leftarrow x$
        \EndIf
        \State /* compute as floating point */
        \State $x \leftarrow x / C_{\text{LCM}}$ \Comment{to remove $C_{\text{LCM}}$ factor}
        \State $x \leftarrow xP$ \Comment{to recover original magnitude}
        \State \textbf{return} $x$
    \EndProcedure
    
    \end{algorithmic}
\end{algorithm}

The details of the private weighting protocol are explained in Protocol \ref{proto:1}.
The protocol consists of a setup phase, which is executed only once during the entire training process, and a weighting phase, which is executed in each round of training.
In the setup phase, as depicted in (a-c) of Protocol \ref{proto:1}, the server generates a key-pair for Paillier encryption, while the silos establish shared random seeds through a Diffie–Hellman (DH) key exchange via the server.
Subsequently, in steps of (d-f), the blinded inverses of the user histogram are computed.
In the weighting phase, (a) the server prepares the encrypted weights, (b) the silos compute user-level weighted model deltas in the encrypted world, and (c) the server recovers the aggregated value.
It is important to note that in the Paillier cryptosystem, the plaintext $x$ exists within the additive group modulo $n$, while encrypted data (denoted as $Enc_{\text{p}}(x)$) belongs to the multiplicative group modulo $n^2$ with an order of $n$.
The system allows for operations such as addition of ciphertexts and scalar multiplication and addition on ciphertexts.

\smallskip
\noindent
\textbf{Private user-level sub-sampling.}
Note that our assumption so far is that the results of user-level sub-sampling (i.e., whether a user is sampled) are open to the aggregation server.
This is also the case in Protocol \ref{proto:1}. However, it could be hidden by combining the two-party verifiable sampling scheme with 1-out-of-P Oblivious Transfer (OT) as described in \cite{kato2021preventing}.
As an overview, for each user $u$, the server creates $P-1$ dummy data $Enc_{\text{p}}(0)$ for $Enc_{\text{p}}(B_{\text{inv}}(N_{u}))$ described in the step 2.(a) of Protocol \ref{proto:1}.
When the client performs OT on this data, the selection probability of $Enc_{\text{p}}(B_{\text{inv}}(N_{u}))$ is $\frac{1}{P}$ and that of $Enc_{\text{p}}(0)$ is $\frac{P-1}{P}$.
The selection of $Enc_{\text{p}}(B_{\text{inv}}(N_{u}))$ means that the user is not sampled by the user-level sub-sampling.
In this way, the server does not know which data was retrieved by the client from the OT, and the client cannot know the sampling result due to the Paillier encryption.
However, the expressed probability is likely to be less strict because it can only represent discrete probability distributions.
This process requires extra computational costs for both the server and the silo, proportional to the number of users, and should not be included if it is not necessary.

\subsection{Theoretical analysis}
We provide a theoretical analysis of this private weighting protocol (Protocol \ref{proto:1}) in terms of correctness and privacy.

\smallskip
\noindent \textbf{Correctness.}
The protocol must compute the correct result that is the same as non-secure method.
To this end, we consider the correctness of the aggregated data obtained in each round.

\begin{theorem}[Correctness of Protocol \ref{proto:1}]
\label{theo:correctness}
Let $\sum_{s\in S}\Delta^{s}_t$ with non-secure method be $\Delta$ and the one with the Protocol \ref{proto:1} be $\Delta_{\text{sec}}$, our goal is formally stated as $\Pr[|\Delta - \Delta_{\text{sec}}|_{\infty} > P] < negl$, where $P$ is a precision parameter and $negl$ signifies a negligible value.
\end{theorem}

\smallskip
\noindent \textbf{Privacy.}
In the protocol, both the central server and the silos do not get more information than what is available in the original ULDP-AVG while we perform the enhanced weighting strategy.

\begin{theorem}[Privacy of Protocol \ref{proto:1}]
\label{theo:privacy}
None of the parties learns $n_{s,u}$ other than their own users from the protocol.
\end{theorem}

\section{Experiments}
\label{sec:experiment}

In this section, we report the results of the experimental evaluation of our proposed methods.
We design experiments to answer the following questions: 
\begin{itemize}
    \item How much does our proposed method improve the privacy-utility trade-offs from baselines in terms of ULDP?
    \item How effective are enhanced weighting strategies and user-level sub-sampling in enhancing ULDP-AVG?
    \item How efficient is the proposed private weight protocol? Can it work for real-world data?
\end{itemize}

\noindent
All of our implementations and experimental settings are available\footnote{\url{https://github.com/FumiyukiKato/uldp-fl}}.

\subsection{Settings}
\label{sec:exp:settings}
We evaluate the privacy-utility trade-offs of the proposed methods (ULDP-AVG/ULDP-AVG-w/SGD), along with the  baselines (ULDP-NAIVE/GROUP-$k$) and a non-private baseline (FedAVG with two-sided learning rates \cite{yang2021achieving}, denoted by DEFAULT).
In ULDP-AVG/SGD, we set the weights as $w_{s,u}=1/|S|$ for all $s$ and $u$, the one using $w^{opt}_{s,u}$ is referred to as ULDP-AVG-w.
Regarding ULDP-GROUP-$k$, flags $\mathbf{B}$ are generated for existing records to minimize waste on filtered out records, despite the potential privacy concerns.
Various values, including the maximum number of user records (ULDP-GROUP-max), the median (ULDP-GROUP-median), 2, and 8, are tested as group size $k$ and we report GDP using group-privacy conversion of RDP.
In particular, ULDP-GROUP-max would represent an upper bound on the utility achieved by record-level DP in each silo (such as \cite{lowy2021private} and \cite{liu2022privacy}), since there are no deleted records.
In cases where $k$ is not a power of 2, the computed $\epsilon$ is reported for the largest power of 2 below $k$, showcasing the lower bound of GDP to underscore that $\epsilon$ is large.
The hyperparameters, including global and local learning rates $\eta_g$, $\eta_l$, clipping bound $C$, and local epoch $Q$, are set individually for each method. 
Execution times are measured on macOS Monterey v12.1, Apple M1 Max Chip with 64GB memory with Python 3.9 and 3072-bit security.
Most of the results are averaged over 5 runs and the colored area in the graph represents the standard deviation.

\subsubsection{Datasets}
Datasets used in the evaluation comprise real-world open datasets, including Credicard \cite{creditcard2018kaggle}, well-known image dataset MNIST, and two benchmark medical datasets for cross-silo FL \cite{ogier2022flamby}, HeartDisease and TcgaBrca.
Creditcard is a tabular dataset for credit card fraud detection from Kaggle.
We undersample the dataset and use about 25K training data and a neural network with about 4K parameters.
For MNIST, we use a CNN with about 20K parameters, 60K training data and 10K evaluation data, and assigned silos and users to all of the training data.
For HeartDisease and TcgaBrca, we use the same setting such as number of silos (4 and 6), data assignments to the silos, models, etc. as shown in \cite{ogier2022flamby}.
These two datasets are small and the model has less than 100 parameters.

For all datasets, we need to link all records to each user and silo.
We allocate the records to users and silos as follows.


    

  
    

\smallskip
\noindent \textbf{Record allocation for MNIST and Creditcard.}
We designed two different record distribution patterns, \textit{uniform} and \textit{zipf}, to model how user records are scattered across silos in the MNIST and Creditcard datasets.
Both distributions take the number of users $|U|$ and the number of silos $|S|$.
It associates each record with a user and a silo.
(1) In uniform, every record is assigned to a user with equal probability, and likewise, each record is assigned to a silo with equal probability.
(2) zipf combines two types of Zipf distributions. 
First, the distribution of the number of records per user follows a Zipf distribution.
Then, for each user, the numbers of records are assigned to different silos based on another Zipf distribution. 
Each of the two Zipf distributions takes a parameter $\alpha$ that determines the concentration of the numbers.
In the experiments, we used $\alpha=0.5$ for the first distribution and $\alpha=2.0$ for the second distribution.
This choice is rooted in the observation that the concentration of user records is not as high as the concentration in the silos selected by each user.
For Creditcard and MNIST, the number of silos $|S|$ is fixed at 5.
We used 100, 1000 for Creditcard as $|U|$ and 100, 10000 for MNIST.
For MNIST, we can require each user to have only 2 labels at most for non-i.i.d.

\smallskip
\noindent \textbf{Record allocation for HeartDisease and TcgaBrca.}
For the HeartDisease and TcgaBrca datasets, we adopted the same two distributions \textit{uniform} and \textit{zipf} as the above-mentioned ones.
In the benchmark datasets HeartDisease and TcgaBrca, all records are already allocated to silos and the number of records of each silo is fixed.
Therefore, the design of the user-record allocation is slightly different.
(1) In uniform, all records belong to one of the users with equal probability without allocation to silos.
(2) In zipf, the number of records for a user is first generated according to a Zipf distribution, and 80\% of the records are assigned to one silo, and the rest to the other silos with equal probability.
The priority of the silo is chosen randomly for each user. 
We used $\alpha=0.5$ for the parameter of the Zipf.
In TcgaBrca, Cox-Loss is used for loss function \cite{ogier2022flamby}, which needs more than two records for calculating valid loss and we set more than two records for each silo and user for per-user clipping of ULDP-AVG.


\subsection{Results}

\begin{figure}
    \centering
    \begin{subfigure}{\linewidth}
    \includegraphics[width=0.49\linewidth]{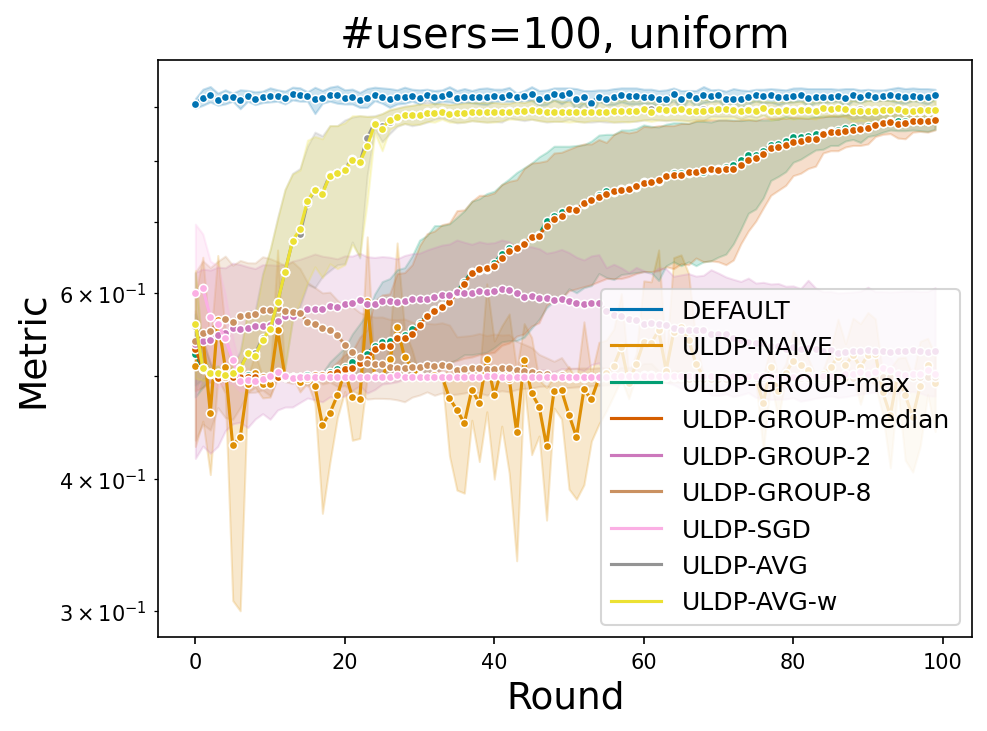}
    \hfill
    \includegraphics[width=0.49\linewidth]{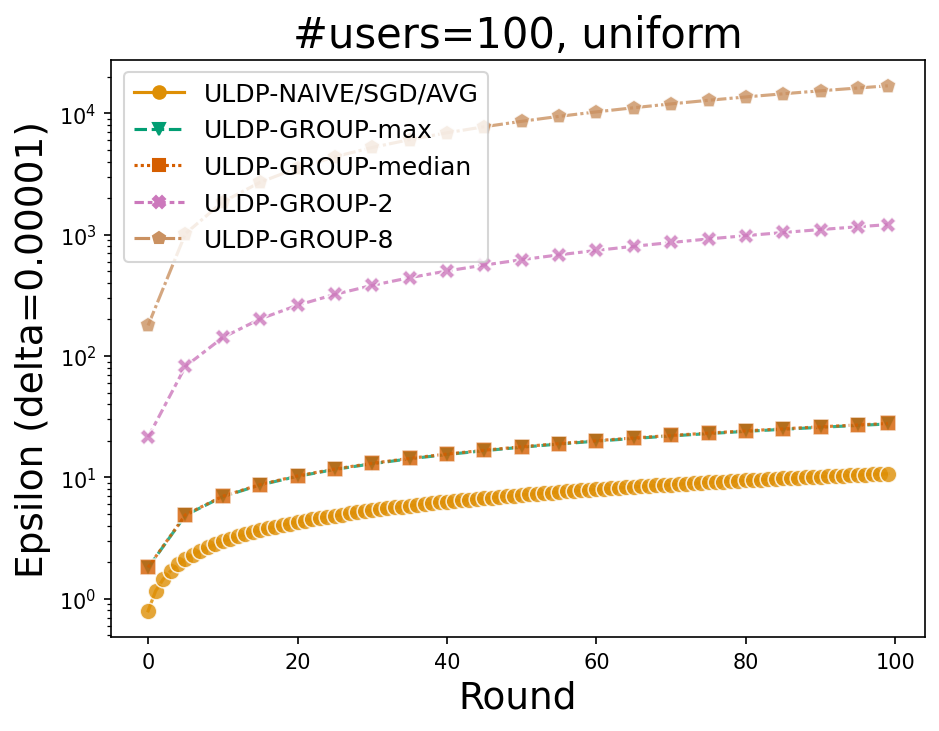}
    \caption{$n \approx 246$ ($|U|=100$), uniform.}
    \label{fig:creditcard_a}

    \includegraphics[width=0.49\linewidth]{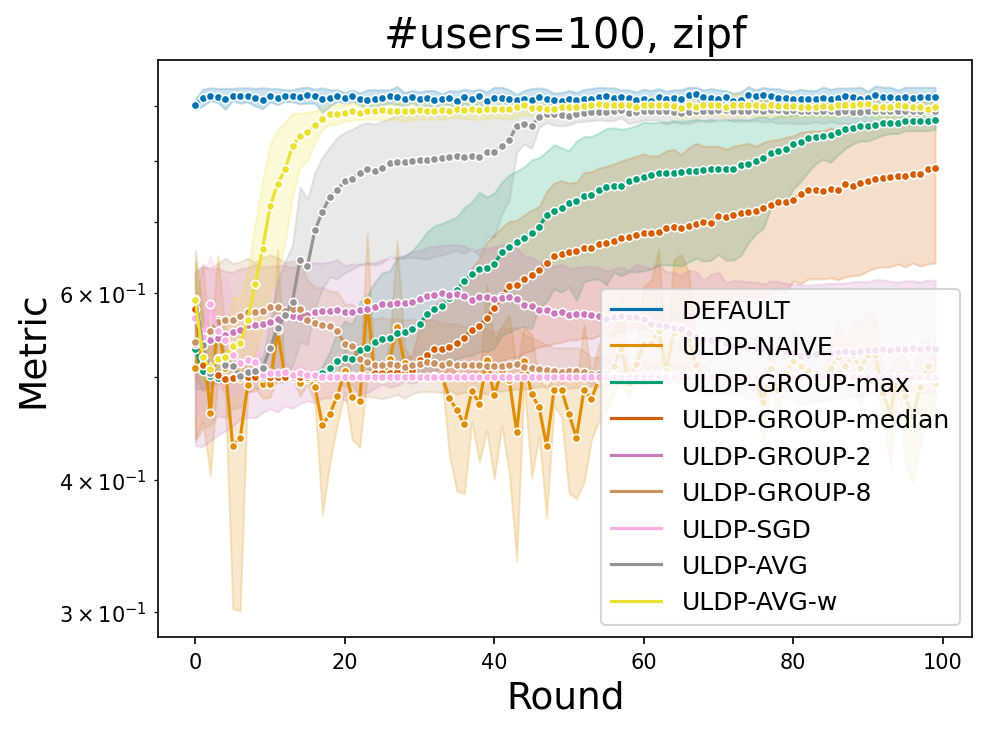}
    \hfill
    \includegraphics[width=0.49\linewidth]{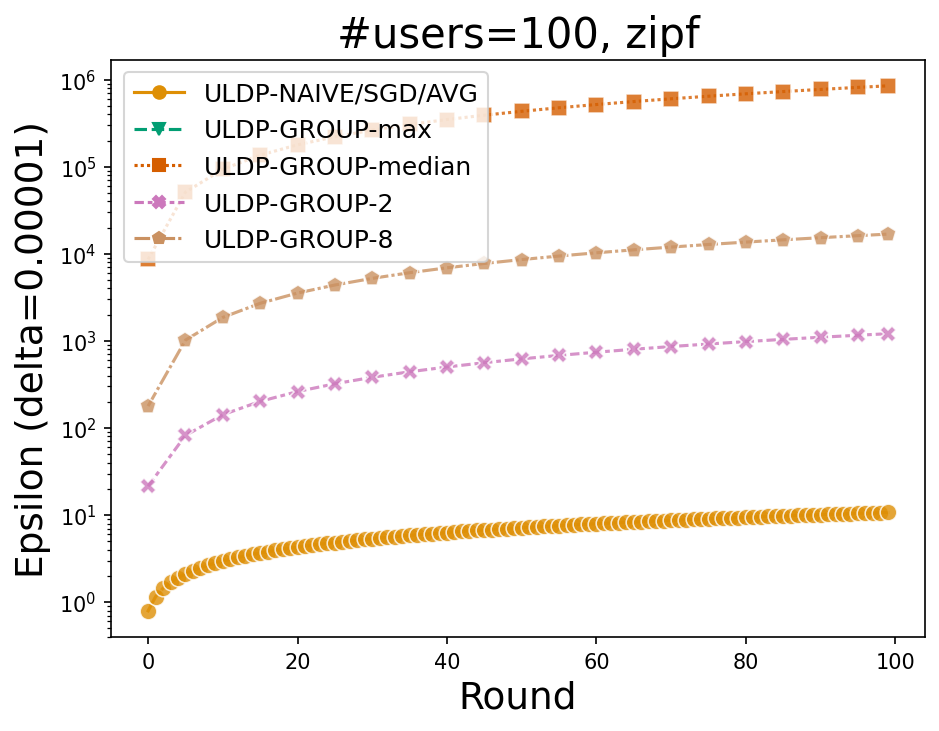}
    \caption{$n \approx 246$ ($|U|=100$), zipf.}
    \label{fig:creditcard_b}

    \includegraphics[width=0.49\linewidth]{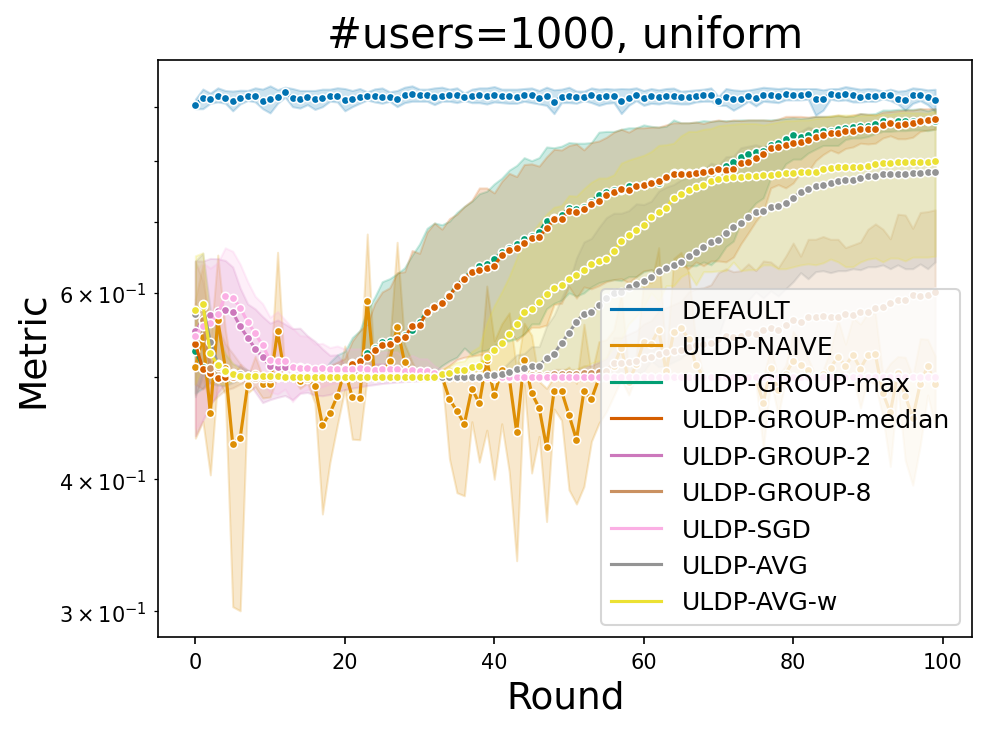}
    \hfill
    \includegraphics[width=0.49\linewidth]{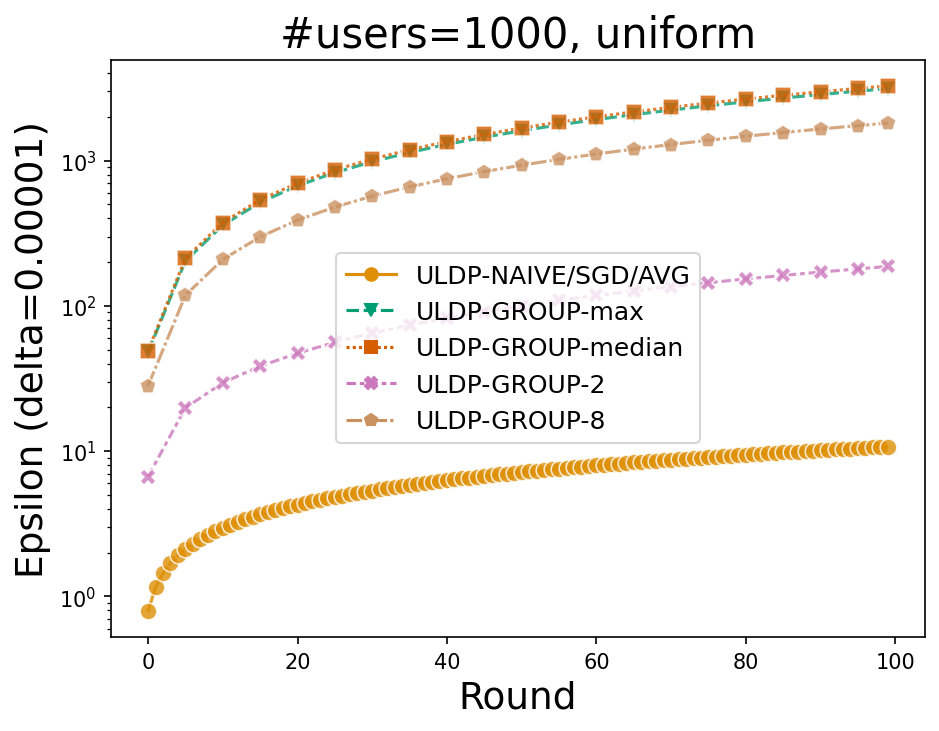}
    \caption{$n \approx 25$ ($|U|=1000$), uniform.}
    \label{fig:privacy_utility_creditcard_1000_uniform}

    \includegraphics[width=0.49\linewidth]{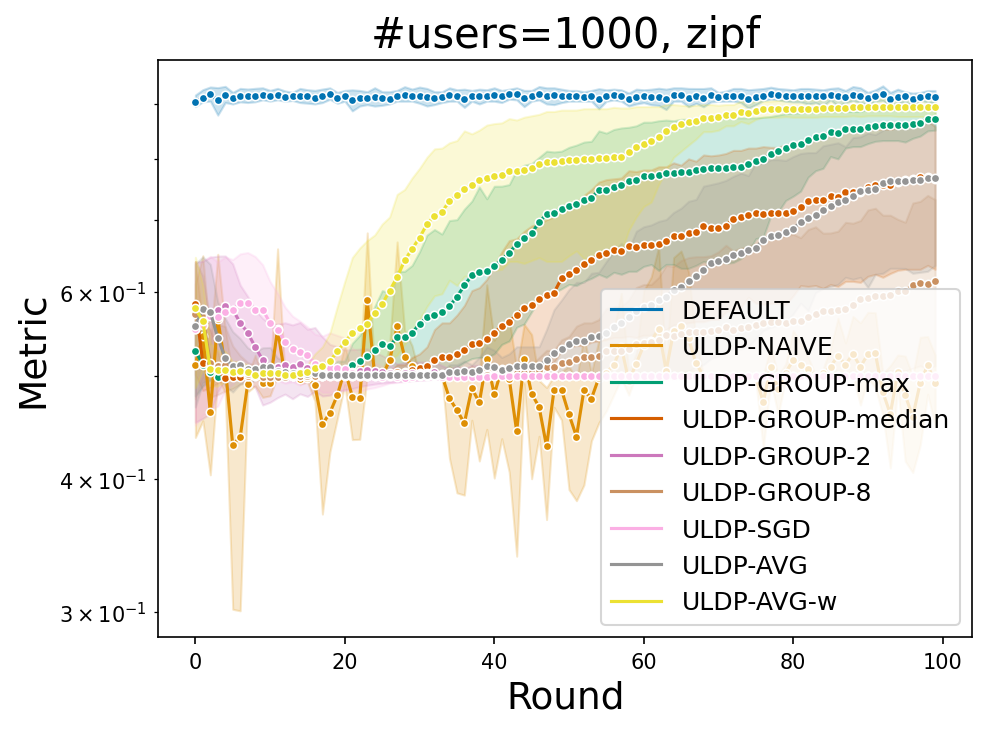}
    \hfill
    \includegraphics[width=0.49\linewidth]{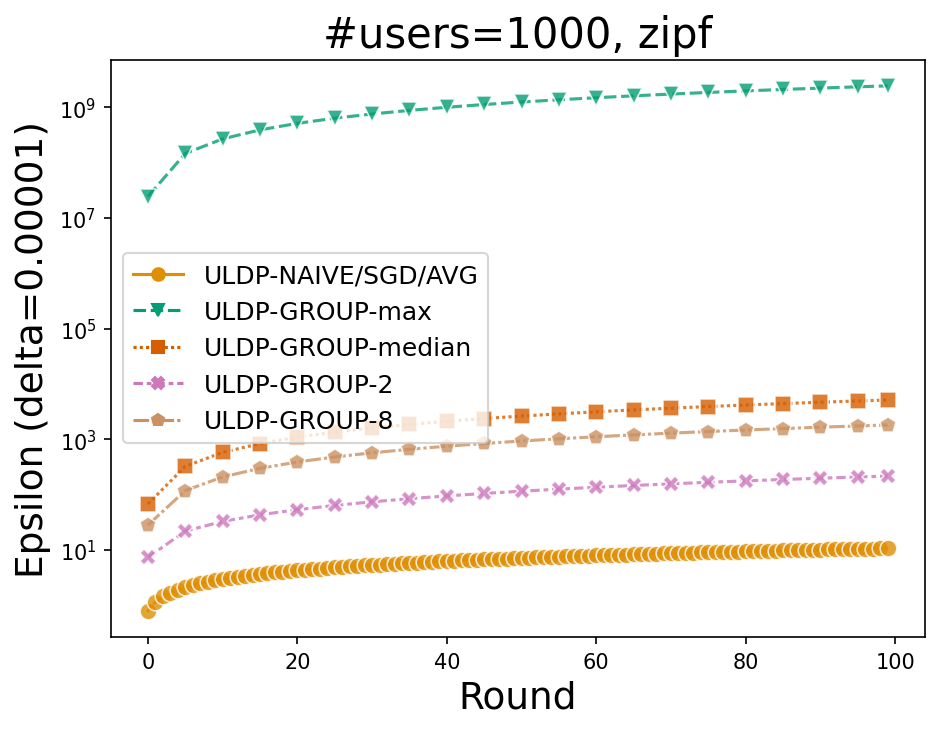}
    \caption{$n \approx 25$ ($|U|=1000$), zipf.}
    \label{fig:creditcard_d}
    \end{subfigure}
    \caption{Privacy-utility trade-offs on Creditcard dataset: Test Accuracy (Left), Privacy (Right).}
    \label{fig:creditcard}
\end{figure}

\begin{figure}
    \centering
    \begin{subfigure}{\linewidth}
        \includegraphics[width=0.32\linewidth]{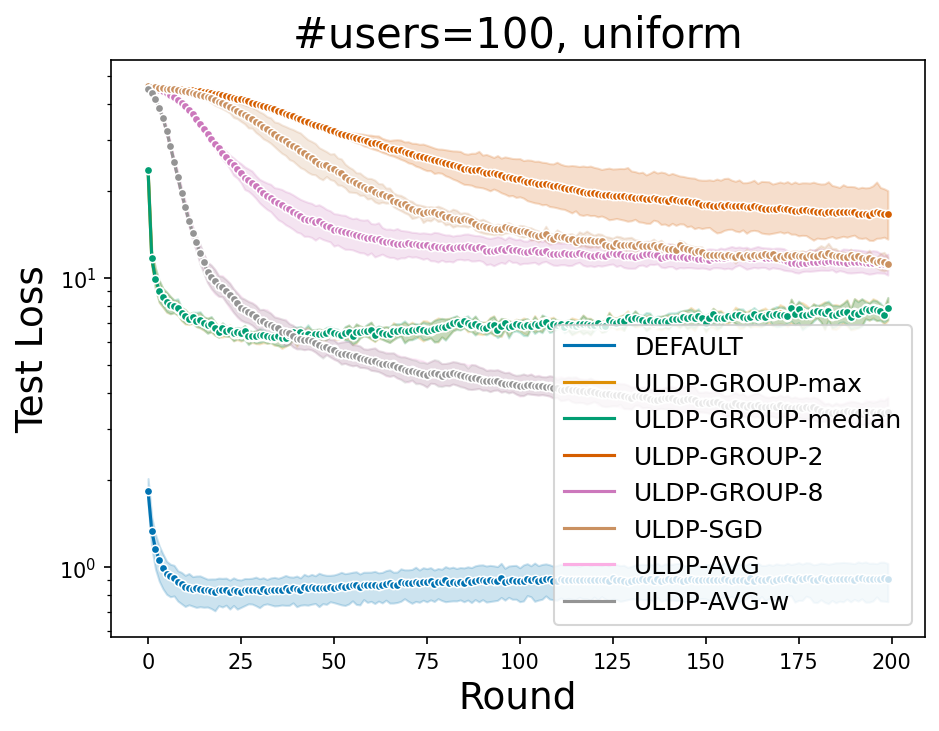}
        \hfill
        \includegraphics[width=0.32\linewidth]{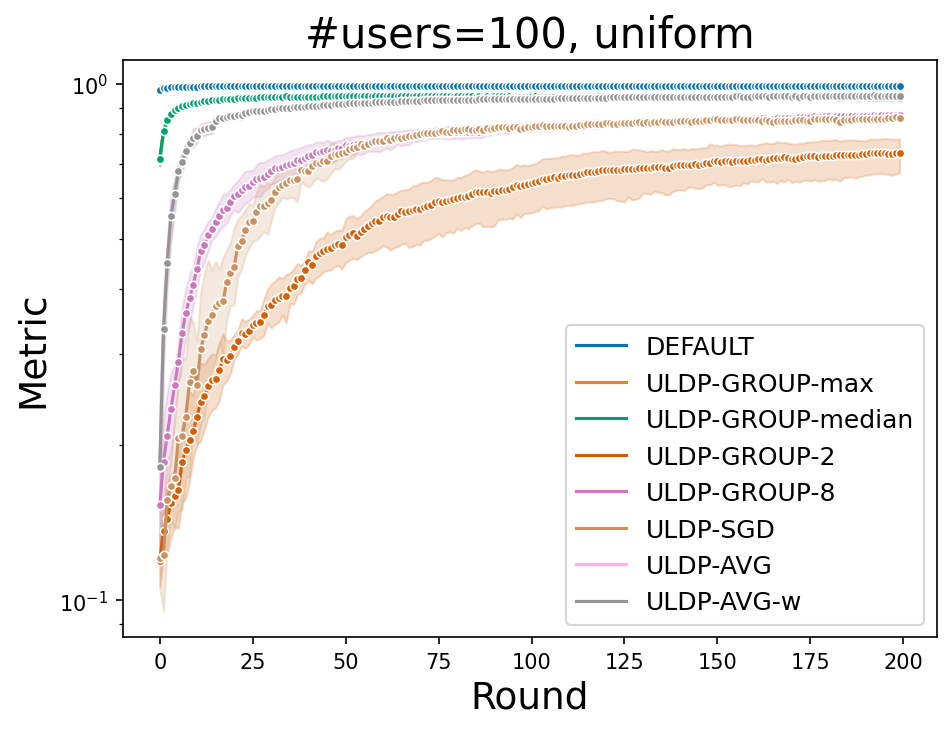}
        \hfill
        \includegraphics[width=0.32\linewidth]{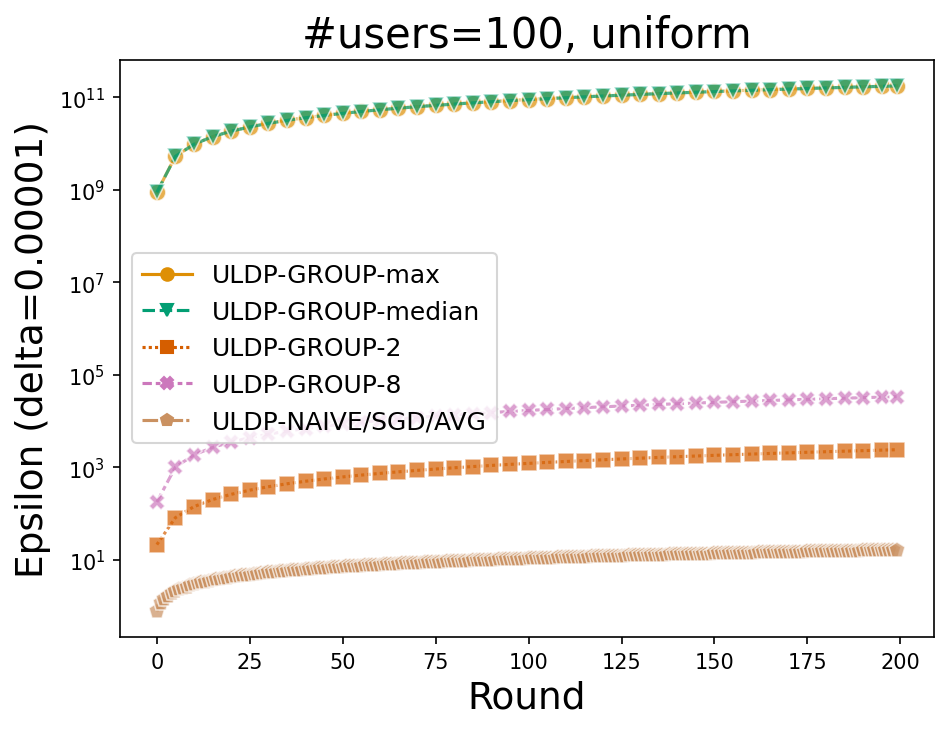}
        \caption{$n \approx 600$ ($|U|=100$), uniform.}
        \label{fig:privacy_utility_mnist_100_uniform}
    
        \includegraphics[width=0.32\linewidth]{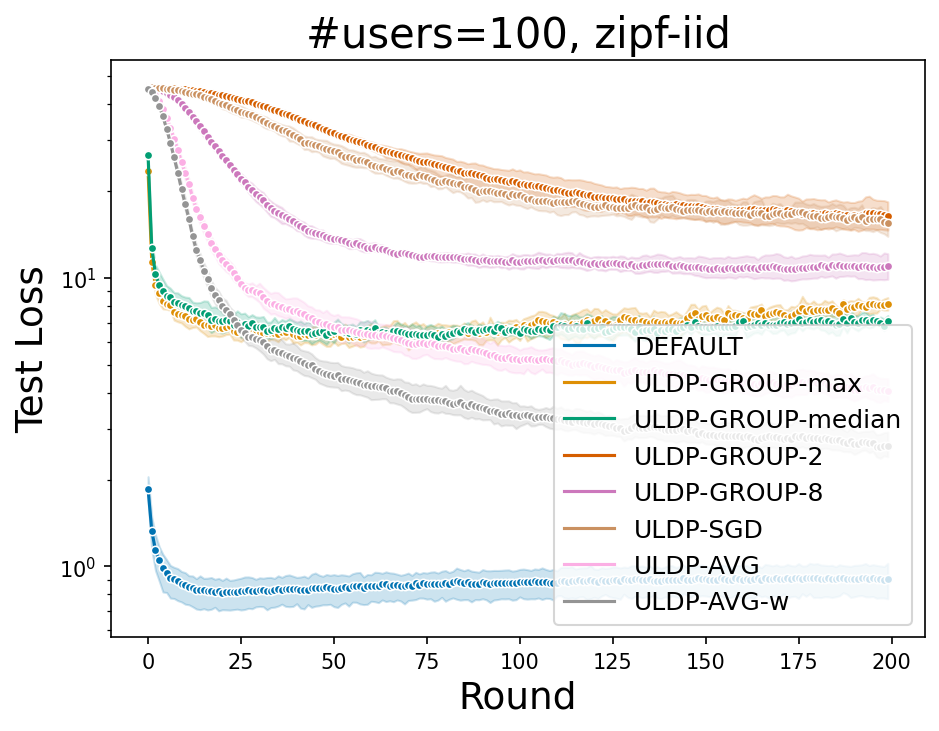}
        \hfill
        \includegraphics[width=0.32\linewidth]{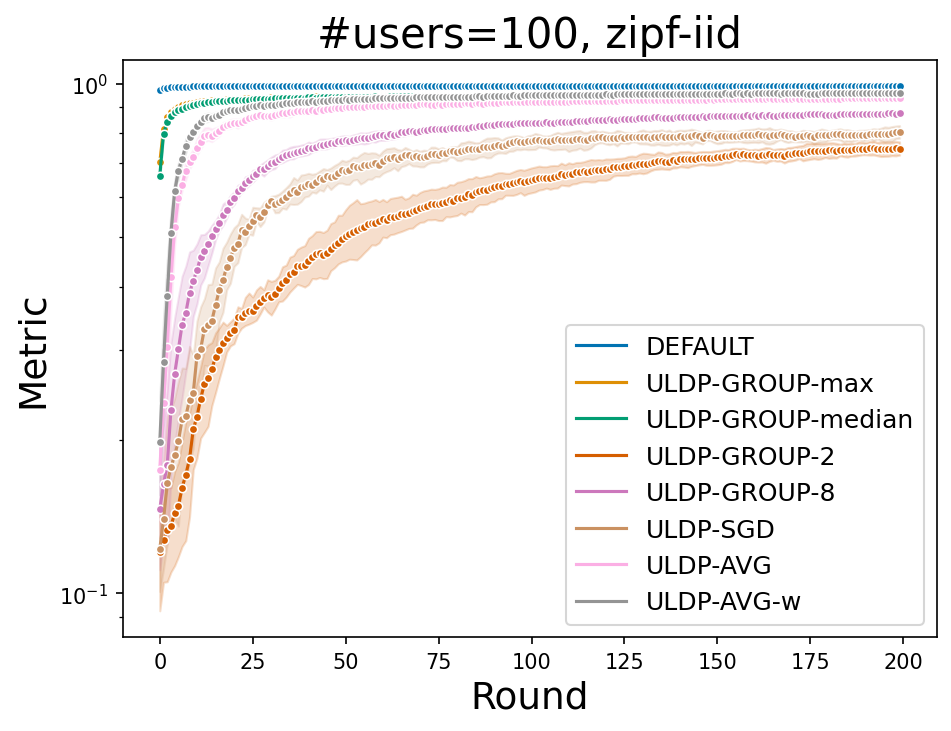}
        \hfill
        \includegraphics[width=0.32\linewidth]{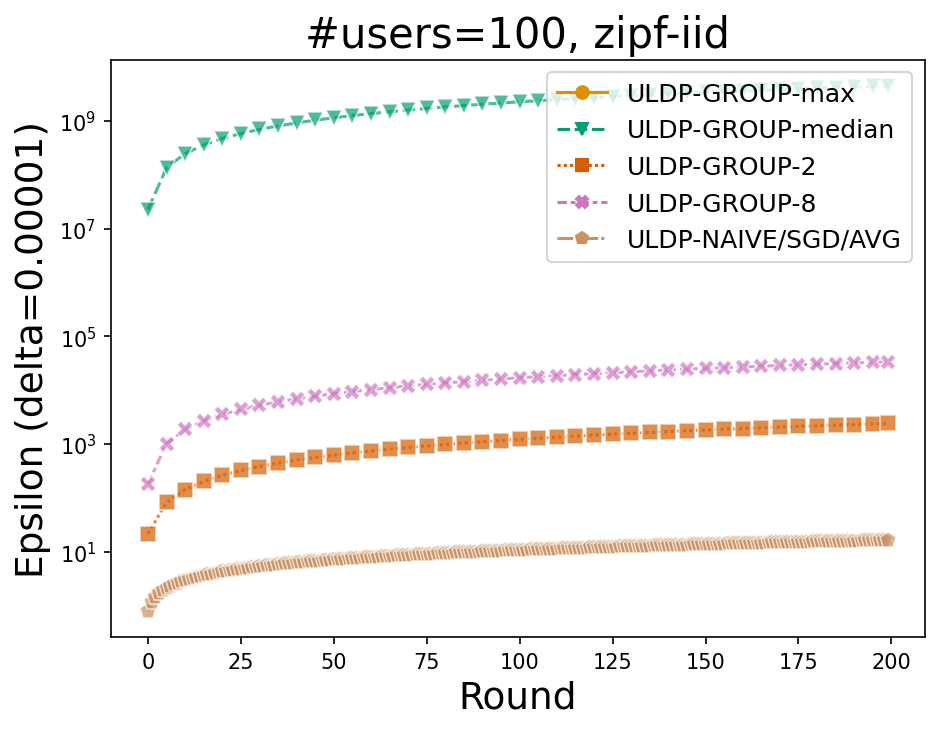}
        \caption{$n \approx 600$ ($|U|=100$), zipf, iid.}
        \label{fig:privacy_utility_mnist_100_zipf-iid}

        \includegraphics[width=0.32\linewidth]{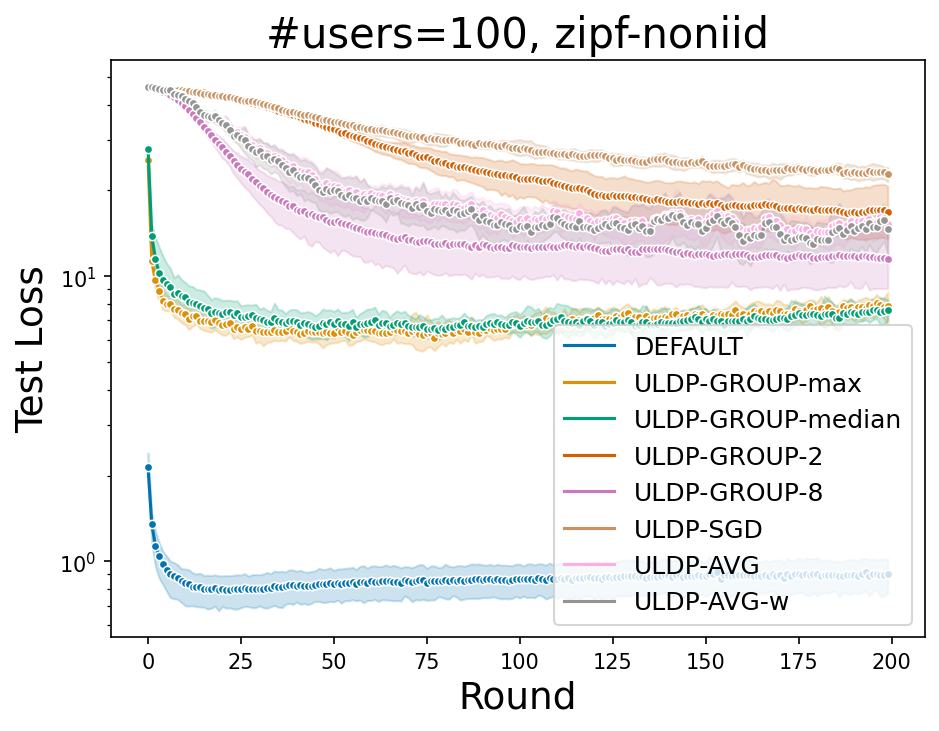}
        \hfill
        \includegraphics[width=0.32\linewidth]{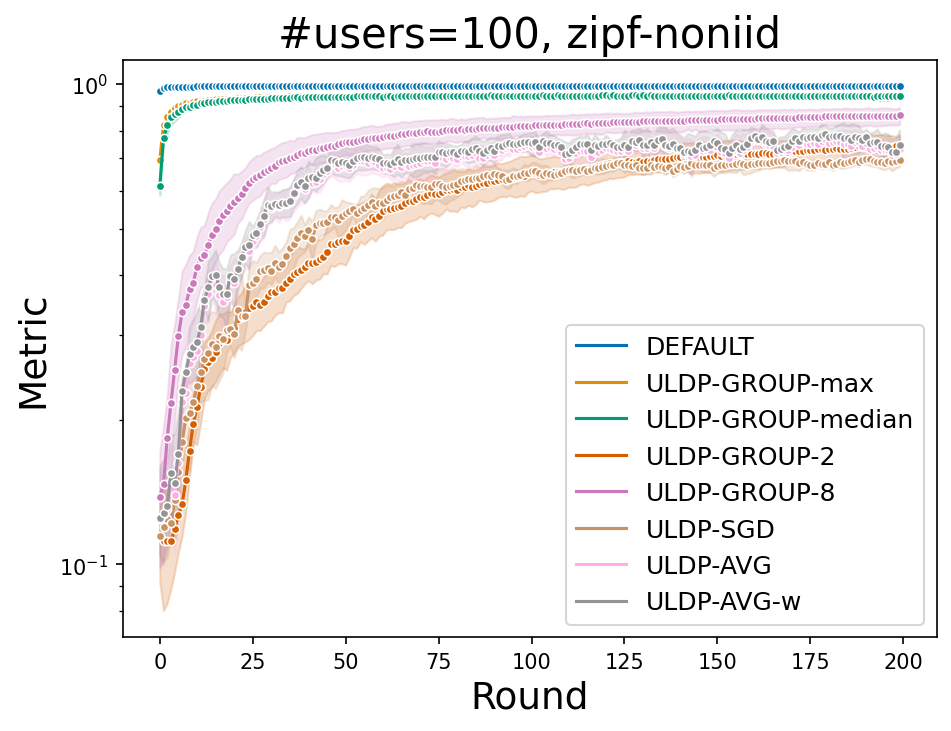}
        \hfill
        \includegraphics[width=0.32\linewidth]{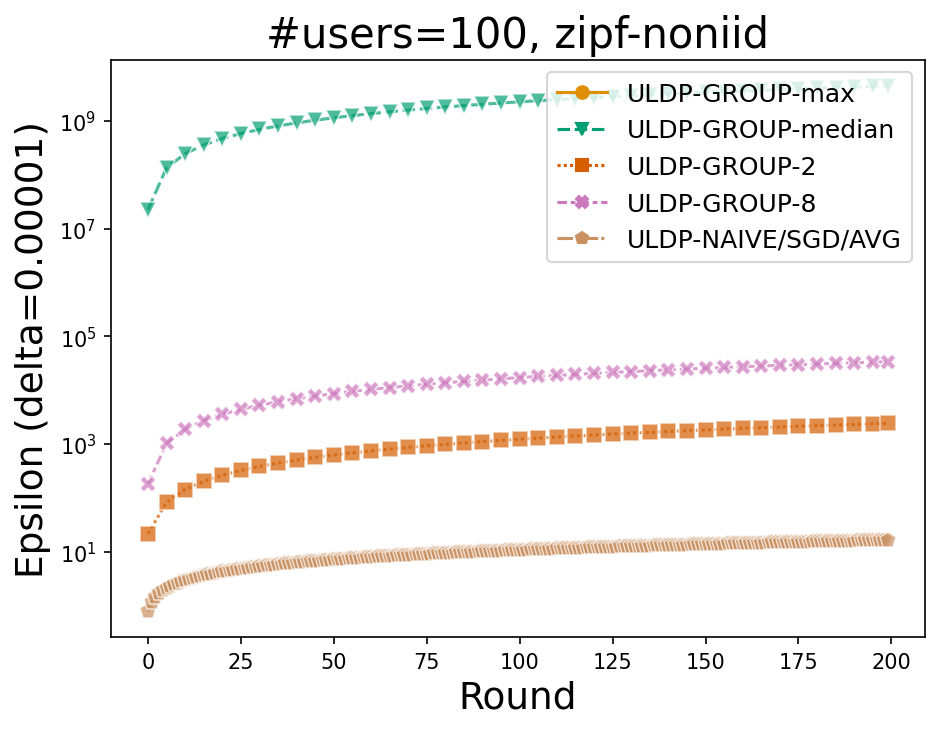}
        \caption{$n \approx 600$ ($|U|=100$), zipf, non-iid.}
        \label{fig:privacy_utility_mnist_100_zipf-noniid}

        \includegraphics[width=0.32\linewidth]{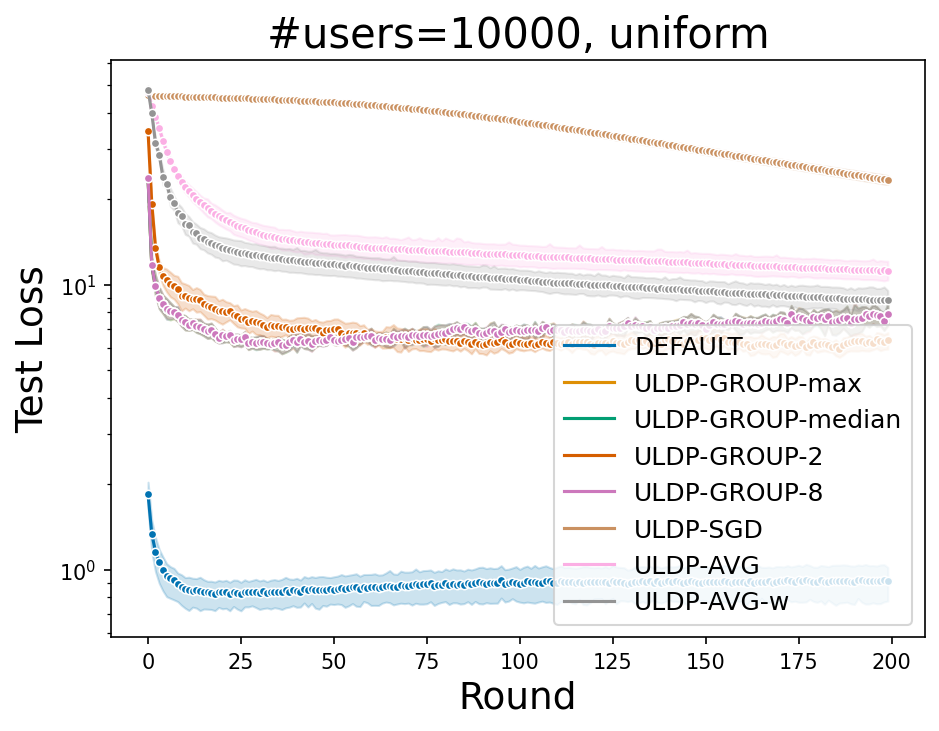}
        \hfill
        \includegraphics[width=0.32\linewidth]{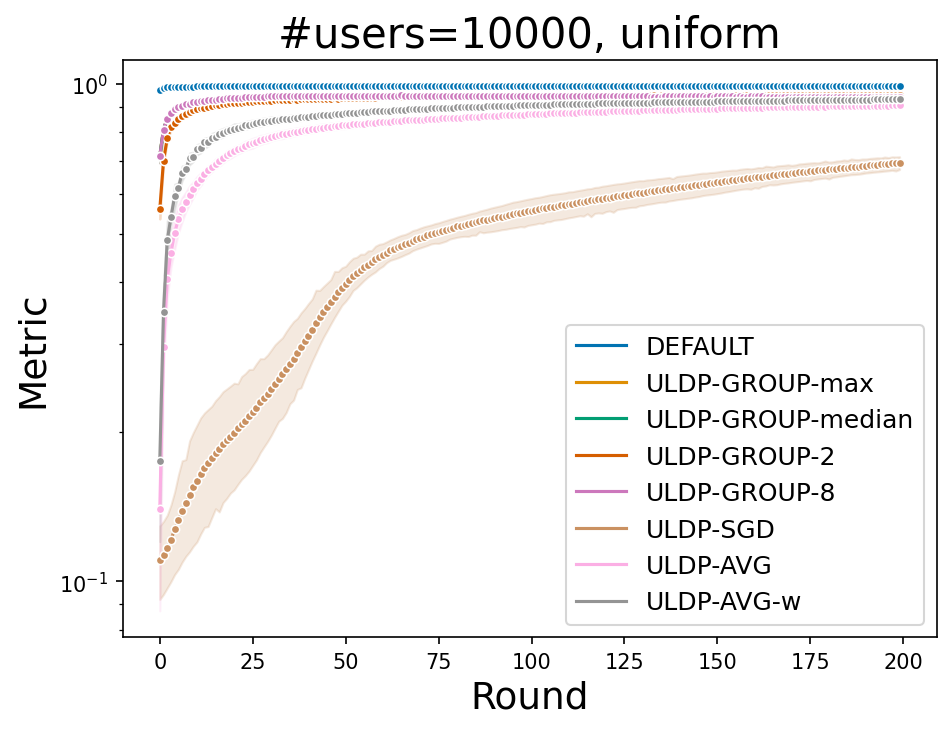}
        \hfill
        \includegraphics[width=0.32\linewidth]{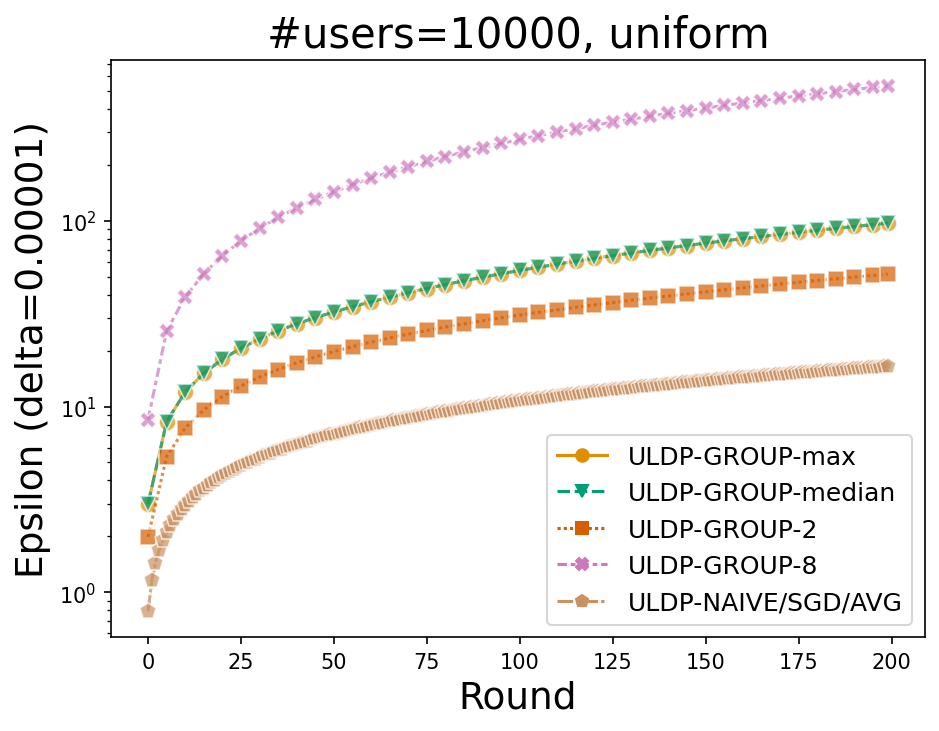}
        \caption{$n \approx 6$ ($|U|=10000$), uniform, iid.}
        \label{fig:privacy_utility_mnist_10000_uniform}

        \includegraphics[width=0.32\linewidth]{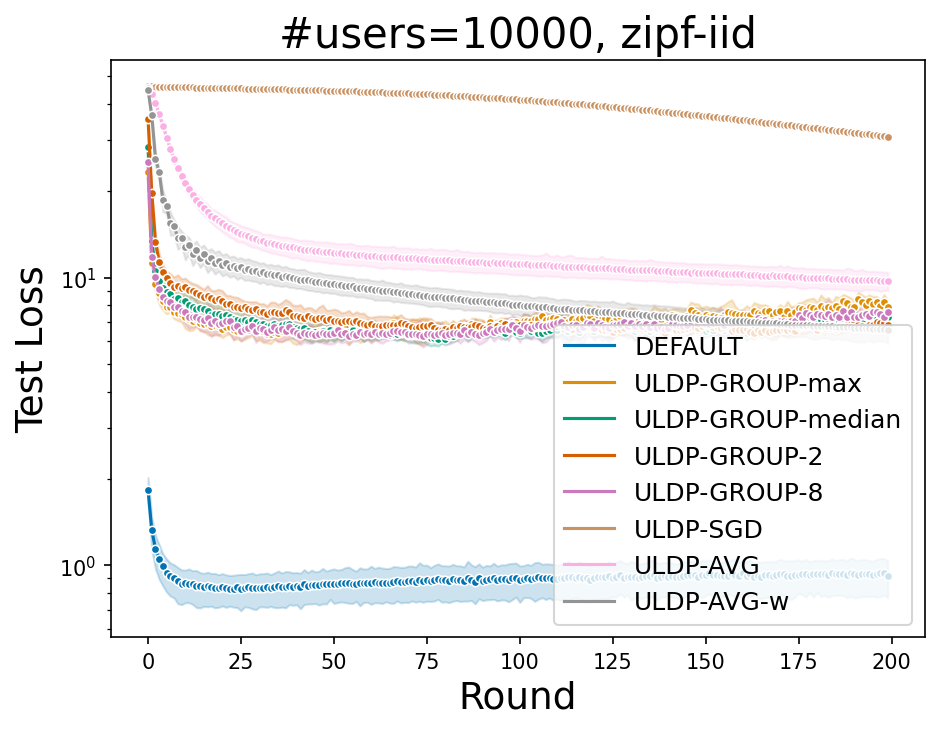}
        \hfill
        \includegraphics[width=0.32\linewidth]{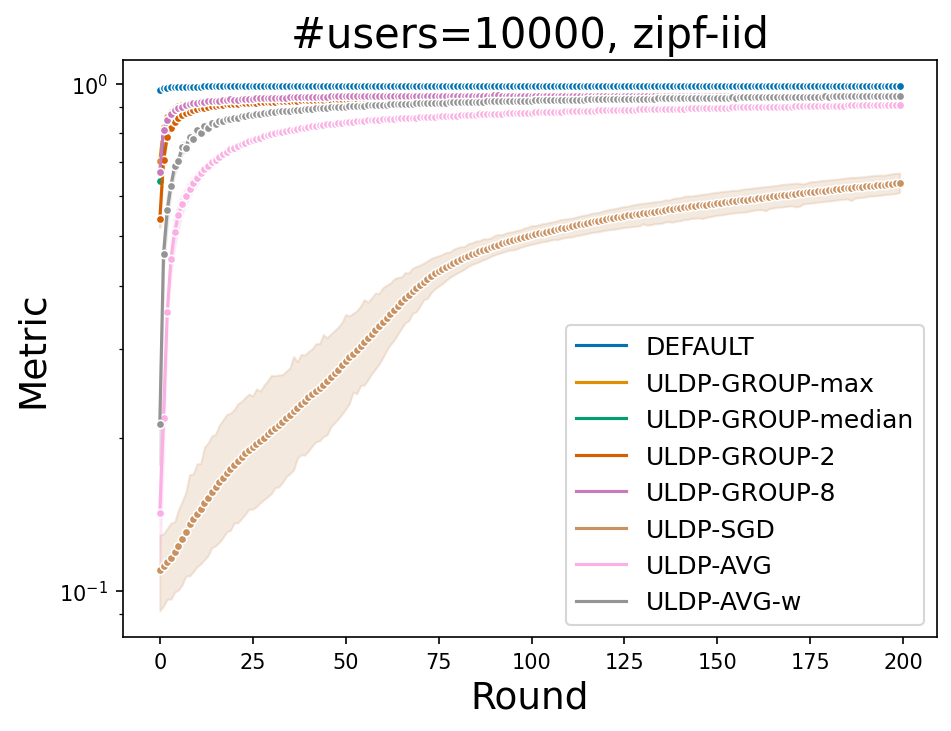}
        \hfill
        \includegraphics[width=0.32\linewidth]{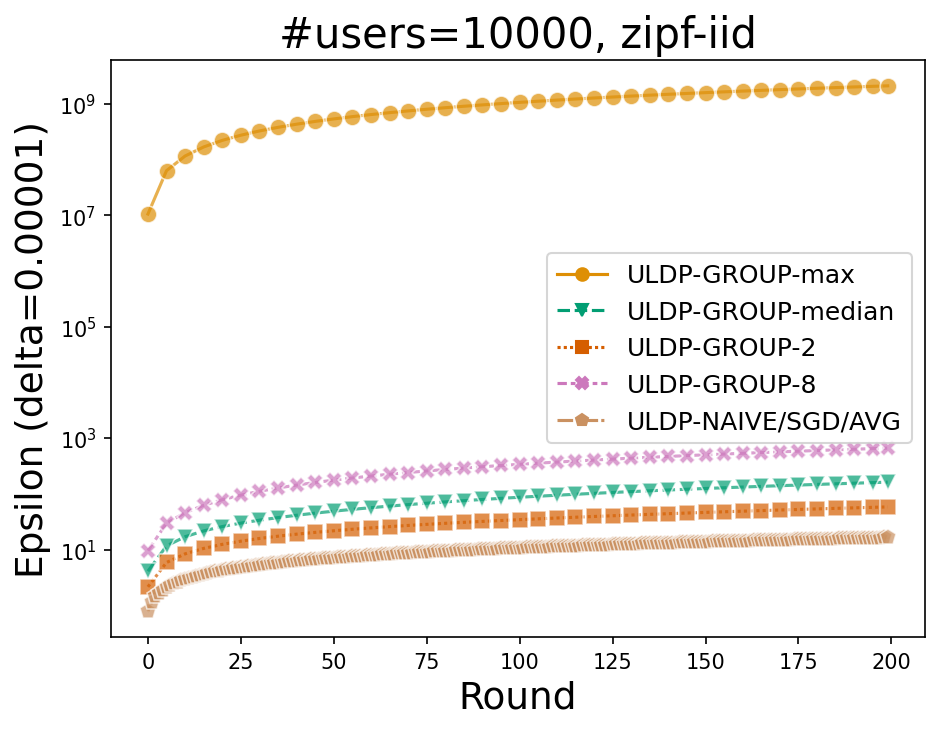}
        \caption{$n \approx 6$ ($|U|=10000$), zipf, iid.}
        \label{fig:privacy_utility_mnist_10000_zipf-iid}
    
        \includegraphics[width=0.32\linewidth]{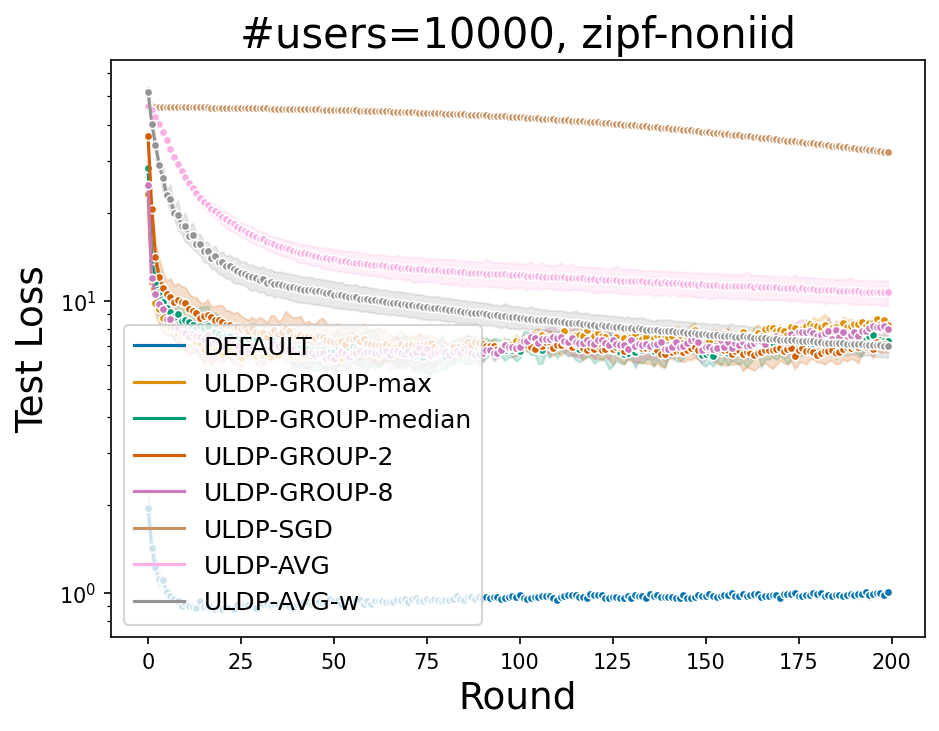}
        \hfill
        \includegraphics[width=0.32\linewidth]{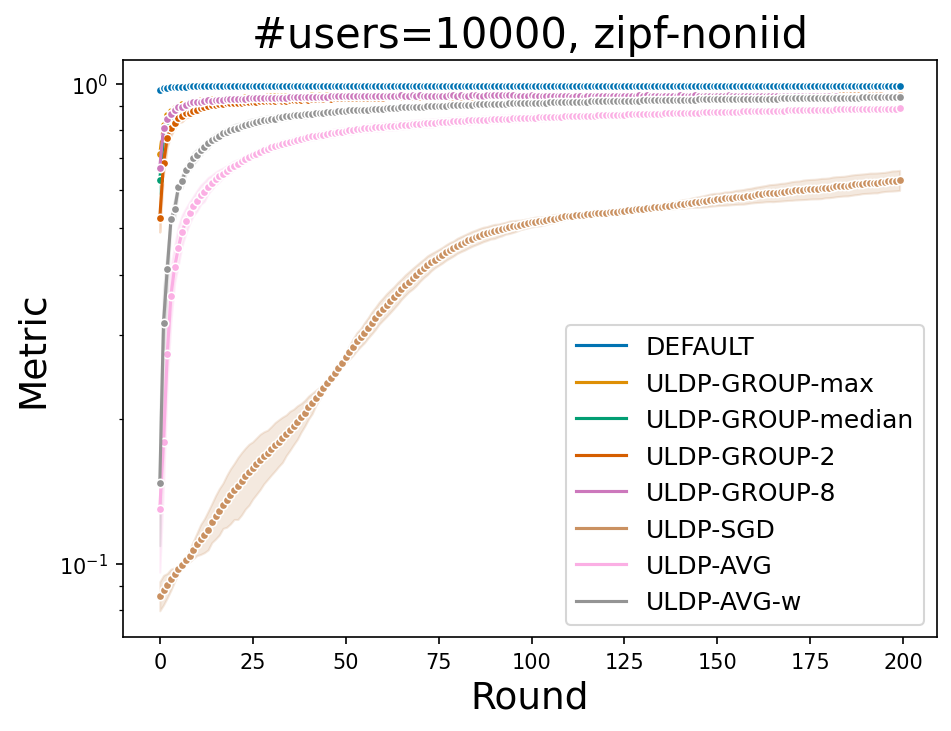}
        \hfill
        \includegraphics[width=0.32\linewidth]{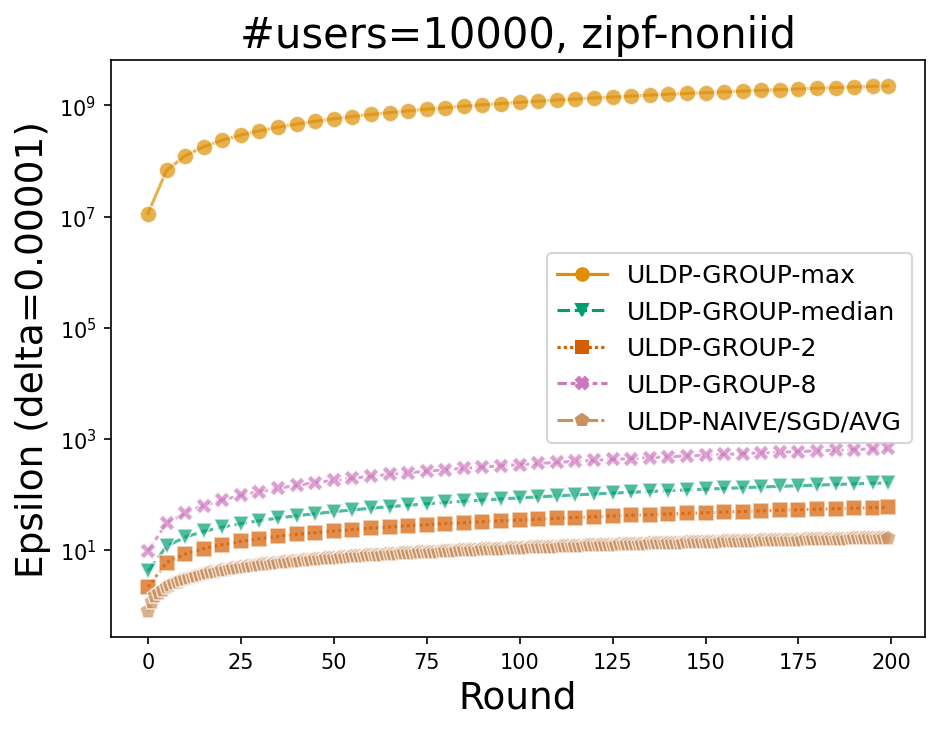}
        \caption{$n \approx 6$ ($|U|=10000$), zipf, non-iid.}
        \label{fig:privacy_utility_mnist_10000_zipf-noniid}
    \end{subfigure}
    \caption{Privacy-utility trade-offs on MNIST dataset: Test Loss (Left), Accuracy (Middle), Privacy (Right).}
    \label{fig:mnist}
\end{figure}

\begin{figure}
    \centering
        \begin{subfigure}{0.95\linewidth}
            \centering
            \includegraphics[width=0.48\linewidth]{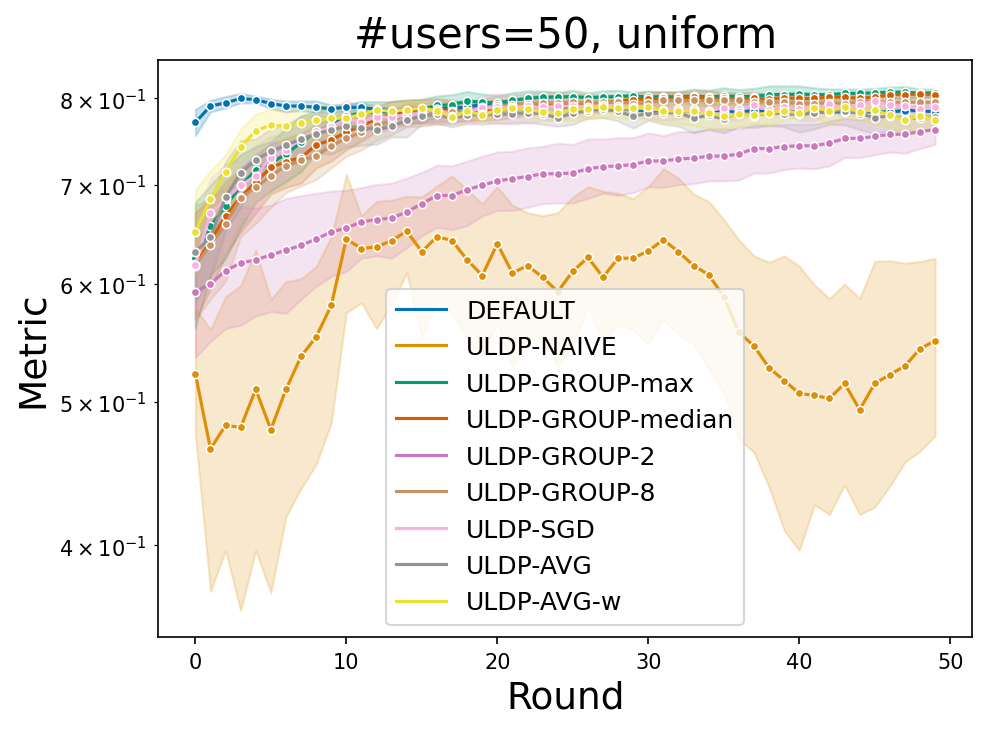}
            \hfill
            \includegraphics[width=0.48\linewidth]{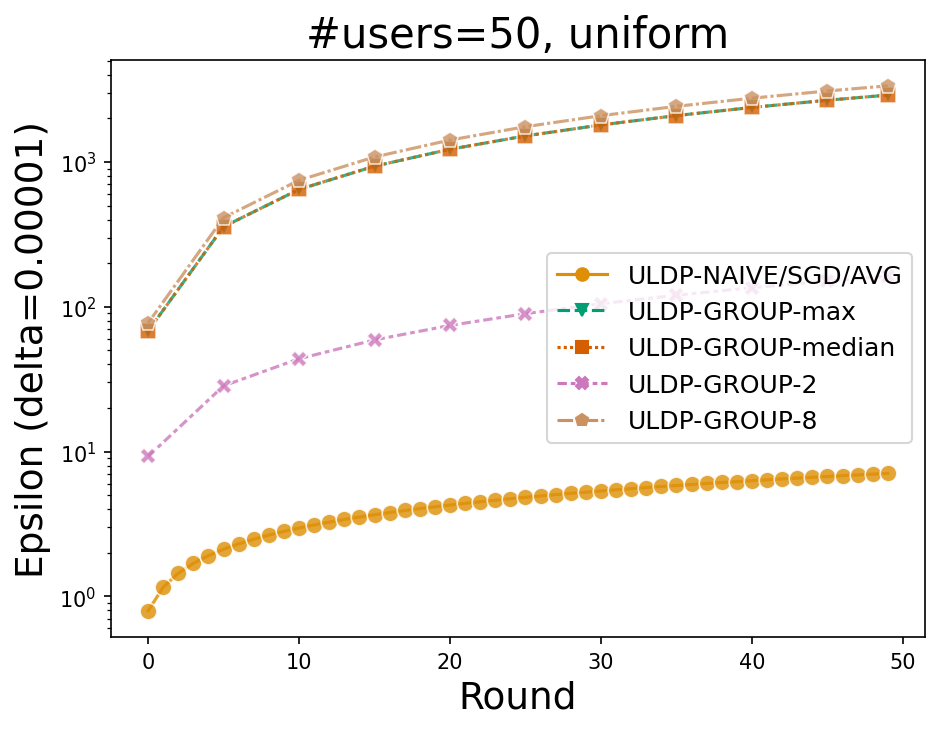}
            \caption{$n \approx 10$ ($|U|=50$), uniform.}
            \label{fig:heart_disease_a}
        
            \includegraphics[width=0.48\linewidth]{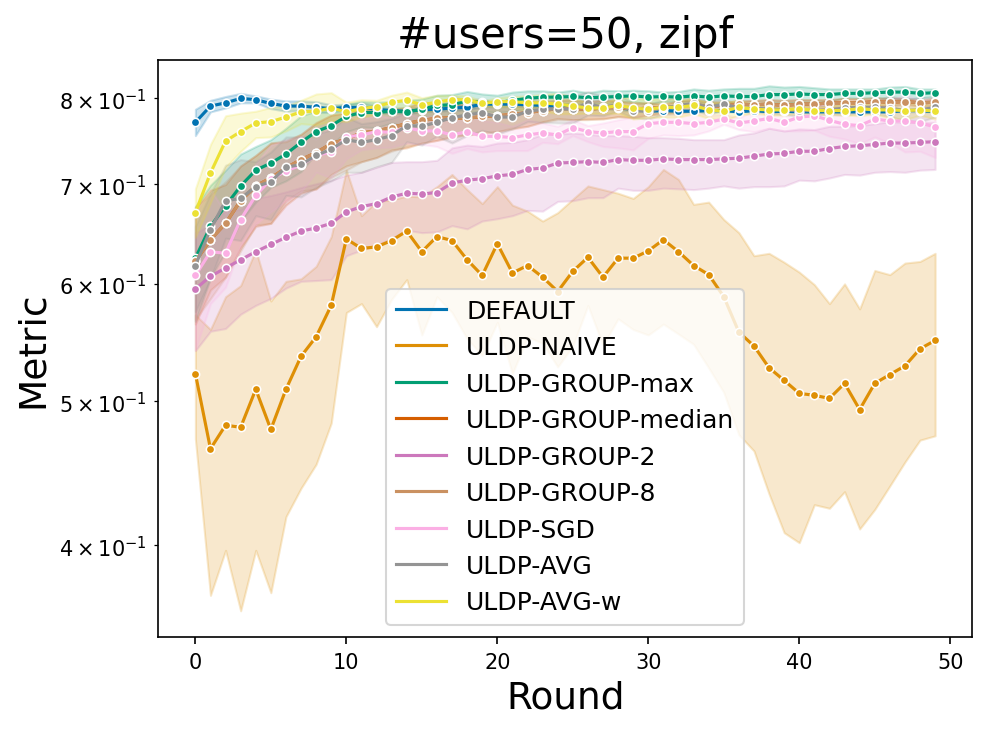}
            \hfill
            \includegraphics[width=0.48\linewidth]{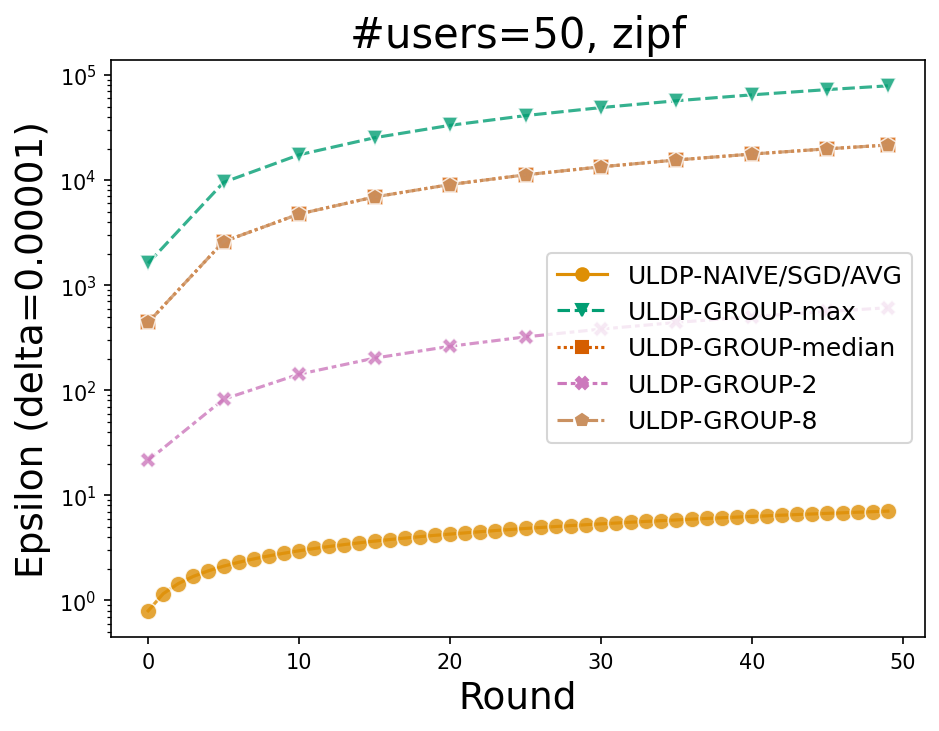}
            \caption{$n \approx 10$ ($|U|=50$), zipf.}
            \label{fig:heart_disease_b}
        
            \includegraphics[width=0.48\linewidth]{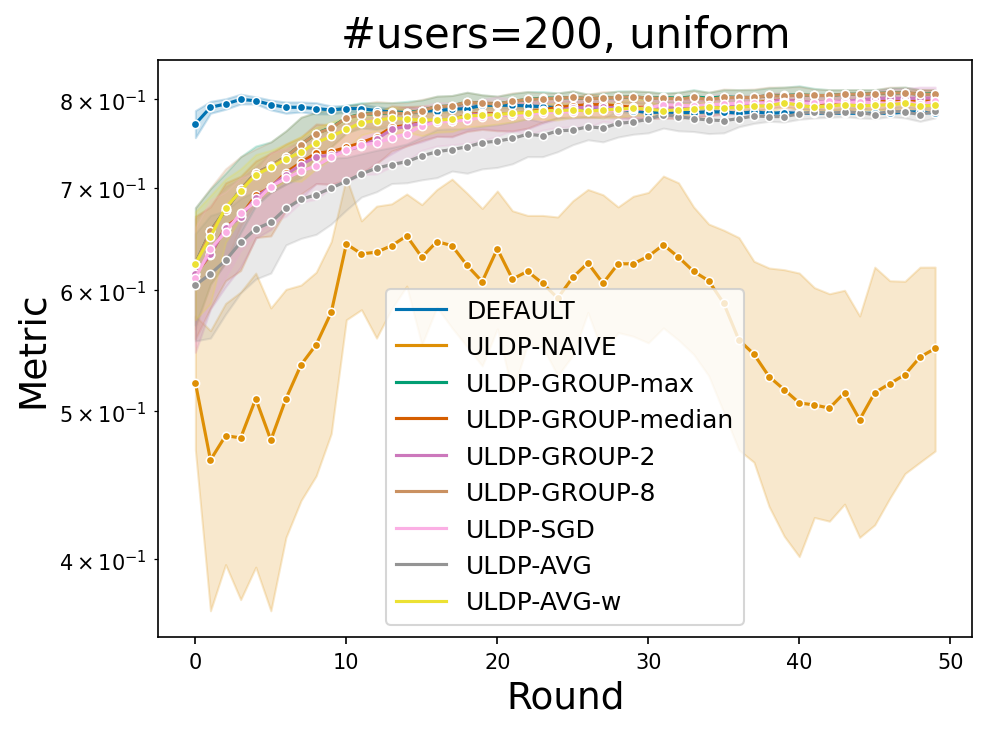}
            \hfill
            \includegraphics[width=0.48\linewidth]{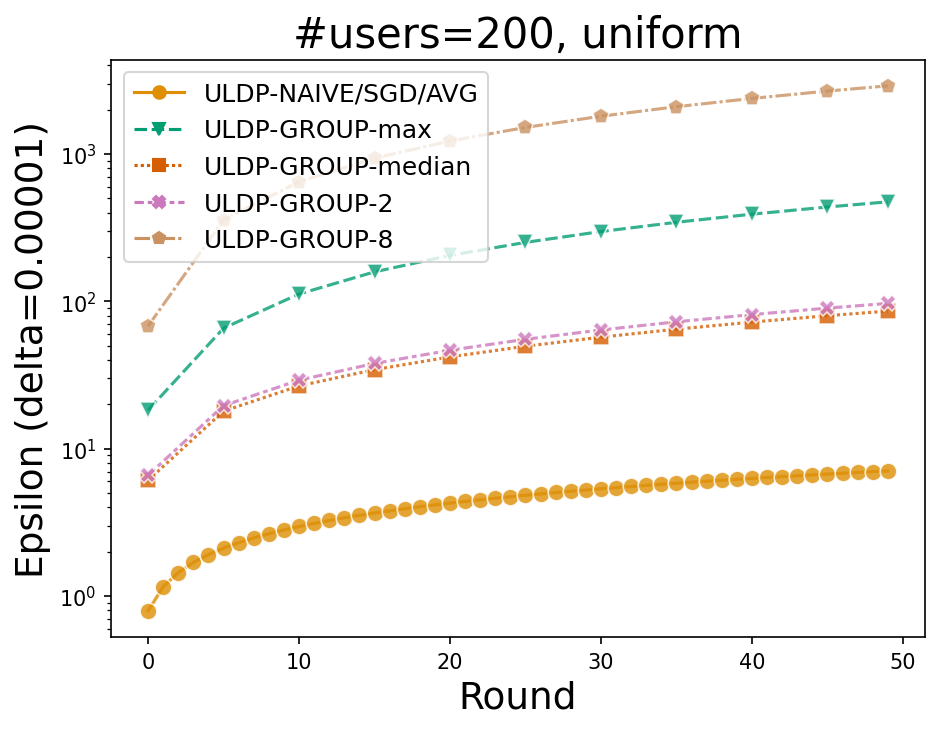}
            \caption{$n \approx 2.5$ ($|U|=200$), uniform.}
            \label{fig:heart_disease_c}
        
            \includegraphics[width=0.48\linewidth]{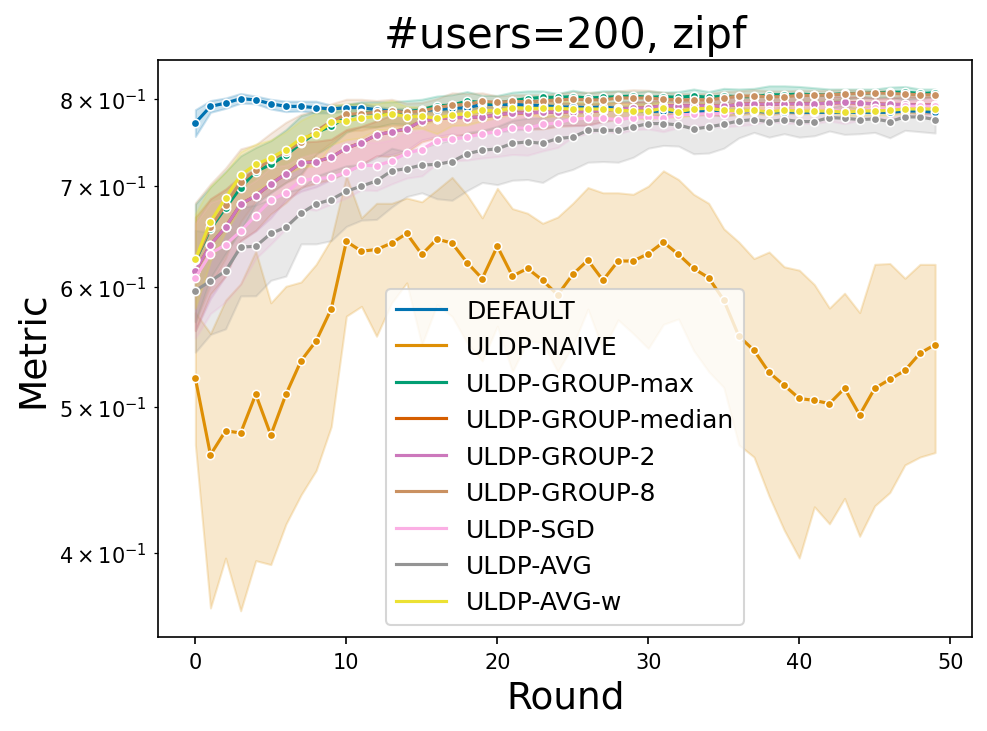}
            \hfill
            \includegraphics[width=0.48\linewidth]{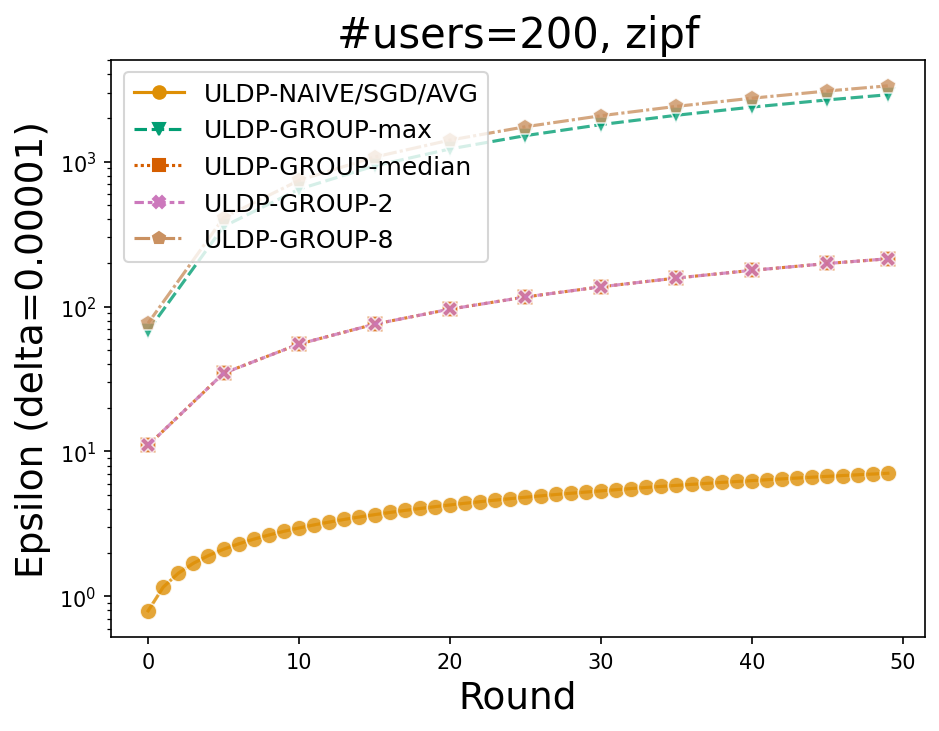}
            \caption{$n \approx 2.5$ ($|U|=200$), zipf.}
            \label{fig:heart_disease_d}
        \end{subfigure}
    \caption{HeartDisease.}
    \label{fig:heart_disease}
\end{figure}

\begin{figure}
    \centering
        \begin{subfigure}{0.95\linewidth}
            \centering
            \includegraphics[width=0.48\linewidth]{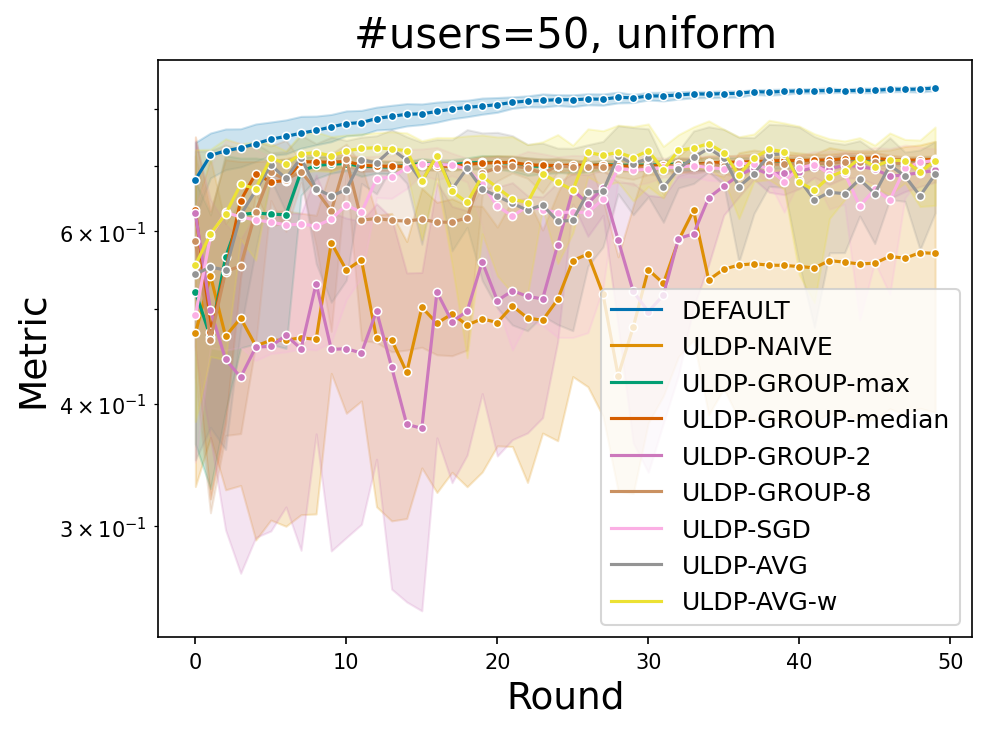}
            \hfill
            \includegraphics[width=0.48\linewidth]{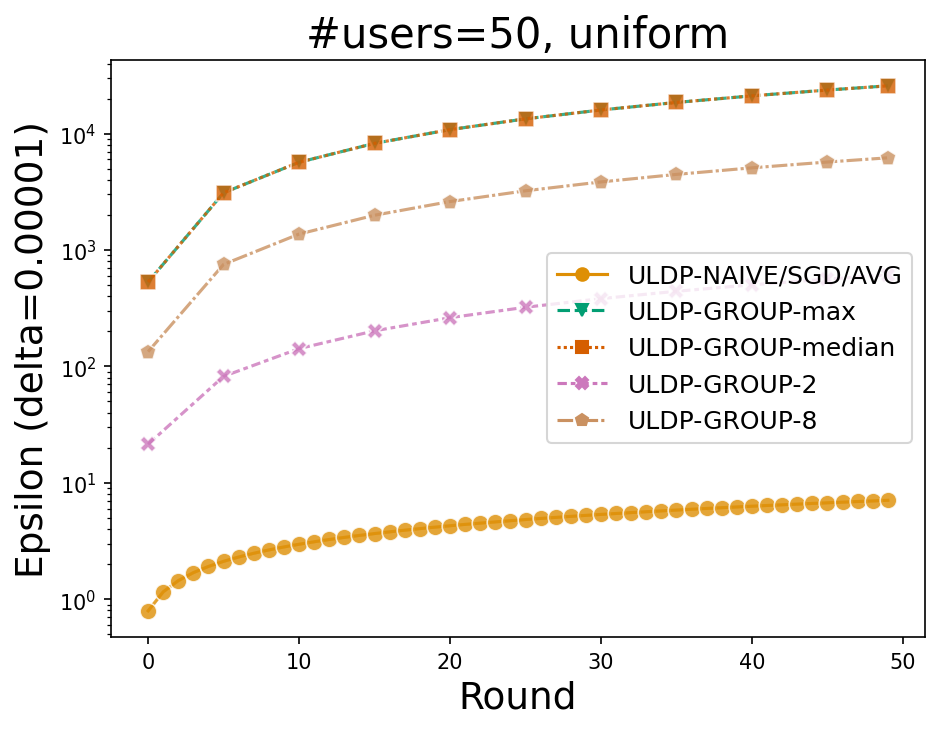}
            \caption{$n \approx 17$ ($|U|=50$) uniform.}
            \label{fig:tcga_brca_a}
        
            \includegraphics[width=0.48\linewidth]{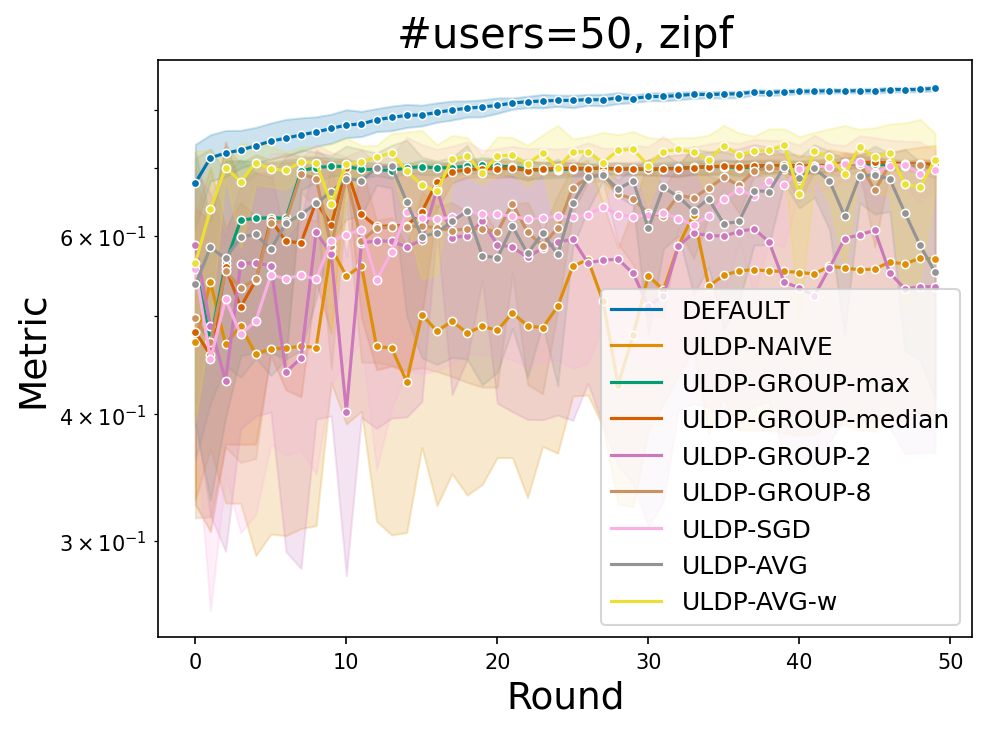}
            \hfill
            \includegraphics[width=0.48\linewidth]{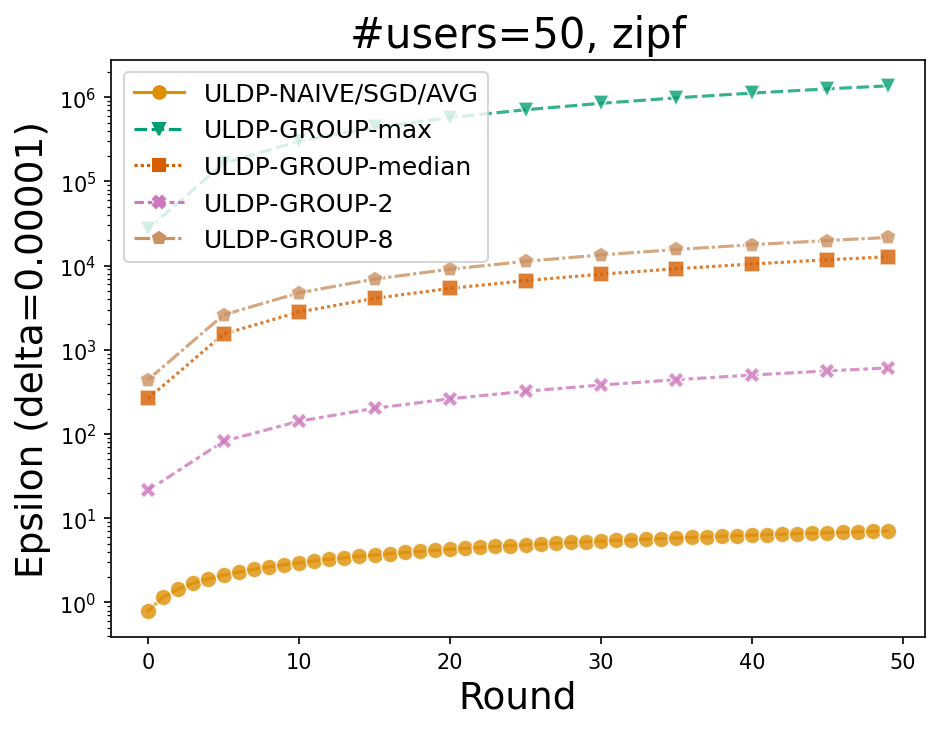}
            \caption{$n \approx 17$ ($|U|=50$), zipf.}
            \label{fig:tcga_brca_b}
        
            \includegraphics[width=0.48\linewidth]{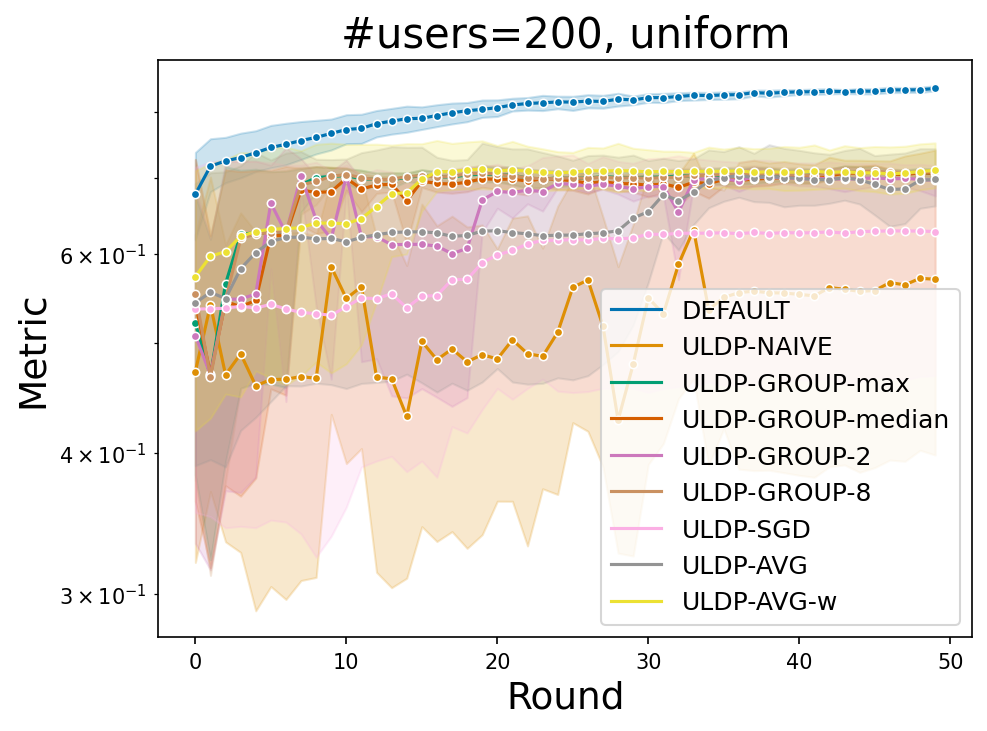}
            \hfill
            \includegraphics[width=0.48\linewidth]{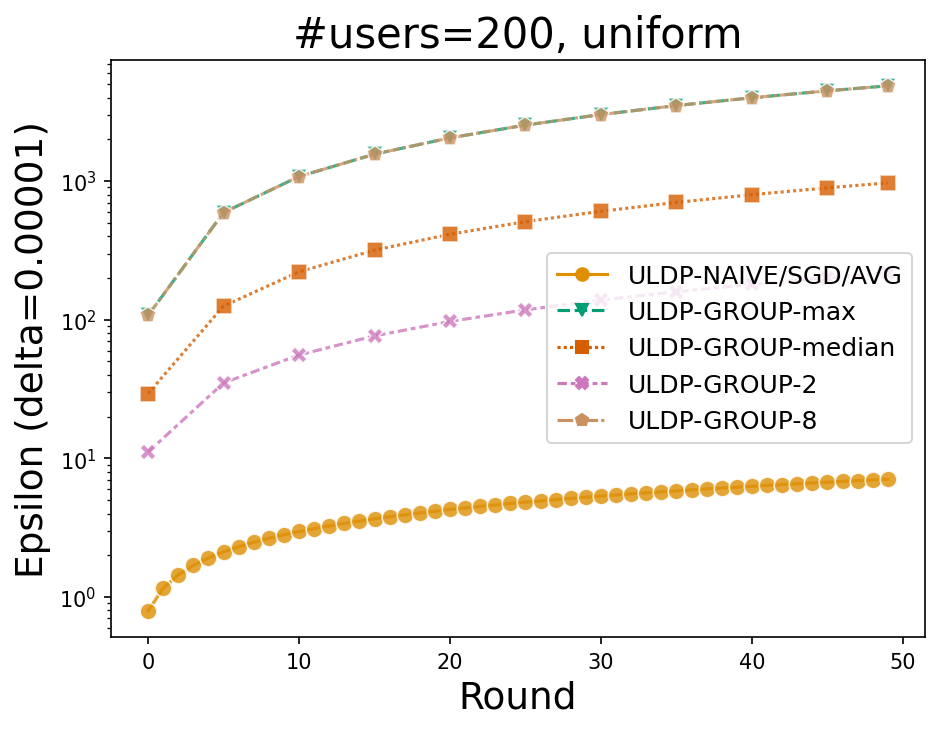}
            \caption{$n \approx 4$ ($|U|=200$), uniform.}
            \label{fig:tcga_brca_c}
        
            \includegraphics[width=0.48\linewidth]{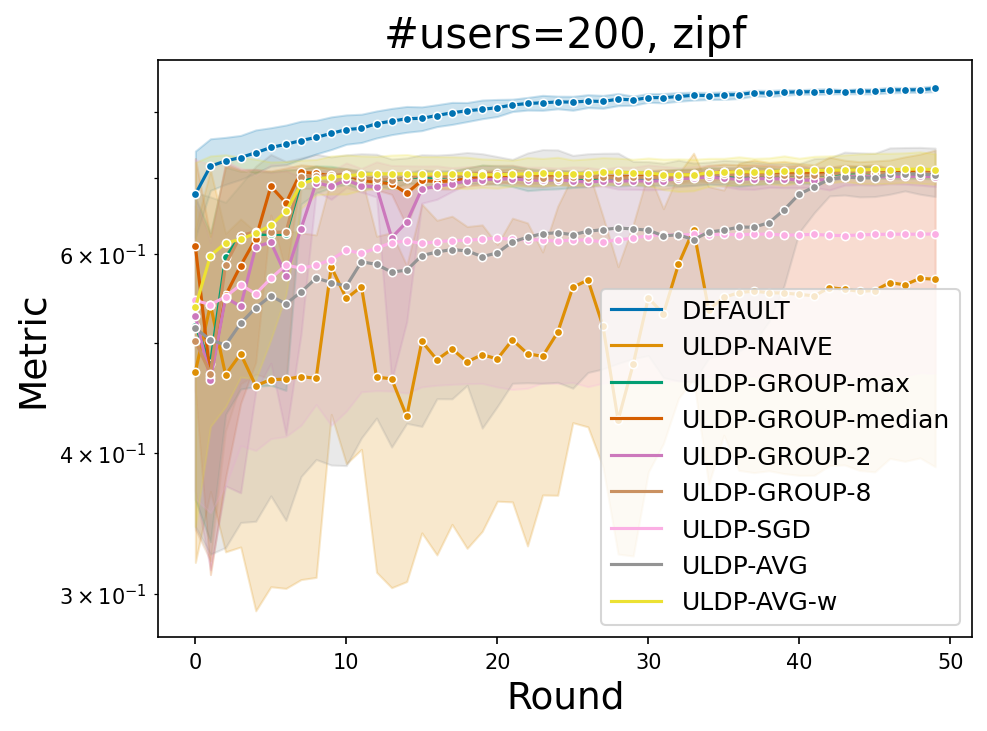}
            \hfill
            \includegraphics[width=0.48\linewidth]{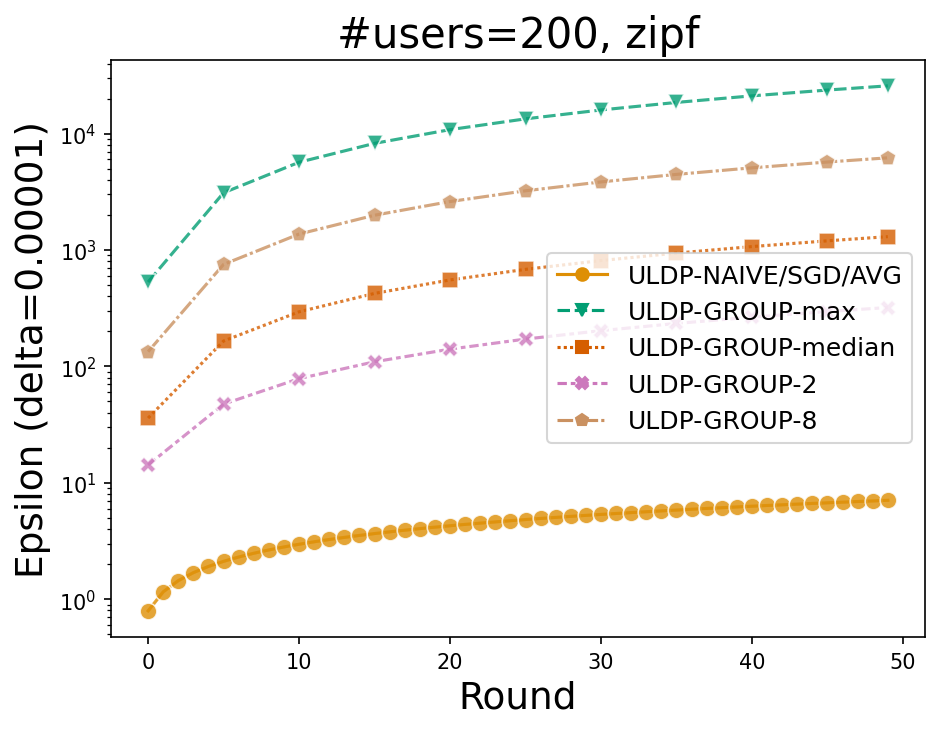}
            \caption{$n \approx 4$ ($|U|=200$), zipf.}
            \label{fig:tcga_brca_d}
        \end{subfigure}
    \caption{TcgaBrca.}
    \label{fig:tcga_brca}
\end{figure}

\textbf{Privacy-utility trade-offs under ULDP.}
Figures \ref{fig:creditcard} show the utility and privacy evaluation results on Creditcard.
The average number of records per user (denoted as $n$) in entire silos and the distribution changes for each figure.
All experiments used a fixed noise parameter $\sigma=5.0$ and $\delta=10^{-5}$, utility metrics (Accuracy for Creditcard) are displayed on the left side and accumulated privacy consumption $\epsilon$ for ULDP are depicted on the right side.
Note that the privacy bounds for ULDP-GROUP-$k$ are derived from the local DP-SGD and depend on not only the group size $k$ but also the size of the local training dataset.

Overall, the proposed method ULDP-AVG/SGD achieves competitive utility with fast convergence and high accuracy, while achieving considerably small privacy bounds, which means the significantly better privacy-utility trade-offs compared to baselines.
We observe that the baseline method, ULDP-NAIVE, has low accuracy and that ULDP-GROUP-$k$ requires much larger privacy budgets, which is consistent with the analysis on the conversion of group privacy described earlier.
The convergence speed of ULDP-AVG is faster than that of ULDP-SGD, which is the same as that of DP-FedAVG/SGD.
Nevertheless, there is still a gap between ULDP-AVG and the non-private method (DEFAULT) in terms of convergence speed and ultimately achievable accuracy, as a price for privacy.
Also, as shown in Figure \ref{fig:privacy_utility_creditcard_1000_uniform}, for small $n$ (i.e., a large number of users), ULDP-GROUP-max/median show higher accuracy than ULDP-AVG.
This is likely due to the overhead from finer datasets at user-level, which increases the bias compared to DP-FedAVG, as seen in the theoretical convergence analysis for ULDP-AVG.

Figure \ref{fig:heart_disease}, \ref{fig:tcga_brca}, and \ref{fig:mnist} show privacy-utility trade-offs on HeartDisease, TcgaBrca, and MNIST, respectively.
All experiments used a fixed noise parameter (noise multiplier) $\sigma=5.0$ and $\delta=10^{-5}$, utility metrics (Accuracy for HeartDisease and MNIST, C-index for TcgaBrca) are plotted on the left side, and accumulated privacy consumption $\epsilon$ for ULDP are plotted on the right side.
For clarity, the test loss is shown on the left-hand side for MNIST.
The average number of records per user (denoted as $n$) in entire silos and the distribution (uniform/zipf) changes for each figure.

In all datasets, consistently, ULDP-AVG is competitive in terms of utility, ULDP-AVG-w shows much faster convergence, and ULDP-SGD shows slower convergence.
ULDP-NAIVE achieves a low privacy bound; however, its utility is much lower than other methods.
ULDP-GROUP-$k$ show reasonably high utility, especially in settings where $n$ is small.
This is because the records to be removed due to the number of records per user being over group size $k$ is small.
However, the ULDP privacy bound ULDP-GROUP-$k$ achieves ends up being very large.
Note that the privacy bounds for ULDP-GROUP-$k$ are derived from the local DP-SGD and depend on not only the group size $k$ but also the size of the local training dataset.
The exceptions are cases where the local data set size is large and the number of records per user is very small as in Figures \ref{fig:privacy_utility_mnist_10000_uniform}, \ref{fig:privacy_utility_mnist_10000_zipf-iid}, and \ref{fig:privacy_utility_mnist_10000_zipf-noniid}.
In these cases, ULDP-GROUP-2 achieves a reasonably small privacy bound.
In other words, if the number of user records is fixed at one or two in the scenario, and the number of training records is large (it is advantageous for ULDP-GROUP because the record-level sub-sampling rate in DP-SGD becomes small), it could be better to use ULDP-GROUP.

\smallskip
\noindent
\textbf{Effect of non-IID data.}
The results for the MNIST, non-i.i.d, and $|U|=100$ case highlight a weak point of ULDP-AVG.
Note that non-i.i.d here is at user-level and DEFAULT and ULDP-GROUP are less affected by non-i.i.d because they train per silo rather than per user.
As Figure \ref{fig:privacy_utility_mnist_100_zipf-noniid} shows, the convergence of ULDP-AVG is worse compared to other results.
It suggests that ULDP-AVG may emphasize the bad effects of user-level non-i.i.d. distribution that were not an issue with normal cross-silo FL because the gradient is not computed at the user level as in ULDP-AVG.
On the other hand, this is less problematic when the number of users is large as shown in Figure \ref{fig:privacy_utility_mnist_10000_zipf-noniid}.
This is due to the relatively smaller effect of individual user overfitting caused by non-i.i.d. distribution as the number of users increases.
Another guideline could be to use existing non-iid cross-device FL methods \cite{ghosh2020efficient} adapted to the user-level in the silo (e.g., turning client-level clustering into user-level clustering, etc.). However, it should be non-trivial, since it should not violate ULDP.

\begin{figure}[t]
    \begin{subfigure}{0.32\linewidth}
    \includegraphics[width=\linewidth]{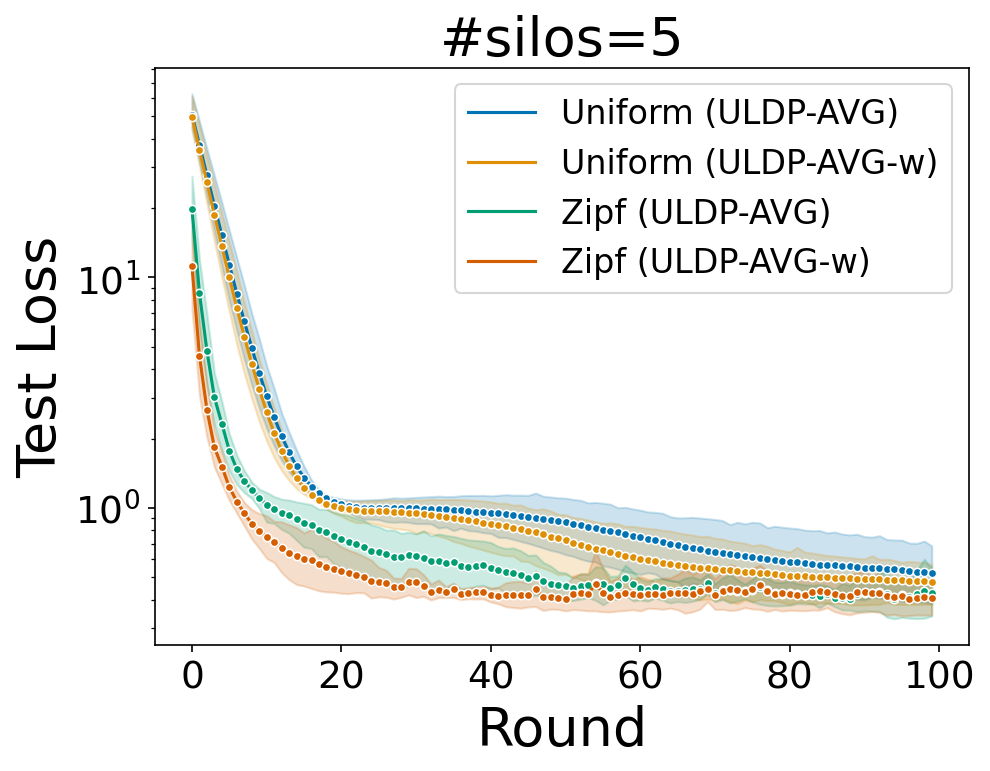}
    \end{subfigure}%
    \hfill
    \begin{subfigure}{0.32\linewidth}
    \includegraphics[width=\linewidth]{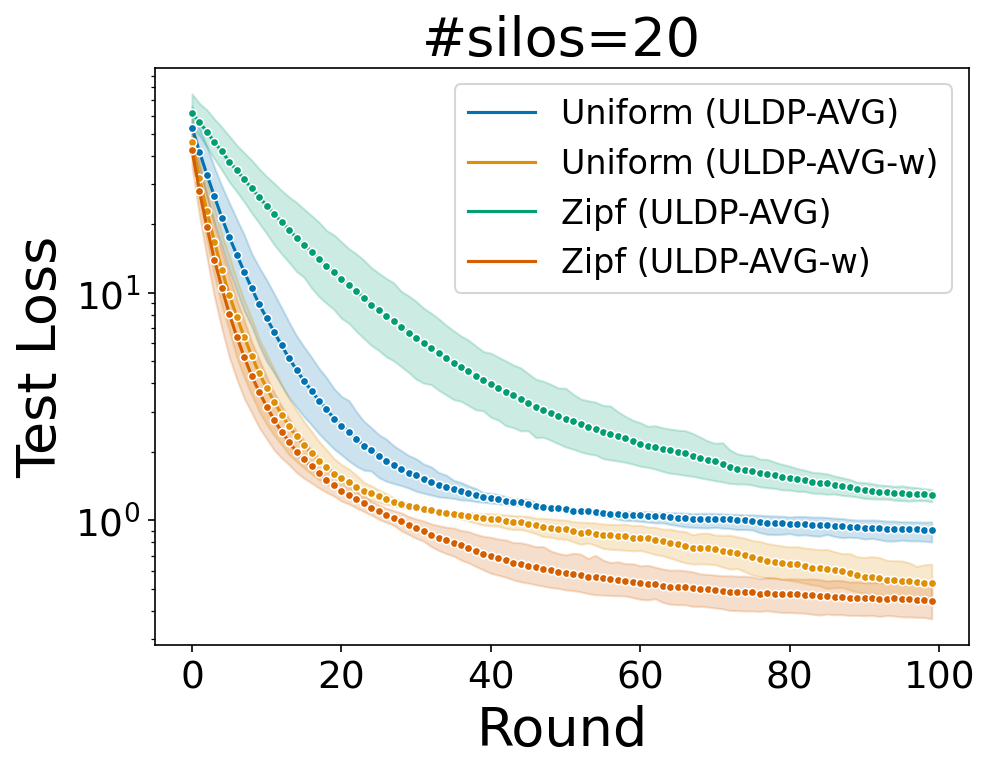}
    \end{subfigure}
    \hfill
    \begin{subfigure}{0.32\linewidth}
    \includegraphics[width=\linewidth]{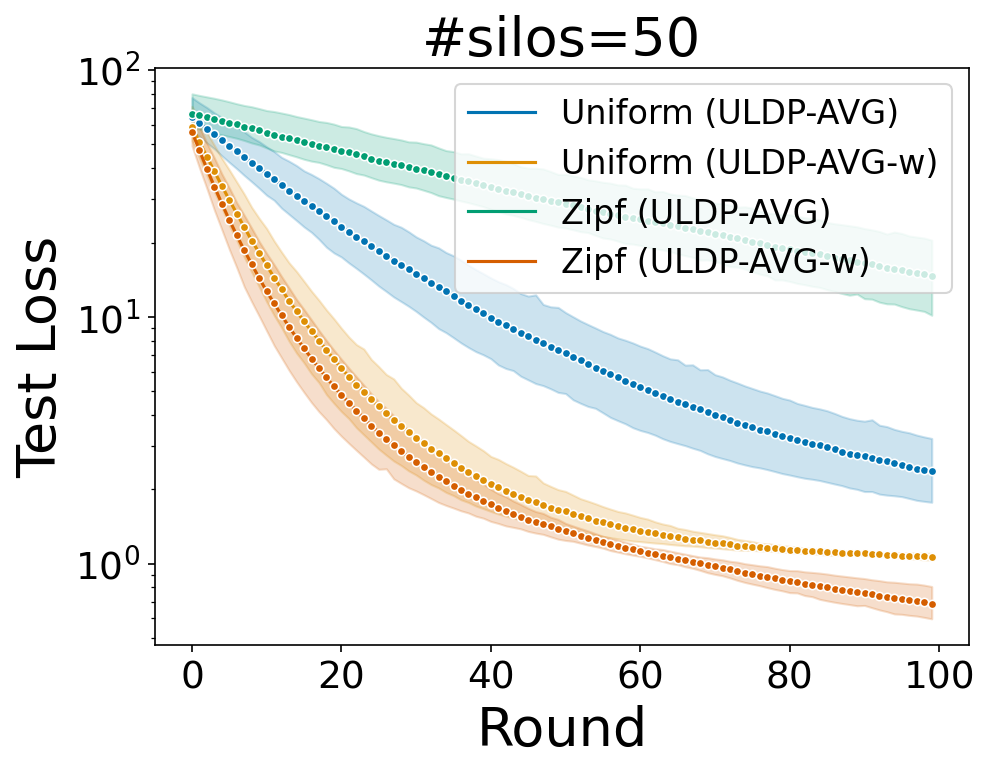}
    \end{subfigure}
    \caption{Test loss of Creditcard: Weighting method is effective, especially in skewed distribution in many silos.}
    \label{fig:testloss-optimal_weighting_creditcard}
\end{figure}


\smallskip
\noindent
\textbf{Effectiveness of enhanced weighting strategy.}
To highlight the effectiveness of the enhanced weighting strategy, Figure \ref{fig:testloss-optimal_weighting_creditcard} shows the test losses of the Creditcard dataset on different record distributions with ULDP-AVG and ULDP-AVG-w.
We present the results with various numbers of silos: 5, 20, and 50.
The need for the better weighting strategy is emphasized by the distribution of the records and the number of silos $|S|$.
When there are large skews in the user records, as in the Zipf distribution, giving equal weights (i.e., ULDP-AVG) results in inefficiency and opens up a large gap from ULDP-AVG-w.
This trend becomes even more significant as $|S|$ increases because all weights become smaller in ULDP-AVG.

\begin{figure}[t]
    \begin{subfigure}{0.98\linewidth}
        \begin{subfigure}{0.48\linewidth}
            \includegraphics[width=\linewidth]{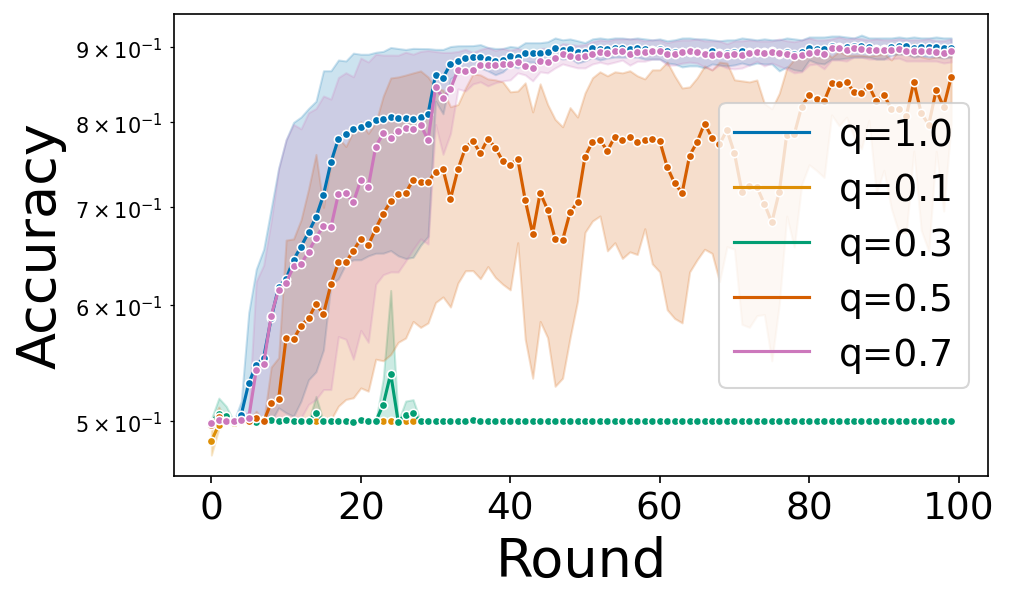}
        \end{subfigure}
        \hfill
        \begin{subfigure}{0.48\linewidth}
            \includegraphics[width=\linewidth]{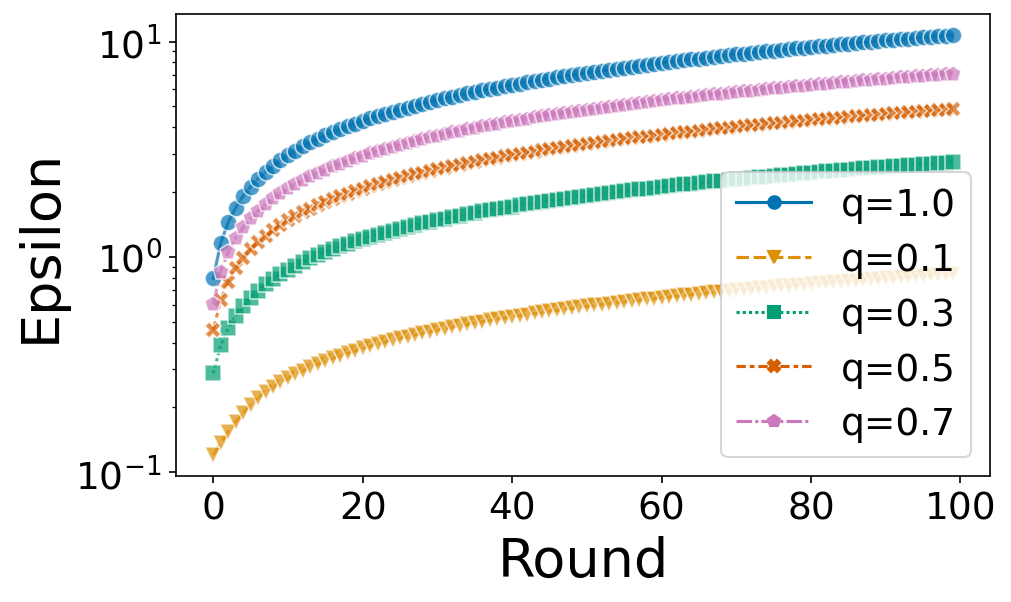}
        \end{subfigure}
    
        \caption{Creditcard.}
        \label{fig:user_level_subsampling_creditcard}
    
        \begin{subfigure}{0.48\linewidth}
            \includegraphics[width=\linewidth]{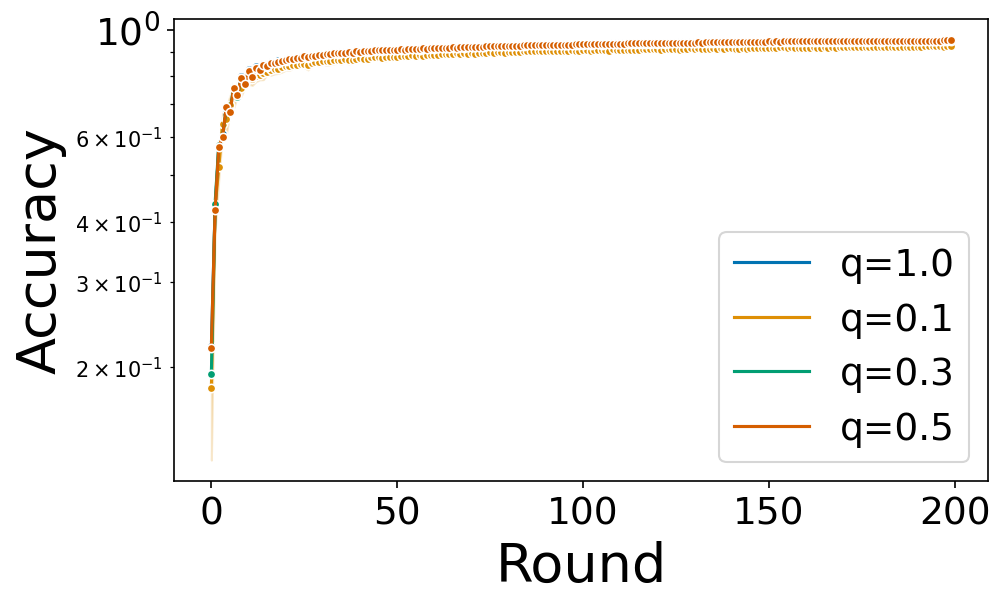}
        \end{subfigure}
        \hfill
        \begin{subfigure}{0.48\linewidth}
            \includegraphics[width=\linewidth]{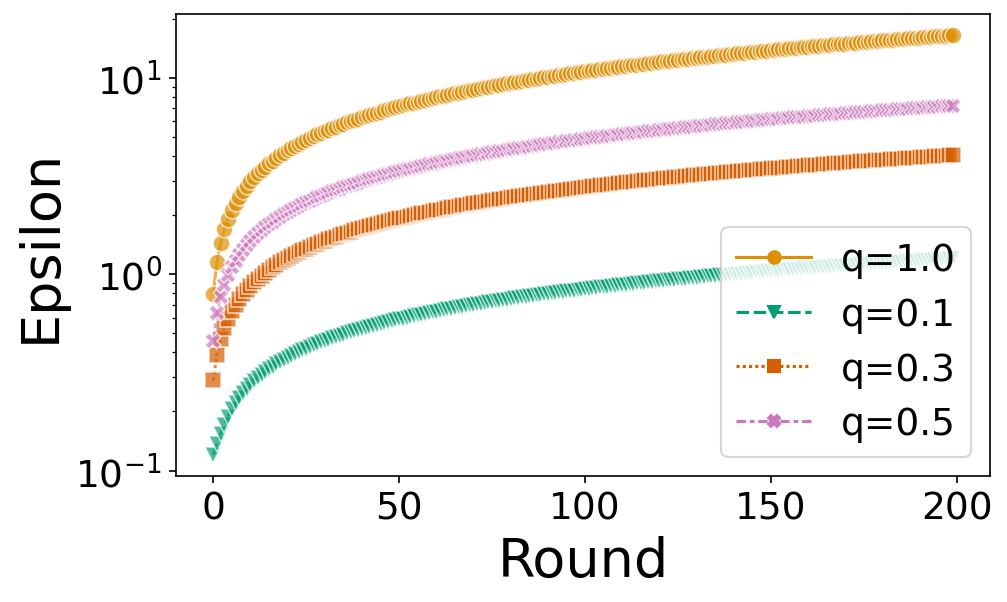}
        \end{subfigure}
        \caption{MNIST.}
        \label{fig:user_level_subsampling_mnist}
    \end{subfigure}
    \caption{User-level sub-sampling achieves a more competitive privacy-utility trade-off.}
\end{figure}

\smallskip
\noindent
\textbf{Effect of user-level sub-sampling.}
We evaluate the effect of user-level sub-sampling. 
Figure \ref{fig:user_level_subsampling_creditcard} illustrates how user-level sub-sampling affects the privacy-utility trade-offs on the Creditcard dataset with 1000 users.
We report the test accuracy and ULDP privacy bounds for various sampling rates $q=0.1, 0.3, 0.5, 0.7, 1.0$. 
Basically, a tighter privacy bound is obtained at the expense of utility.
As the results show, the degradation of utility due to sub-sampling could be acceptable to some extent (e.g., $q=0.7$) and there could be an optimal point for each setting.
Figure \ref{fig:user_level_subsampling_mnist} illustrates how user-level sub-sampling affects the privacy-utility trade-offs on MNIST with 10000 users, with sampling rates $q=0.1, 0.3, 0.5, 1.0$.
The results show that while privacy is greatly improved, there is less degradation in utility.
This is due to the fact that there are a sufficient number of users, i.e., 10000.
In the case of a larger user base, the effect of sub-sampling is greater and more important.

\begin{figure}[t]
  \begin{subfigure}{0.45\linewidth}
      \includegraphics[width=\linewidth]{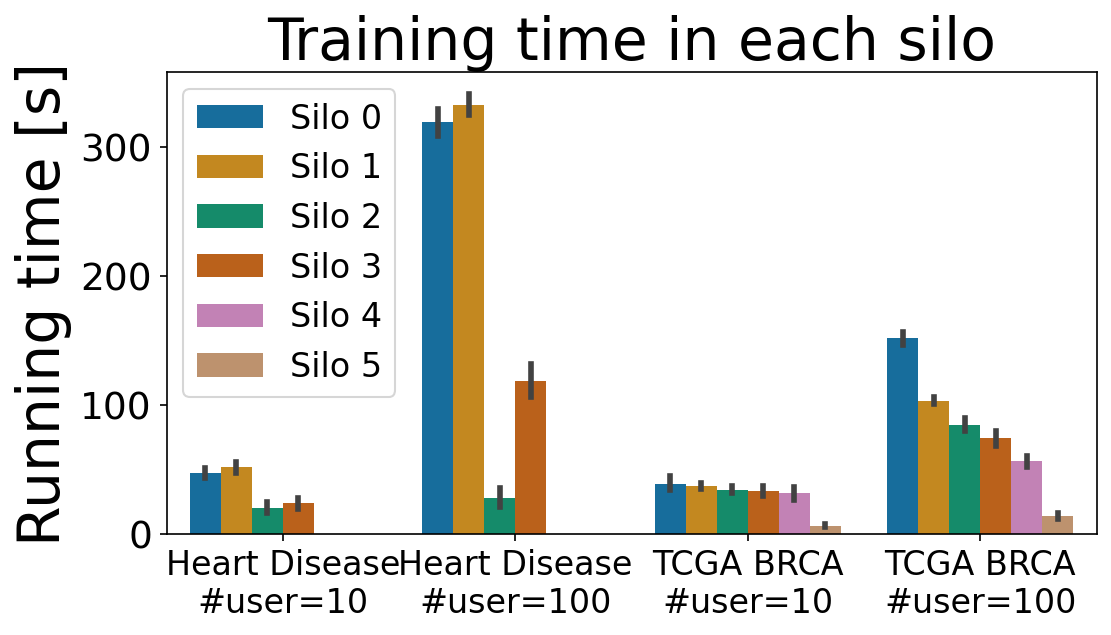}
  \end{subfigure}
  \hfill
  \begin{subfigure}{0.45\linewidth}
      \includegraphics[width=\linewidth]{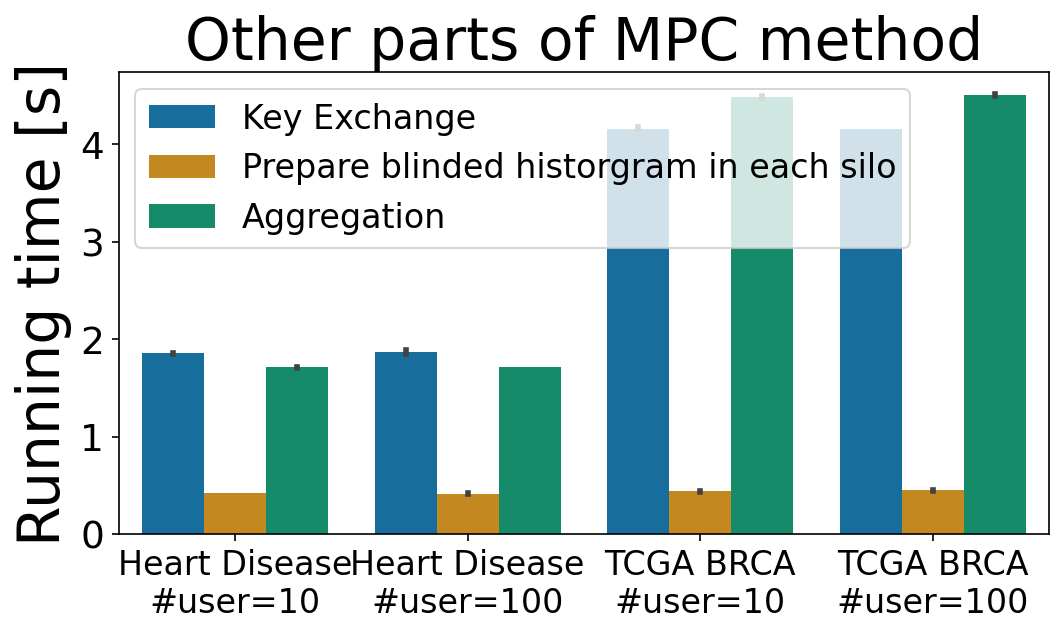}
  \end{subfigure}

  \caption{With a small model, the private weighting protocol has a practical execution time.}
  \label{fig:private_weighting_medical_runtime}
\end{figure}

\begin{figure}[t]
    \begin{subfigure}{\linewidth}
        \includegraphics[width=\linewidth]{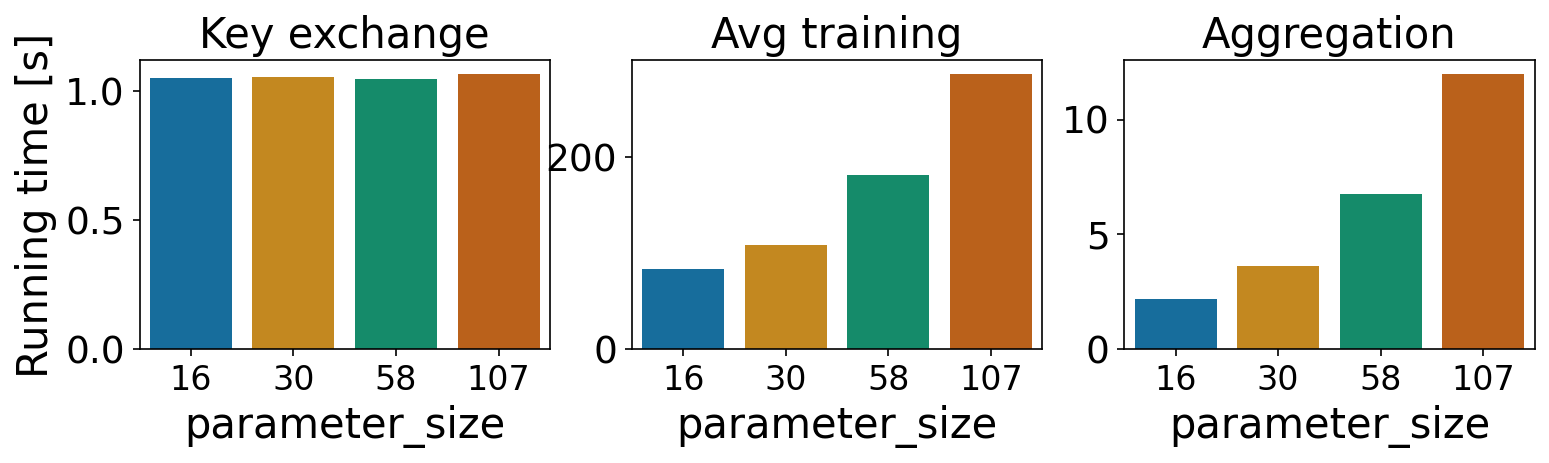}
        \includegraphics[width=\linewidth]{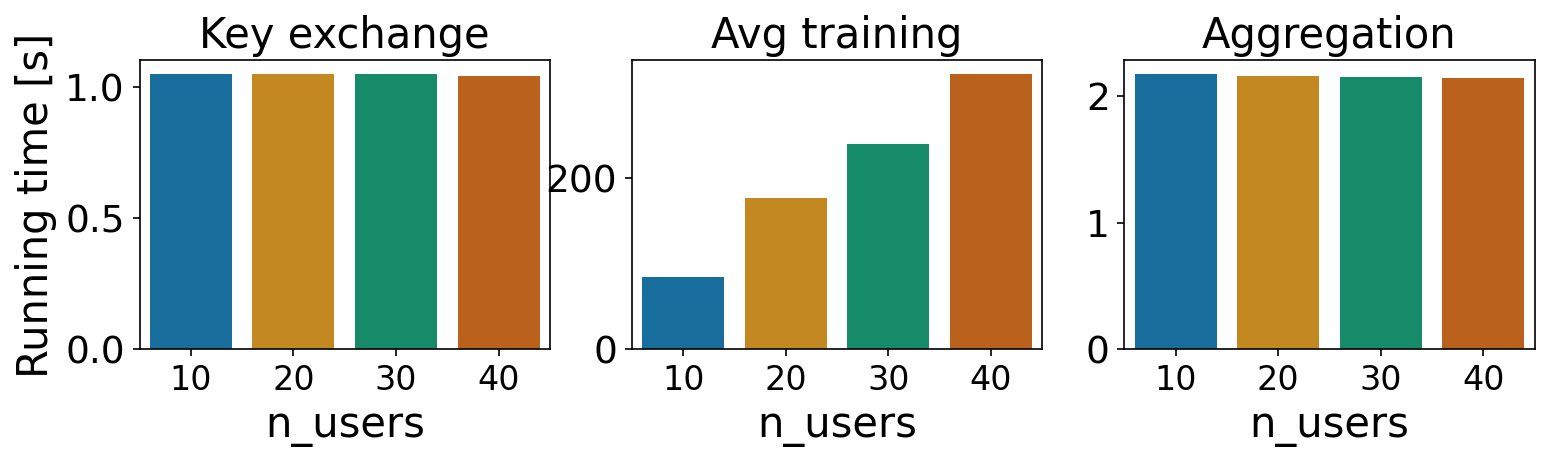}
  \end{subfigure}
    \caption{The dominant execution time grows linearly with parameter size (Top) and/or the number of users (Bottom).}
    \label{fig:private_weighting_artificial_runtime}
\end{figure}

\textbf{Overhead of private weighting protocol.}
We evaluate the overheads of execution time with the private weighting protocol.
Figure \ref{fig:private_weighting_medical_runtime} shows the execution times following HeartDisease and TcgaBrca with the number of users 10 and 100, respectively, with a skewed (zipf) distribution.
These two benchmark scenarios of cross-silo FL from \cite{ogier2022flamby} use small models.
The left figure shows the time required for local training in each silo, and the right figure shows the execution time for key exchange, preparation of blinded histograms, and aggregation.
As shown in the figure, the execution time of local training is dominant and it increases with a larger number of users.
Overall, it shows realistic execution times under these benchmark scenarios \cite{ogier2022flamby}.

Figure \ref{fig:private_weighting_artificial_runtime} shows the execution times of our proposed private weighting protocol (Protocol \ref{proto:1}) with an artificial dataset with 10000 samples and a model with 16 parameters, 20 users, and 3 silos as default.
These are on a considerably small scale.
The major time-consuming parts of the protocol are key exchange, training in each silo, and aggregation on the server.
The top three figures show the execution times on each part of the protocol with various parameter sizes from 16 to 107 and the bottom figures show the results on various number of users from 10 to 40.
The execution time of local training is averaged by silos.
The dominant part is the local training part, which is considered to be an overhead due to the computation with the Paillier encryption, which grows linearly with parameter size and/or the number of users.
The larger parameter size increases the aggregation time on the server as well.
Our implementation is based on the Python library \cite{python_paillier}, which itself could be made faster by software implementation or hardware accelerators \cite{yang2020fpga}.
However, it can be challenging to apply to larger models, such as DNNs, because the execution time increases linearly on a non-negligible scale with parameter size.
Therefore, extending the proposed method to deep models with millions of parameters is a future challenge.
It may be possible to replace such software-based encryption methods by using hardware-assisted Trusted Execution Environments, which have recently attracted attention in the FL field \cite{mo2021ppfl, kato2023olive}.

\section{Conclusion}
\label{sec:conclusion}
This study aimed to integrate user-level DP into FL, providing practical privacy guarantees for the trained model in general cross-silo scenarios.
We proposed the first cross-silo ULDP FL framework where a user can have multiple records across silos.
We designed an algorithm using per-user weighted clipping to directly satisfy ULDP instead of group-privacy.
In addition, we developed an enhanced weighting strategy that improves the utility of our proposed method and a novel protocol that performs it privately.
Finally, we demonstrated the effectiveness of the proposed method on several real-world datasets and showed that it performs significantly better than existing methods.
We also verified that our proposed private protocol works in realistic time in existing cross-silo FL benchmark scenarios.
For future work, research into the more scalable private protocols would be considered.
Also, it would be an independent and interesting direction to empirically compare the privacy protection of user/record-level DP in FL in terms of particular attack aspects such as user/record-level membership inference \cite{10.5555/3361338.3361469}.

\balance

\bibliographystyle{ACM-Reference-Format}
\bibliography{sample}

\clearpage
\appendix

\begin{table*}[t]
\centering
\resizebox{\textwidth}{!}{
\begin{tabular}{|p{0.3cm}|p{0.3cm}|p{2.4cm}|p{4cm}|p{4cm}|p{6cm}|}
\hline

\hline
\multicolumn{6}{|c|}{\textbf{CDP (Central DP) or DDP (Distributed DP)}} \\ \hline
\multicolumn{3}{|c|}{\textbf{Type of DP \& FL}} & \multicolumn{1}{c|}{\textbf{Description}} & \multicolumn{1}{c|}{\textbf{Protection Strength}} & \multicolumn{1}{c|}{\textbf{Trade-offs}} \\ \hline

\multicolumn{3}{|p{3cm}|}{Record-level DP (Centralized ML work \cite{abadi2016deep})} & Original definition of DP. & Basic: Privacy protection for each record. & (good) High utility. (bad) Insufficient privacy protection for users with multiple records, which often occurs in FL. In the case of CDP, trust in the server is necessary, but in many cases this is not problematic in FL because DDP \cite{kairouz2021distributed} is guaranteed with secure aggregation \cite{bonawitz2017practical}. \\ \hline

\quad \quad & \multicolumn{2}{|p{2.7cm}|}{Cross-silo FL \cite{10.14778/3503585.3503592, lowy2021private, lowy2023private, liu2022privacy}} & Independent DP protection applied specifically to silos, i.e., silo-specific DP. & Basic: Privacy protection for each record. & (good) DP algorithms performed in each silo are almost the same as in existing CML. Also, Different privacy budgets can be set for each silo. (bad) Same as record-level DP. \\ \hline

\multicolumn{3}{|p{3cm}|}{User-level DP (Centralized ML works \cite{levy2021learning, liu2020learning})} & Extends the definition of record-level DP by setting the neighboring database based on all records of any single user. & Strong: Privacy protection for user participants. & (good) Practical privacy protection for users with multiple records. (bad) Potential for greater utility loss compared to record-level DP. \\ \hline

\quad \quad & \multicolumn{2}{|p{2.7cm}|}{Cross-device FL \cite{geyer2017differentially, mcmahan2017learning, kairouz2021distributed, agarwal2021skellam, zhang2022understanding, girgis2021shuffled}} & Assume each device such as mobile and IoT has all of a single user's training data. & Strong: Privacy protection for user participants. & (good) Simple yet effective algorithm based on the assumption that each user has all data in at most one device. (bad) No major weakness. \\ \hline

\quad \quad & \quad \quad & \multicolumn{1}{|p{2.4cm}|}{Shuffling \cite{erlingsson2020encode, girgis2021shuffled}} & DDP is guaranteed by de-identifying the LDP-guaranteed data with a shuffler. & Strong: Privacy protection for user participants. Also satisfies Local DP. & (good) Compared to other Central DP variants, the shuffler reduces trust in the server. (bad) Utility does not exceed the user-level DP in the cross-device FL. \\ \hline

\quad \quad & \multicolumn{2}{|p{2.7cm}|}{\textcolor{red}{Cross-silo FL (\textbf{ours})}} & \textcolor{red}{Assume one user can have multiple records and the records exist across silos.} & \textcolor{red}{Strong: Privacy protection for user participants.} & \textcolor{red}{(good) It is possible to achieve a utility close to that of record-level DP with some algorithm designs. (bad) Need to design specific algorithms to satisfy user-level DP across silos.} \\ \hline

\multicolumn{3}{|p{3cm}|}{Group DP in cross-silo FL (e.g., Appendix of \cite{lowy2023private})} & Extends record-level DP to any group with up to $k$ records. & Strong: Privacy protection for any group participation. & (good) Any record-level DP algorithm can be applied without modification. (bad) Super-linear privacy bound degradation with respect to $k$ occurs when using approximate DP (i.e., non-zero $\delta$). Please see Section \ref{sec:preliminaries} in detail. \\ \hline 

\hline \hline

\multicolumn{6}{|c|}{\textbf{Local DP}} \\ \hline
\multicolumn{3}{|c|}{\textbf{Type of DP \& FL}} & \multicolumn{1}{c|}{\textbf{Description}} & \multicolumn{1}{c|}{\textbf{Protection Strength}} & \multicolumn{1}{c|}{\textbf{Trade-offs}} \\ \hline

\multicolumn{3}{|p{3cm}|}{Local DP in cross-device FL \cite{truex2020ldp}} & Privacy is ensured by local randomization before data aggregation. & Strongest: Privacy protection for user's input data itself. & (good) Unnecessary to trust server. (bad) Significant noise is necessary and hard to use for high-dimensional data. \\ \hline

\quad \quad & \multicolumn{2}{|p{2.7cm}|}{User-level DP \cite{wei2021user}} & A variant of Local DP in cross-device FL setting that can use a different privacy budget for each device. & Strongest: Privacy protection for user's input data itself. & (good) Different privacy budgets can be set for each user. (bad) Same as Local DP. \\ \hline

\multicolumn{3}{|p{3cm}|}{Local DP in cross-silo FL \cite{wang2022safeguarding}} & Assume that silos and users are separated and each user applies the randomization mechanism before data is collected in the silo. (\cite{wang2022safeguarding} assume implicitly that each user belongs to at most one silo.) & Strongest: Privacy protection for user's input data itself. & (good) Users do not have to trust the silos as well as the server. (bad) Same as Local DP. Strong assumption that LDP is guaranteed for the data before passing it to the client performing local training. \\ \hline

\end{tabular}
}
\caption{Comparison of Existing Differential Privacy Variants in Federated Learning.}
\label{table:dp_comparison}
\end{table*}

\section{DP Variants in FL}

Table \ref{table:dp_comparison} describes the comparison of existing DP variants in FL.
Initially, the table data can be divided into Central DP (CDP) or Distributed DP (DDP) (first to seventh rows) and Local DP (LDP) (eighth to tenth rows).
DDP is a variant of CDP that does not require trust in the server, and the privacy achievable is almost identical to that of CDP \cite{kairouz2021distributed}.
This is typically realized by hiding clients' raw data from the server through cryptographic techniques such as Secure Aggregation \cite{bonawitz2017practical} and distributed noise that regenerates Gaussian noise \cite{kairouz2021distributed}.

Within CDP, there are Record-level DP (first and second rows), User-level DP (third to sixth rows), and Group DP (seventh row).
Even within Record-level DP, cross-silo FL (second row) sometimes considers a specific privacy model known as silo-specific DP \cite{10.14778/3503585.3503592, lowy2021private, lowy2023private, liu2022privacy}.
Most research on User-level DP assumes cross-device FL (fourth and fifth rows) \cite{geyer2017differentially, mcmahan2017learning, kairouz2021distributed, agarwal2021skellam, zhang2022understanding, girgis2021shuffled}.
Shuffling DP-FL \cite{erlingsson2020encode, girgis2021shuffled} also aims to achieve DDP at the user level and is included in this category, yet it simultaneously guarantees a slight LDP (fifth row).
Our paper targets User-level DP in Cross-silo FL (sixth row). 
Among the studies on Record-level DP in cross-silo FL, some explore extending DP guarantees to multiple records using an extension to Group-DP (seventh row) (e.g., Appendix of \cite{lowy2023private}).

LDP primarily targets cross-device FL (eighth and ninth rows) (such as \cite{truex2020ldp}).
However, a different definition of User-level DP, albeit with the same name as our definition, has also been proposed (ninth row) \cite{wei2021user}.
Note that this is a variant of Local DP in a cross-device FL setting that can use a different privacy budget for each device.
Furthermore, LDP-compliant Cross-silo FL has also been proposed \cite{wang2022safeguarding}.
This comes with the special assumption that data can be collected from an independent \textit{user plane} at the data collection phase, ensuring LDP.

\section{Omitted Definition}

\begin{definition}[Sensitivity]
\label{def:sensitivity}
The sensitivity of a function $f$ for any two neighboring inputs $D, D' \in \mathcal{D}$ is:
\begin{equation}
\nonumber
    \Delta_{f} = \sup_{D, D' \in \mathcal{D}} \|f(D)-f(D')\|.
\end{equation}
where $\lVert\cdot\rVert$ is a norm function defined in $f$'s output domain.
\end{definition}

\noindent
We consider $\ell2$-norm ($\lVert\cdot\rVert_{2}$) as $\ell2$-sensitivity due to adding a Gaussian noise.

\section{Omitted Figures}
\begin{figure}[!t]

  \begin{subfigure}{\linewidth}
    \includegraphics[width=\linewidth]{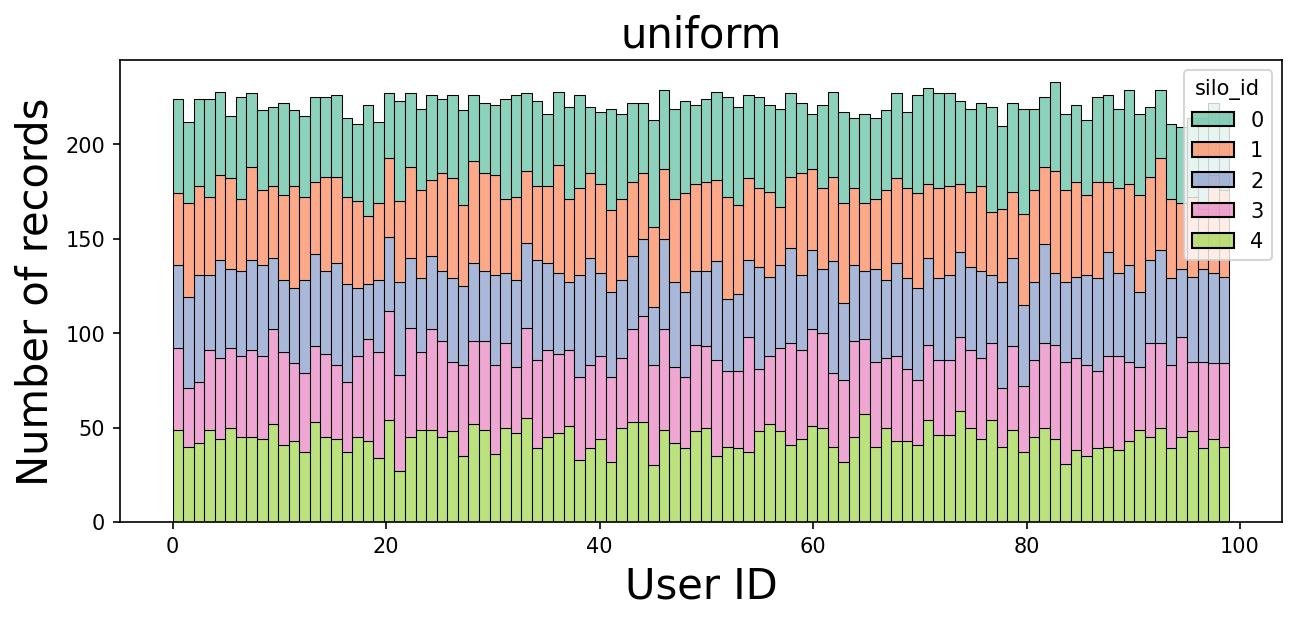}
    \caption{uniform.}
    \label{fig:record_distribution_creditcard_100_uniform}
  \end{subfigure}%

  \begin{subfigure}{\linewidth}
    \includegraphics[width=\linewidth]{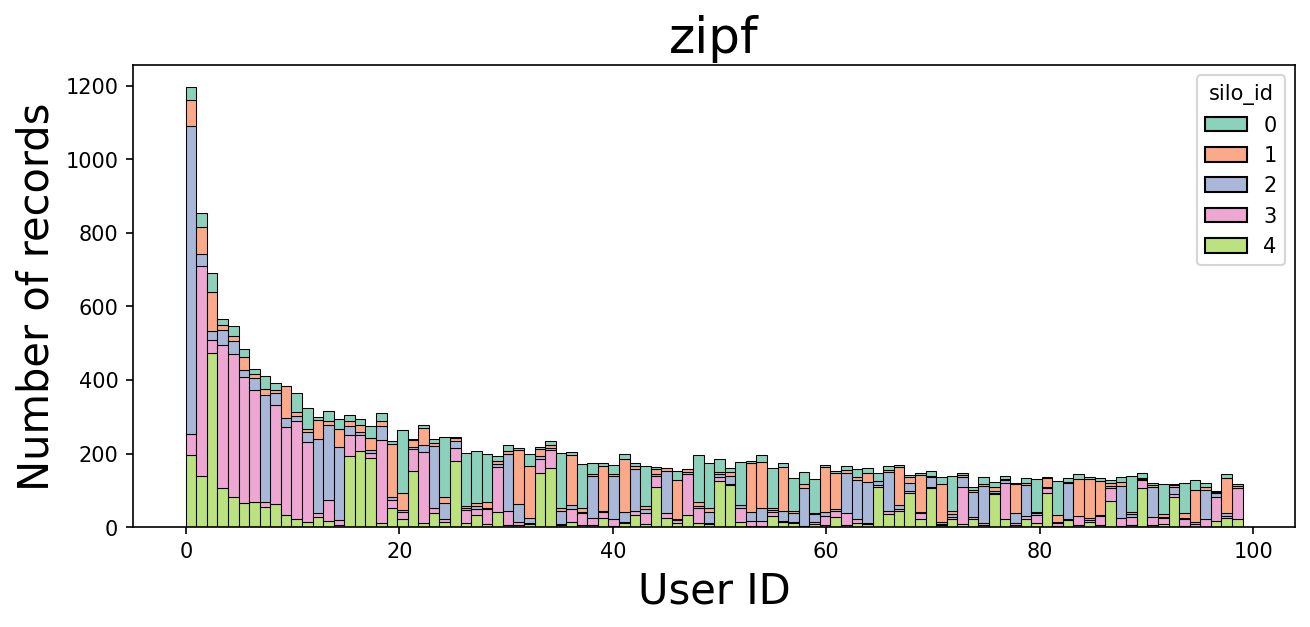}
    \caption{zipf.}
    \label{fig:record_distribution_creditcard_100_zipf}
  \end{subfigure}
  
  \caption{Example of record allocation on Creditcard.}
  \label{fig:record_distribution}
\end{figure}

Here we illustrate the examples of record allocation described in Section \ref{sec:exp:settings}.
For Creditcard and MNIST, the number of silos $|S|$ is fixed at 5.
We used 100, 1000 for Creditcard as $|U|$.
For example, when $|U|=100$ and $|S|=5$, the record distribution of Creditcard dataset is shown in Figure \ref{fig:record_distribution}.
The number of records is plotted for each user and color-coded for each silo.

\section{Proofs}

\subsection{Proof of Theorem \ref{theo:uldp_naive}}
At each round $t$, due to the clipping operation (Line 14 in Algorithm \ref{alg:naive}), for each silo $s \in S$, any user's contribution to $\Delta^{s}_t$ (Line 6) is limited to at most $C$ (regardless of the number of user records).
Since a single user may exist in any silo, they contribute at most $|S|C$ to $\sum_{s\in S}{\Delta^{s}_{t}}$, which means the user-level sensitivity is $|S|C$.
Therefore, when each silo adds Gaussian noise with variance $\sigma^2 C^2 |S|$, the aggregate $\sum_{s\in S}{\Delta^{s}_{t}}$ includes Gaussian noise with variance $\sigma^2 C^2 |S|^2$.
Then, by Lemma \ref{lemma:rdp_gaussian}, it satisfies $(\alpha, \frac{\alpha}{2\sigma^2})$-RDP for $\alpha > 1$. 
And after $T$ rounds, it satisfies $(\alpha, \frac{T\alpha}{2\sigma^2})$-RDP by Lemma \ref{lemma:rdp_composition}.
Finally, by Lemma \ref{lemma:rdp_conversion}, we obtain the final result.

\subsection{Proof of Theorem \ref{theo:uldp_group}}
In each silo $s \in S$, by performing DP-SGD with $Q$ epochs and $T$ rounds, we achieve record-level $(\alpha, \rho_s = \rho(\sigma, \gamma, QT))$-RDP.
The actual value of $\rho_s$ is calculated numerically as described in \cite{mironov2019r}.
RDP is known to satisfy parallel composition \cite{10.5555/3454287.3454599}.
That is, for disjoint databases $D_{1}$ and $D_{2}$, if $\mathcal{M}_1$ is $(\alpha, \rho_1)$-RDP and $\mathcal{M}_2$ is $(\alpha, \rho_2)$-RDP, their combined release $(\mathcal{M}_{1}(D_{1}), \mathcal{M}_2(D_{2}))$  satisfies $(\alpha, \mathrm{max}\{\rho_1, \rho_2\})$-RDP.
Since input databases $D_s$ for any silo $s \in S$ are disjoint from others, after $T$ rounds, the trained model satisfies $(\alpha, \rho=\max_{s\in S}\rho_s)$-RDP for the entire cross-silo database $D=D_{1} \oplus ... \oplus D_{|S|}$.
Then, by applying Lemma \ref{lemma:rdp_group_privacy} and Lemma \ref{lemma:rdp_conversion}, for any integer $k$ that is a power of 2 and any $\alpha \ge 2^{k+1}$, it also satisfies $(k, 3^{k}\rho + \log{((\frac{\alpha}{2^k}-1)/\frac{\alpha}{2^k})} - (\log{\delta} + \log{\frac{\alpha}{2^k}})/(\frac{\alpha}{2^k}-1), \delta)$-GDP.
By filtering with $\mathbf{B}$, we ensure that any user has at most $k$ records in $D$.
The final result is obtained by Proposition \ref{prop:gdp_uldp}.
(An alternative method to compute $\epsilon$ is to use Lemma \ref{lemma:normaldp_group_privacy} instead of Lemma \ref{lemma:rdp_group_privacy}.)

\subsection{Proof of Theorem \ref{theo:uldp_avg}}
For any round $t$, due to the clipping operation (Line 16), for each silo $s \in S$ any user's contribution to $\Delta^{s}_t$ (Line 6) is limited to at most $w_{s,u}C$ (regardless of the number of user records).
Therefore, for $\sum_{s\in S}{\Delta^{s}_{t}}$ (Line 6), any user's contribution is at most $\sum_{s\in S}w_{s,u}C = C$ since $\sum_{s \in S}{w_{s,u}}=1$.
That means the user-level sensitivity is just $C$.
Since each silo adds Gaussian noise with variance $\sigma^2 C^2/|S|$, $\sum_{s\in S}{\Delta^{s}_{t}}$ includes Gaussian noise with variance $\sigma^2 C^2$.
Then, by Lemma \ref{lemma:rdp_gaussian}, it satisfies $(\alpha, \frac{\alpha}{2\sigma^2})$-RDP for $\alpha > 1$ in a user-level manner.
And after $T$ rounds, it satisfies $(\alpha, \frac{T\alpha}{2\sigma^2})$-RDP by Lemma \ref{lemma:rdp_composition}.
Finally, by Lemma \ref{lemma:rdp_conversion}, we obtain the result.

\subsection{Proof of Theorem \ref{theo:correctness}}
Initially, with regards to secure aggregation, the additive pair-wise masks cancel out as shown in \cite{bonawitz2017practical}.
For the difference, silo must participate in any rounds in our cross-silo FL.
When collecting the histogram, additive masks are canceled out as follows:
{\small
    \begin{equation}
    \label{eq:hist_sec_agg}
        B(N_{u}) = \sum_{s\in S}{B(n_{s,u})} \nonumber 
        \quad + \sum_{s < s'}{\underbrace{(r^{u}_{s,s'} - r^{u}_{s',s})}_{\text{canceled out}}} - \sum_{s > s'}{\underbrace{(r^{u}_{s,s'} - r^{u}_{s',s})}_{\text{canceled out}}} \nonumber
    \end{equation}
}
\unskip The same applies to secure aggregation for model delta.
The mask is not directly added to the ciphertext (within the multiplication group of $n^2$); instead, scalar addition within $\mathbb{F}_{n}$ that takes advantage of the homomorphic property of the Paillier cryptosystem is utilized.
As secure aggregation doesn't result in any degradation of the aggregation outcome when all terms are within $\mathbb{F}_{n}$, our focus shifts to other components.
Note that there are errors due to the handling of fixed-point numbers on a finite field.
From the protocol description, $Enc_{\text{p}}(\Delta_{\text{sec}})$ is analyzed as follows:
{\small
    \begin{equation}
    \label{eq:enc_sec_delta}
    \begin{aligned}
        Enc_{\text{p}}(\Delta_{\text{sec}}) &= \sum_{s\in S}{Enc_{\text{p}}(\Delta^{s}_{t})} = \sum_{s\in S}{\sum_{u\in U}{Enc_{\text{p}}(\Tilde{\Delta}^{s,u}_{t})} + z'_{s}} \nonumber \\
        &= \sum_{s\in S}{\sum_{u\in U} E^{s,u}_{t} n_{s,u} r_{u} C_{\text{LCM}} Enc_{\text{p}}(B_{\text{inv}}(N_{u})) + Z^{s}_{t}C_{\text{LCM}}} \nonumber \\
        &\stackrel{(1)}{=} Enc_{\text{p}}(\sum_{s\in S}{(\sum_{u\in U}{(E^{s,u}_{t} n_{s,u} r_{u} C_{\text{LCM}} B_{\text{inv}}(N_{u}))}) + Z^{s}_{t}C_{\text{LCM}}}) \nonumber \\
        &= Enc_{\text{p}}(\sum_{s\in S}{(\sum_{u\in U}{(E^{s,u}_{t} n_{s,u} r_{u} C_{\text{LCM}} (r_{u}N_{u})^{-1}}) + Z^{s}_{t}C_{\text{LCM}}})) \nonumber \\
        &\stackrel{(2)}{=} Enc_{\text{p}}(\sum_{s\in S}{(\sum_{u\in U}{(E^{s,u}_{t} n_{s,u} C_{-N_u}}) + Z^{s}_{t}C_{\text{LCM}}})) \nonumber \\
    \end{aligned}
    \end{equation}
}
\unskip where $C_{-N_u}$ is the result of modular multiplication between $C_{\text{LCM}}$ and the modular multiplicative inverse of $N_u$, $E^{s,u}_{t}=\textsc{Encode}(\Tilde{\Delta}^{s,u}_{t}, P, n)$ and $Z^{s}_{t}=\textsc{Encode}(z^{s}_{t}, P, n)$.
Equation (1) is because all of $B_{\text{inv}}$, $E^{s,u}_{t}$, $n_{s,u}$, $r_{u}$ and $C_{\text{LCM}}$ $\in \mathbb{F}_n$.
At (2), the modulo inverse of $r_u$ is canceled out because $r_u \in \mathbb{F}_n$ and $r_u$ almost always has a modulo inverse.
However, this does not hold when $r_u$ and $n$ are not coprime.
Let $p$ and $q$ be two large primes used in the Paillier cryptosystem, such that $n = pq$, then the probability of a random $r_u \in \mathbb{F}_n$ and $n$ are not coprime is 
{
    \begin{equation}
        \frac{n - 1 - \phi(n)}{n-1} = \frac{n - 1 - (p-1)(q-1)}{n-1},
    \end{equation}
}
\unskip where $\phi$ is Euler's totient function.
This probability is negligibly small when n is sufficiently large.
In the case of user-level sub-sampling, $B_{\text{inv}}(N_{u})$ is set 0 and we see that only the model delta for user $u$ is all zeros, which produces exactly the same result as if the unselected users did not participate in the training round.
The important condition is that if $N_u \le N_{\text{max}}$ and $N_u$ is a factor of $C_{\text{LCM}}$, $C_{\text{LCM}}/N_u$ is always divisible on $\mathbb{Z}$ and the result is the same as on $\mathbb{F}_n$.
Also if $\sum_{s\in S}{(\sum_{u\in U}{(E^{s,u}_{t} n_{s,u} C_{-N_u}}) + Z^{s}_{t}C_{\text{LCM}}}) \in \mathbb{Z}$ is smaller than $n$, it yields the same results on $\mathbb{Z}$ and $\mathbb{F}_n$.
Hence, when these conditions are satisfied, decrypted value $\Delta_{\text{sec}} \in \mathbb{F}_n$ obtains the same result on $\Delta_{\text{sec}} \in \mathbb{Z}$.
After decryption, we consider $\Delta_{\text{sec}} \in 
\mathbb{R}$.
The final result is decoded by $\textsc{Decode}(\Delta_{\text{sec}}, P, C_{\text{LCM}}, n) \in \mathbb{R}$ as follows:
{\small
    \begin{equation}
    \label{eq:dec_sec_delta}
    \begin{aligned}
        \textsc{Decode}(\Delta_{\text{sec}}, P, C_{\text{LCM}}, n) = \sum_{s\in S}{(\sum_{u\in U}{(\frac{n_{s,u}}{N_u}\Tilde{\Delta}'_{s,u,t})} + Z'_{s,t}}) \nonumber
    \end{aligned}
    \end{equation}
}
\unskip where $\Tilde{\Delta}'_{s,u,t}$ is the same as the result of computing $\Tilde{\Delta}^{s,u}_{t}$ in fixed-point with precision $P$, and $Z'_{s,t}$ is the same as $z^{s}_{t}$.

Therefore, the correctness of these calculations is satisfied when two conditions (1) $N_u \le N_{\text{max}}$ and \\
(2) $\sum_{s\in S}{(\sum_{u\in U}{(E^{s,u}_{t} n_{s,u} C_{-N_u}}) + Z^{s}_{t}C_{\text{LCM}}}) < n$, are satisfied.
To satisfy (1), $N_{\text{max}}$ must be sufficiently large, and a larger $N_{\text{max}}$ leads to a larger $C_{\text{LCM}}$.
Hence, when we take $n$ is large, these conditions can be satisfied unless the parameters or noise take on unrealistically large values.

For example, suppose the range of the noise and aggregated model parameters is $[-10^{10}, 10^{10}]$, $P=10^{-10}$, $N_{\text{max}}=2000$ and $\lambda$ is 3072-bit security, we have $n > 10^{924}$ by Paillier cryptosystem, $E^{s,u}_{t} < 10^{20}$ , $Z^{s}_{t} < 10^{20}$ and $C_{\text{LCM}} < 10^{867}$.
Then, \\ 
$\sum_{s\in S}{(\sum_{u\in U}{(E^{s,u}_{t} n_{s,u} C_{-N_u}}) + Z^{s}_{t}C_{\text{LCM}}}) < 10^{888} < n$ and we satisfy the condition (2).
$C_{\text{LCM}}$ grows exponentially with respect to $N_{\text{max}}$. 
One possible solution for this is that we can make it very small by limiting the number of records per user to several values. 
For example, $\{10^{1}, 10^{2}, 10^{3}, 10^{4}\}$ then $C_{\text{LCM}}=10^{4}$.

\subsection{Proof of Theorem \ref{theo:privacy}}

Our approach relies on a private weighted sum aggregation technique employing the Paillier cryptosystem \cite{paillier1999public, liu2022privacy} and secure aggregation \cite{bonawitz2017practical}.
Formal security arguments for these methodologies are available in their respective sources.
The protocol is fully compatible with these works because all data exchanged is handled on $\mathbb{F}_{n}$ including random masks for secure aggregation.

A different view of the server from these basic methods is $B(N_{u})$ for all $u$, which is multiplicatively blinded aggregated histogram.
Since $B(n_{s,u})$ for $s$ and $u$ is securely hidden by secure aggregation, only $B(N_{u})$ is the meaningful server view.
$B(N_{u})$ is $r_uN_{u}\, (\text{mod}\, n)$ and this is randomly distributed on $\mathbb{F}_{n}$ if $r_u$ is uniformly distributed on $\mathbb{F}_{n}$.
Also, the inverse of $B_{\text{inv}}(N_{u})$ is uniformly distributed as well.
This is because the multiplication operation in a finite field is closed and bijective.
Therefore, it is information-theoretically indistinguishable and private.
Such multiplicative blinding has been also used in \cite{damgaard2007efficient}.
The view of the silos is the same as that in \cite{liu2022privacy}, even though the contents of the weights are sensitive, and privacy is protected by Paillier encryption.
Note that the security of the initial DH key exchange and the security of the Paillier cryptosystem follow the security parameter $\lambda$, which is an input to the protocol.

\newpage
\onecolumn

\noindent
\section{Convergence analysis of ULDP-AVG}
\label{appendix:convergence}

In this section, we theoretically analyze ULDP-AVG and give a convergence analysis to compare it with existing methods.
To this end, we use the following assumptions.
These assumptions, commonly used in prior researches \cite{yang2021achieving, zhang2022understanding, 50448}, facilitate the derivation of convergence properties from the boundedness of derivatives, including bounded Lipschitz constants.
In particular, the second assumption, $\sigma_g$, quantifies the heterogeneity of non-i.i.d. data between silos in FL and is used in many previous studies.
$\sigma_g=0$ corresponds to the i.i.d. setting.
For each $s \in S$ and $u \in U$, we assume access to an unbiased stochastic gradient $g^{s,u}_{t,q}$ of the true local gradient $\nabla f_{s,u}(x)$ for $s$ and $u$.

\begin{assumption}[Lipschitz Gradient]
\label{ass:lipschitz}
    The function $f_{s,u}$ is L-smooth for all silo $s \in S$ and user $u \in U$, i.e., $\lVert \nabla f_{s,u}(x) - \nabla f_{s,u}(y) \rVert \le L \lVert x-y \rVert$, for all $x,y \in \mathbb{R}^d$.
\end{assumption}

\begin{assumption}[Bounded Variances]
\label{ass:local_and_global_variance}
    The function $f_{s,u}$ is $\sigma_l$-locally-bounded, i.e., the variance of each local gradient estimator is bounded as $\mathbb{E}[\lVert g^{s,u}_{t,q} - \nabla f_{s,u}(x^{s,u}_{t,q})\rVert^2] \le \sigma_l^2$ for all $s$, $u$ and $t, q$. And the functions are $\sigma_g$-globally-bounded, i.e., the global gradient variance is bounded as $\lVert\nabla f_{s,u}(x_t) - f(x_t)\rVert^2 \le \sigma_g^2$ for all $s, u, t$.
\end{assumption}

\begin{assumption}[Globally Bounded Gradients]
\label{ass:global_bounded_grad}
    For all $s, u, t, q$, gradient is $G$-bounded, i.e., $\lVert g^{s,u}_{t,q}\rVert \le G$.
\end{assumption}

\noindent
Then, we derive the following convergence theorem, enabling comparison with existing methods such as FedAVG and DP-FedAVG.
\begin{theorem}[Convergence analysis on ULDP-AVG]
\label{theorem:uldp_conv}
For ULDP-AVG, with assumptions \ref{ass:lipschitz}, \ref{ass:local_and_global_variance} and \ref{ass:global_bounded_grad} and $\min_{x}{f(x)} \ge f^*$, let local and global learning rates $\eta_l$, $\eta_g$ be chosen as s.t. $\textstyle{\eta_{g}\eta_{l}\le \frac{1}{3QL\Bar{\alpha}_{t}}}$ and $\textstyle{\eta_{l} < \frac{1}{\sqrt{30}QL}}$, we have,
{
\begin{align}
\label{eq:conv_results}
    \cfrac{1}{T}\sum_{t=0}^{T-1}{\mathbb{E}\left[ \left\lVert \nabla f(x_t) \right\rVert^2\right]}
    \le & \cfrac{1}{c T\eta_{g}\eta_{l}Q|S|} \left(\mathbb{E}\left[\frac{f(x_0)}{\underbar{C}} \right] - \mathbb{E}\left[\frac{f^*}{\bar{C}} \right]\right)
    + \frac{5}{2c} L^2 Q \eta_{l}^2 (\sigma_{l}^2 + 6 Q \sigma_{g}^2) + \cfrac{3 \bar{C}L \eta_{g}\eta_{l} \sigma_l^2}{2c|S|^2|U|} + \cfrac{L \eta_g \sigma^2 C^2 d}{2c\underbar{C}\eta_l Q |S||U|^2} \notag \\
    +& A_{1}\sum_{t=0}^{T-1}{\mathbb{E}\left[\sum_{s \in S}\sum_{u \in U}{\left(\left|\alpha^{s,u}_{t} - \Tilde{\alpha}^{s,u}_{t}\right| + \left|\Tilde{\alpha}^{s,u}_{t} - \Bar{\alpha}_{t}\right|\right)}\right]}
    + A_{2}\sum_{t=0}^{T-1}{\mathbb{E}\left[\sum_{s \in S}\sum_{u \in U}{\left(\left|\alpha^{s,u}_{t} - \Tilde{\alpha}^{s,u}_{t}\right|^2 + \left|\Tilde{\alpha}^{s,u}_{t} - \Bar{\alpha}_{t}\right|^2\right)}\right]}
\end{align}
}

\noindent
where $c>0$, \\
$\bar{C} := \max_{s, u, t}\left(\frac{C}{\max{(C, \eta_{l}\lVert\mathbb{E}\left[\sum_{q\in [Q]}g^{s,u}_{t,q}\right]\rVert)}} \right)$,
$\underline{C} := \min_{s, u, t}\left(\frac{C}{\max{(C, \eta_{l}\lVert\mathbb{E}\left[\sum_{q\in [Q]}g^{s,u}_{t,q}\right]\rVert)}} \right)$,
$\alpha^{s,u}_{t} := \frac{w_{s,u}C}{\max{(C, \eta_{l}\lVert\sum_{q\in [Q]}g^{s,u}_{t,q}\rVert)}}$,
$\Tilde{\alpha}^{s,u}_{t} := \frac{w_{s,u}C}{\max{(C, \eta_{l}\lVert\mathbb{E}\left[\sum_{q\in [Q]}g^{s,u}_{t,q}\right]\rVert)}}$,
$\Bar{\alpha}_{t} := \frac{1}{|S||U|}\sum_{s \in S}\sum_{u \in U}{\Tilde{\alpha}^{s,u}_{t}}$,
$A_{1}:=\frac{G^2}{c\underline{C}|U|T}$
and $A_{2}:=\frac{3L\eta_{g}\eta_{l}QG^2}{2c\underline{C}|U|T}$.
\end{theorem}

\begin{remark}
{\normalfont
    The first three terms recover the standard convergence bound up to constants for FedAVG \cite{yang2021achieving} when considering participants in FL as user-silo pairs (i.e., we have $|S||U|$ participants).
    The asymptotic bound is $O(\frac{1}{\sqrt{|S||U|QT}} + \frac{1}{T})$.
    Theorem \ref{theorem:uldp_conv} achieves this when we choose the global and local learning rates $\eta_{g} = |S|\sqrt{|S||U|Q}$ and $\eta_{l} = \frac{1}{\sqrt{T}QL}$, respectively.
    This requires a learning rate $|S|$ times larger than the usual FedAVG with $|S||U|$ participants, which can be interpreted as coming from the constraint on the weights that total 1 for each user.
    The fourth term is the convergence overhead due to Gaussian noise addition, and the last fifth and sixth terms are the overhead due to bias from the clippings.
    If both the noise and the clipping bias are zero, the convergence rate is asymptotically the same as the FedAVG convergence rate.
}
\end{remark}

\begin{remark}
{\normalfont
    The fourth term, accounting for the overhead due to noise, differs slightly from DP-FedAVG.
    This term is inversely proportional to $|S||U|^2$. 
    As highlighted in the previous remark, if the global learning rate is set as $\eta_{g} = |S|\sqrt{|U|Q}$ in ULDP-AVG, this term becomes proportional to $\sqrt{|U|}/|U|^2$, consistent with the case of DP-FedAVG with $|U|$ participants.
}
\end{remark}

\begin{remark}
\label{remark:grad_var}
{\normalfont
    The fifth and sixth terms correspond to the overhead due to the clipping biases $\left|\alpha^{s,u}_{t} - \Tilde{\alpha}^{s,u}_{t}\right|$ and $\left|\Tilde{\alpha}^{s,u}_{t} - \Bar{\alpha}_{t}\right|$, respectively.
    The quantity $\left|\alpha^{s,u}_{t} - \Tilde{\alpha}^{s,u}_{t}\right|$ represents the local gradient variance across all users in all silos and can be made zero by full-batch gradient descent.
    The noteworthy term is $\left|\Tilde{\alpha}^{s,u}_{t} - \Bar{\alpha}_{t}\right|$.
    As discussed in the analysis of \cite{zhang2022understanding}, this term is influenced by the structure of the neural network and data heterogeneity.
    It roughly quantifies how much the magnitudes of all gradients deviate from the global mean.
    We may be able to minimize these values by selecting appropriate weights $\mathbf{W}$, guided by the following optimization problem:
    {\small
        \begin{align}
            & \min_{\mathbf{W}}{\sum_{s \in S}\sum_{u \in U}\left|\Tilde{\alpha}^{s,u}_{t} - \Bar{\alpha}_{t}\right|}, \; \text{s.t.,} \; w_{s,u} > 0, \forall{u}, \sum_{s\in S}{w_{s,u}}=1  \notag \\
            & \left(= \sum_{s \in S}\sum_{u \in U}\left|w_{s,u} C_{s,u} - \frac{1}{|S||U|}\sum_{s' \in S}\sum_{u' \in U}{w_{s',u'}C_{s',u'}} \right|\right) \notag
        \end{align}
        where $C_{s,u} := \frac{C}{\max{(C, \eta_{l}\lVert\mathbb{E}\left[\sum_{q\in [Q]}g^{s,u}_{t,q}\right]\rVert)}}$.
    }
    
    \noindent
    However, determining the optimal weights is challenging because we cannot predict the gradients' norms in advance, which could also cause another privacy issue. In next section, we present a heuristic weighting strategy that aims to reduce this bias. 
    
}
\end{remark}

\subsection{Proof of Theorem \ref{theorem:uldp_conv}}

The many of techniques in the following proof are seen in \cite{yang2021achieving, zhang2022understanding, 50448}.

For convenience, we define following quantities:
\begin{align}
  \alpha^{s,u}_{t} &:= \cfrac{w_{s,u}C}{\max{(C, \eta_{l}\lVert\sum_{q\in [Q]}g^{s,u}_{t,q}\rVert)}},\;
  \Tilde{\alpha}^{s,u}_{t} := \cfrac{w_{s,u}C}{\max{(C, \eta_{l}\lVert\mathbb{E}\left[\sum_{q\in [Q]}g^{s,u}_{t,q}\right]\rVert)}},\;
  \Bar{\alpha}_{t} := \cfrac{1}{|S||U|}\sum_{s \in S}\sum_{u \in U}{\Tilde{\alpha}^{s,u}_{t}}, \notag \\
  \Delta^{s,u}_{t} &:= - \eta_{l}\sum_{q\in [Q]}{g^{s,u}_{t,q}} \cdot \alpha^{s,u}_{t},\;
  \Tilde{\Delta}^{s,u}_{t} := - \eta_{l}\sum_{q\in [Q]}{g^{s,u}_{t,q}} \cdot \Tilde{\alpha}^{s,u}_{t},\;
  \Bar{\Delta}^{s,u}_{t} := - \eta_{l}\sum_{q\in [Q]}{g^{s,u}_{t,q}} \cdot \Bar{\alpha}_{t}, \\
  \breve{\Delta}^{s,u}_{t} &:= - \eta_{l}\sum_{q\in [Q]}{\nabla f_{s,u}(x^{s,u}_{t,q})} \cdot \Bar{\alpha}_{t},\;
  \Delta^{s}_{t} := \sum_{u \in U}{\Delta^{s,u}_{t}},\;
  \Tilde{\Delta}^{s}_{t} := \sum_{u \in U}{\Tilde{\Delta}^{s,u}_{t}},\;
  \Bar{\Delta}^{s}_{t} := \sum_{u \in U}{\Bar{\Delta}^{s,u}_{t}},\;
  \breve{\Delta}^{s}_{t} := \sum_{u \in U}{\breve{\Delta}^{s,u}_{t}}.
\end{align}
\noindent
where expectation in $\Tilde{\alpha}^{s,u}_{t}$ is taken over all possible randomness.
Due to the smoothness in Assumption \ref{ass:lipschitz}, we have
\begin{align}
\label{eq:lipschitz}
    f(x_{x+1}) \le f(x_{t}) + \langle\nabla f(x_{t}), x_{t+1} - x_{t}\rangle + \frac{L}{2}\lVert x_{t+1} - x_{t}\rVert^2.
\end{align}
The model difference between two consecutive iterations can be represented as 
\begin{align}
\label{eq:diffs}
    x_{x+1} - x_{t} = \eta_g \frac{1}{|S||U|}\sum_{s \in S}{(\Delta^{s}_{t} + z^{s}_{t})}
\end{align}
\noindent
with random noise $z^{s}_{t} \sim \mathcal{N}(0, I\sigma^2 C^2 / |S|)$.
Taking expectation of $f(x_{t+1})$ over the randomness at communication round $t$, we have:
\begin{align}
\label{eq:expectation}
    \mathbb{E}\left[f(x_{x+1})\right] &\le f(x_{t}) + \langle\nabla f(x_{t}), \mathbb{E}\left[x_{t+1} - x_{t}\right]\rangle + \frac{L}{2}\mathbb{E}\left[\lVert x_{t+1} - x_{t}\rVert^2\right] \notag \\
    &= f(x_{t}) + \eta_{g}\langle\nabla f(x_{t}), \mathbb{E}\left[\frac{1}{|S||U|}\sum_{s \in S}{(\Delta^{s}_{t} + z^{s}_{t})}\right]\rangle + \frac{L}{2}\eta_{g}^2\mathbb{E}\left[\left\lVert\frac{1}{|S||U|}\sum_{s \in S}{(\Delta^{s}_{t} + z^{s}_{t})}\right\rVert^2\right] \notag \\
    &= f(x_{t}) + \eta_{g}\langle\nabla f(x_{t}), \mathbb{E}\left[\frac{1}{|S||U|}\sum_{s \in S}{\Delta^{s}_{t}}\right]\rangle + \frac{L}{2}\eta_{g}^2\mathbb{E}\left[\left\lVert\frac{1}{|S||U|}\sum_{s \in S}{\Delta^{s}_{t}}\right\rVert^2\right] + \frac{L}{2}\eta_{g}^2\frac{1}{|S|^2|U|^2}\sigma^2 C^2 d
\end{align}
\noindent
where $d$ in the last expression is dimension of $x_t$ and in the last equation we use the fact that $z^{s}_{t}$ is zero mean normal distribution.

Firstly, we evaluate the first-order term of Eq. (\ref{eq:expectation}),
\begin{align}
\label{eq:first_order1}
    \langle\nabla &f(x_{t}), \mathbb{E}\left[\frac{1}{|S||U|}\sum_{s \in S}{\Delta^{s}_{t}}\right]\rangle \notag \\
    &=\langle\nabla f(x_{t}), \mathbb{E}\left[\frac{1}{|S||U|}\sum_{s \in S}{(\Delta^{s}_{t} - \Tilde{\Delta}^{s}_{t})}\right]\rangle + \langle\nabla f(x_{t}), \mathbb{E}\left[\frac{1}{|S||U|}\sum_{s \in S}{(\Tilde{\Delta}^{s}_{t} - \Bar{\Delta}^{s}_{t})}\right]\rangle + \langle\nabla f(x_{t}), \mathbb{E}\left[\frac{1}{|S||U|}\sum_{s \in S}{\Bar{\Delta}^{s}_{t}}\right]\rangle,
\end{align}
and the last term of Eq. (\ref{eq:first_order1}) can be evaluated as follows:
\begin{align}
\label{eq:first_order2}
    \langle\nabla &f(x_{t}), \mathbb{E}\left[\frac{1}{|S||U|}\sum_{s \in S}{\Bar{\Delta}^{s}_{t}}\right]\rangle \notag \\
    &= - \cfrac{\eta_l\Bar{\alpha}_{t}Q}{2}\lVert\nabla f(x_{t})\rVert^2 - \cfrac{\eta_l\Bar{\alpha}_{t}}{2Q}\mathbb{E}\left[\left\lVert\cfrac{1}{\eta_l|S||U|\Bar{\alpha}_{t}}\sum_{s\in S}{\breve{\Delta}^{s}_{t}}\right\rVert^2\right] + \cfrac{\eta_l\Bar{\alpha}_{t}}{2}\mathbb{E}\left[\left\lVert\sqrt{Q}\nabla f(x_{t}) + \cfrac{1}{\sqrt{Q}\eta_l|S||U|\Bar{\alpha}_{t}}\sum_{s\in S}{\breve{\Delta}^{s}_{t}}\right\rVert^2\right].
\end{align}
This is because $\mathbb{E}\Bar{\Delta}^{s}_{t} = \breve{\Delta}^{s}_{t}$ and $\langle a, b \rangle = - \frac{1}{2}\lVert a \rVert^2 - \frac{1}{2}\lVert b \rVert^2 + \frac{1}{2}\lVert a+b \rVert^2$ for any vector $a$, $b$.

We further upper bound the last term of Eq. \ref{eq:first_order2} as:
\begin{align}
\label{eq:first_order3}
    \mathbb{E}\left[\left\lVert\sqrt{Q}\nabla f(x_{t}) + \cfrac{1}{\sqrt{Q}\eta_l|S||U|\Bar{\alpha}_{t}}\sum_{s\in S}{\breve{\Delta}^{s}_{t}}\right\rVert^2\right] &= Q\mathbb{E}\left[\left\lVert\nabla f(x_{t}) + \cfrac{1}{Q\eta_l|S||U|\Bar{\alpha}_{t}}\sum_{s\in S}{(\sum_{u\in U}{(- \eta_{l}\sum_{q\in [Q]}{\nabla f_{s,u}(x^{s,u}_{t,q})} \cdot \Bar{\alpha}_{t}}))}\right\rVert^2\right] \notag \\
    &= Q\mathbb{E}\left[\left\lVert \cfrac{1}{Q|S||U|}\sum_{s\in S}\sum_{u\in U}\sum_{q \in [Q]}{\nabla f_{s,u}(x_{t}) - \nabla f_{s,u}(x^{s,u}_{t,q})}\right\rVert^2\right] \notag \\
    &\stackrel{(a1)}{\le} Q \cdot \cfrac{1}{Q|S||U|} \mathbb{E}\left[\sum_{s\in S}\sum_{u\in U}\sum_{q \in [Q]} {\lVert \nabla f_{s,u}(x_{t}) - \nabla f_{s,u}(x^{s,u}_{t,q})\rVert^2}\right] \notag \\
    &\stackrel{(a2)}{\le} \cfrac{1}{|S||U|} \sum_{s\in S}\sum_{u\in U}\sum_{q \in [Q]}{L^2 \mathbb{E}\left[{\lVert x_{t} - x^{s,u}_{t,q}  \rVert^2}\right]} \notag \\
    &\stackrel{(a3)}{\le} L^2 5 Q^2 \eta_{l}^2 (\sigma_{l}^2 + 6 Q \sigma_{g}^2) + L^2 30 Q^3 \eta_{l}^2 \lVert \nabla f(x_t)\rVert^2,
\end{align}
where we use $\mathbb{E}\left[\lVert X_1 + ... + X_n \rVert^2\right] \le n\mathbb{E}\left[\lVert X_1 \rVert^2 + ... + \lVert X_n \rVert^2 \right]$ at (a1), L-smoothness (Assumption \ref{ass:lipschitz}) at (a2), and Lemma 3 of \cite{50448} at (a3).

Secondly, we evaluate the second-order term in Eq. (\ref{eq:expectation}) as follows:
\begin{align}
\label{eq:second_order1}
    \mathbb{E}&\left[\left\lVert\frac{1}{|S||U|}\sum_{s \in S}{\Delta^{s}_{t}}\right\rVert^2\right] \notag \\
    &\le 3\mathbb{E}\left[\left\lVert\frac{1}{|S||U|}\sum_{s \in S}{\Delta^{s}_{t} - \Tilde{\Delta}^{s}_{t}}\right\rVert^2\right] + 3\mathbb{E}\left[\left\lVert\frac{1}{|S||U|}\sum_{s \in S}{\Delta^{s}_{t} - \Bar{\Delta}^{s}_{t}}\right\rVert^2\right] + 3\mathbb{E}\left[\left\lVert\frac{1}{|S||U|}\sum_{s \in S}{\Bar{\Delta}^{s}_{t}}\right\rVert^2\right].
\end{align}
This is because $(a+b+c)^2 \le 3a^2 + 3b^2 + 3c^2$ holds when $a=A-B$, $b=B-C$, $c=C$ for all vector $A, B, C$.
\noindent
We can bound the expectation in the last term of Eq. (\ref{eq:second_order1}) as follows:
\begin{align}
\label{eq:second_order2}
    \mathbb{E}\left[\left\lVert\frac{1}{|S||U|}\sum_{s \in S}{\Bar{\Delta}^{s}_{t}}\right\rVert^2\right] &= \mathbb{E}\left[\left\lVert\frac{1}{|S||U|}\sum_{s \in S}{\sum_{u\in U}{(\eta_l \sum_{q \in [Q]}{g^{s,u}_{t,q}} \cdot \Bar{\alpha}_{t})}}\right\rVert^2\right] \notag \\
    &\stackrel{(a4)}{=} \eta_l^2 \mathbb{E}\left[\left\lVert\frac{1}{|S||U|}\sum_{s \in S}{\sum_{u\in U}{\sum_{q \in [Q]}{\nabla f_{s,u}(x^{s,u}_{t,q}) \cdot \Bar{\alpha}_{t}}}}\right\rVert^2 + \left\lVert\frac{1}{|S||U|}\sum_{s \in S}{\sum_{u\in U}{\sum_{q \in [Q]}{(\nabla f_{s,u}(x^{s,u}_{t,q}) - g^{s,u}_{t,q}) \cdot \Bar{\alpha}_{t}}}}\right\rVert^2\right] \notag \\
    &= \mathbb{E}\left[\left\lVert\frac{1}{|S||U|}\sum_{s \in S}{\breve{\Delta}^{s}_{t}}\right\rVert^2\right] + \cfrac{\eta_l^2 \Bar{\alpha}_{t}^2}{|S|^2|U|^2}\mathbb{E}\left[\left\lVert \sum_{s \in S}{\sum_{u\in U}{\sum_{q \in [Q]}{(\nabla f_{s,u}(x^{s,u}_{t,q}) - g^{s,u}_{t,q})}}}\right\rVert^2\right] \notag \\
    &\stackrel{(a5)}{\le} \mathbb{E}\left[\left\lVert\frac{1}{|S||U|}\sum_{s \in S}{\breve{\Delta}^{s}_{t}}\right\rVert^2\right] + \cfrac{\eta_l^2 \Bar{\alpha}_{t}^2}{|S|^2|U|^2}\mathbb{E}\left[\sum_{s \in S}{\sum_{u\in U}{\sum_{q \in [Q]}{\left\lVert\nabla f_{s,u}(x^{s,u}_{t,q}) - g^{s,u}_{t,q}\right\rVert^2}}}\right] \notag \\
    &\le \mathbb{E}\left[\left\lVert\frac{1}{|S||U|}\sum_{s \in S}{\breve{\Delta}^{s}_{t}}\right\rVert^2\right] + \cfrac{\eta_l^2 \Bar{\alpha}_{t}^2}{|S|^2|U|^2}\sum_{s \in S}{\sum_{u\in U}{\sum_{q \in [Q]}{\sigma_l^2}}} \notag \\
    &\le \mathbb{E}\left[\left\lVert\frac{1}{|S||U|}\sum_{s \in S}{\breve{\Delta}^{s}_{t}}\right\rVert^2\right] + \cfrac{\eta_l^2 \Bar{\alpha}_{t}^2}{|S||U|}Q\sigma_l^2,
\end{align}
where we use $\mathbb{E}\left[\lVert X \rVert^2\right] = \mathbb{E}\left[\lVert X - \mathbb{E}\left[ X \right]\rVert^2 \right] + \lVert\mathbb{E}\left[ X \right]\rVert^2$ at (a4), and $\mathbb{E}\left[\lVert X_1 + ... + X_n \rVert^2\right] \le \mathbb{E}\left[\lVert X_1 \rVert^2 + ... + \lVert X_n \rVert^2 \right]$ when $\forall i,j, i \not= j$,  $X_i$ and $X_j$ are independent and $\mathbb{E}\left[X_i\right]=0$ and Assumption \ref{ass:local_and_global_variance} at (a5).

Lastly, combining all of above, we have
\begin{align}
\label{eq:last}
    \mathbb{E}\left[f(x_{x+1})\right] &\le f(x_{t}) + \eta_{g}\langle\nabla f(x_{t}), \mathbb{E}\left[\frac{1}{|S||U|}\sum_{s \in S}{(\Delta^{s}_{t} - \Tilde{\Delta}^{s}_{t})}\right]\rangle + \eta_{g}\langle\nabla f(x_{t}), \mathbb{E}\left[\frac{1}{|S||U|}\sum_{s \in S}{(\Tilde{\Delta}^{s}_{t} - \Bar{\Delta}^{s}_{t})}\right]\rangle \notag \\
    &\;\;\;\; - \frac{1}{2}\eta_{g}\eta_{l}\Bar{\alpha}_{t}Q \lVert\nabla f(x_{t}) \rVert^2 - \frac{1}{2}\cfrac{\eta_g}{Q|S|^2|U|^2\eta_l \Bar{\alpha}_{t}}\mathbb{E}\left[\left\lVert\sum_{s \in S}{\breve{\Delta}^{s}_{t}}\right\rVert^2\right] \notag \\
    &\;\;\;\; + \cfrac{\eta_g\eta_l\Bar{\alpha}_{t}}{2}\left(L^2 5 Q^2 \eta_{l}^2 (\sigma_{l}^2 + 6 Q \sigma_{g}^2) + L^2 30 Q^3 \eta_{l}^2 \lVert \nabla f(x_t)\rVert^2\right) \notag \\
    &\;\;\;\; + \frac{3}{2}L\eta_{g}^2 \mathbb{E}\left[\left\lVert\frac{1}{|S||U|}\sum_{s \in S}{\Delta^{s}_{t} - \Tilde{\Delta}^{s}_{t}}\right\rVert^2\right] + \frac{3}{2}L\eta_{g}^2 \mathbb{E}\left[\left\lVert\frac{1}{|S||U|}\sum_{s \in S}{\Tilde{\Delta}^{s}_{t} - \Bar{\Delta}^{s}_{t}}\right\rVert^2\right] + \frac{3}{2}L\eta_{g}^2 \frac{1}{|S|^2|U|^2}\mathbb{E}\left[\left\lVert\sum_{s \in S}{\breve{\Delta}^{s}_{t}}\right\rVert^2\right] \notag \\
    &\;\;\;\; + \frac{3}{2}\cfrac{\eta_g^2 \eta_l^2 \Bar{\alpha}_{t}^2 LQ\sigma_l^2}{|S||U|} + \frac{1}{2}\cfrac{L \eta_g^2 \sigma^2 C^2 d}{|S|^2|U|^2} \notag \\
    &= f(x_{t}) - \left( \frac{1}{2}\eta_{l}\eta_{g}\Bar{\alpha}_{t}Q - \cfrac{\eta_g\eta_l\Bar{\alpha}_{t}}{2}L^2 30 Q^3 \eta_{l}^2 \right) \lVert \nabla f(x_t)\rVert^2 \notag \\
    &\;\;\;\; - \left( \frac{1}{2}\cfrac{\eta_g}{Q|S|^2|U|^2\eta_l \Bar{\alpha}_{t}} - \frac{3}{2}L\eta_{g}^2 \frac{1}{|S|^2|U|^2} \right) \mathbb{E}\left[\left\lVert\sum_{s \in S}{\breve{\Delta}^{s}_{t}}\right\rVert^2\right] \notag \\
    &\;\;\;\; + \cfrac{\eta_g\eta_l\Bar{\alpha}_{t}}{2}L^2 5 Q^2 \eta_{l}^2 (\sigma_{l}^2 + 6 Q \sigma_{g}^2) + \frac{3}{2}\cfrac{\eta_g^2 \eta_l^2 \Bar{\alpha}_{t}^2 LQ\sigma_l^2}{|S||U|} + \frac{1}{2}\cfrac{L \eta_g^2 \sigma^2 C^2 d}{|S|^2|U|^2} \notag \\
    &\;\;\;\; + \underbrace{\eta_{g}\langle\nabla f(x_{t}), \mathbb{E}\left[\frac{1}{|S||U|}\sum_{s \in S}{(\Delta^{s}_{t} - \Tilde{\Delta}^{s}_{t})}\right]\rangle + \eta_{g}\langle\nabla f(x_{t}), \mathbb{E}\left[\frac{1}{|S||U|}\sum_{s \in S}{(\Tilde{\Delta}^{s}_{t} - \Bar{\Delta}^{s}_{t})}\right]\rangle}_{A_1} \notag \\
    &\;\;\;\; + \underbrace{\frac{3}{2}L\eta_{g}^2 \mathbb{E}\left[\left\lVert\frac{1}{|S||U|}\sum_{s \in S}{\Delta^{s}_{t} - \Tilde{\Delta}^{s}_{t}}\right\rVert^2\right] + \frac{3}{2}L\eta_{g}^2 \mathbb{E}\left[\left\lVert\frac{1}{|S||U|}\sum_{s \in S}{\Tilde{\Delta}^{s}_{t} - \Bar{\Delta}^{s}_{t}}\right\rVert^2\right]}_{A_2} \notag \\
    &\stackrel{(a6)}{\le} f(x_{t}) - \eta_{l}\eta_{g}\Bar{\alpha}_{t}Q \left( \frac{1}{2} - 15 L^2 Q^2 \eta_{l}^2 \right) \lVert \nabla f(x_t)\rVert^2 \notag \\
    &\;\;\;\; + \cfrac{\eta_g\eta_l\Bar{\alpha}_{t}}{2}L^2 5 Q^2 \eta_{l}^2 (\sigma_{l}^2 + 6 Q \sigma_{g}^2) + \frac{3}{2}\cfrac{\eta_g^2 \eta_l^2 \Bar{\alpha}_{t}^2 LQ\sigma_l^2}{|S||U|} + \frac{1}{2}\cfrac{L \eta_g^2 \sigma^2 C^2 d}{|S|^2|U|^2} + A_1 + A_2 \notag \\
    &\stackrel{(a7)}{\le} f(x_{t}) - c\eta_{l}\eta_{g}\Bar{\alpha}_{t}Q \lVert \nabla f(x_t)\rVert^2 + \cfrac{\eta_g\eta_l\Bar{\alpha}_{t}}{2}L^2 5 Q^2 \eta_{l}^2 (\sigma_{l}^2 + 6 Q \sigma_{g}^2) + \frac{3}{2}\cfrac{\eta_g^2 \eta_l^2 \Bar{\alpha}_{t}^2 LQ\sigma_l^2}{|S||U|} + \frac{1}{2}\cfrac{L \eta_g^2 \sigma^2 C^2 d}{|S|^2|U|^2} + A_1 + A_2
\end{align}
where (a6) follows from $\left( \frac{1}{2}\frac{\eta_g}{Q|S|^2|U|^2\eta_l \Bar{\alpha}_{t}} - \frac{3}{2}L\eta_{g}^2 \frac{1}{|S|^2|U|^2} \right) \ge 0$ if $\eta_{g}\eta_{l}\le \frac{1}{3QL\Bar{\alpha}_{t}}$ and replacing the last terms with $A_1$ and $A_2$, and (a7) holds because there exists a constant $c > 0$ satisfying $\left( \frac{1}{2} - 15 L^2 Q^2 \eta_{l}^2 \right) > c > 0$ if $\eta_{l} < \frac{1}{\sqrt{30}QL}$.
Divide both sides of (\ref{eq:last}) by $c\eta_{l}\eta_{g}Q\Bar{\alpha}_{t}$, sum over $t$ from $0$ to $T-1$, divide both sides by $T$, and rearrange, we have
\begin{align}
\label{eq:last2}
    \cfrac{1}{T}\sum_{t=0}^{T-1}{\mathbb{E}\left[ \left\lVert \nabla f(x_t) \right\rVert^2\right]}
    \le \; & \cfrac{1}{c T\eta_{g}\eta_{l}Q} \left(\mathbb{E}\left[\frac{f(x_0)}{\Bar{\alpha}_0} \right] - \mathbb{E}\left[\frac{f(x_T)}{\Bar{\alpha}_{T-1}} \right]\right) \notag \\
    &\;\;\; + \frac{5}{2c} L^2 Q \eta_{l}^2 (\sigma_{l}^2 + 6 Q \sigma_{g}^2) + \cfrac{3 L \eta_{g}\eta_{l} \sigma_l^2}{2c|S||U|} \cfrac{1}{T}\sum_{t=0}^{T-1}{\Bar{\alpha}_t} + \cfrac{L \eta_g \sigma^2 C^2 d}{2c\eta_l Q |S|^2|U|^2} \cfrac{1}{T}\sum_{t=0}^{T-1}{\frac{1}{\Bar{\alpha}_t}} \notag \\
    &\;\;\; + \underbrace{\cfrac{1}{c\eta_{l}Q} \cfrac{1}{T}\sum_{t=0}^{T-1}{\frac{1}{\Bar{\alpha}_t}}{\mathbb{E}\left[\langle\nabla f(x_{t}), \mathbb{E}\left[\frac{1}{|S||U|}\sum_{s \in S}{(\Delta^{s}_{t} - \Tilde{\Delta}^{s}_{t})}\right]\rangle + \langle\nabla f(x_{t}), \mathbb{E}\left[\frac{1}{|S||U|}\sum_{s \in S}{(\Tilde{\Delta}^{s}_{t} - \Bar{\Delta}^{s}_{t})}\right]\rangle\right]}}_{B_1} \notag \\
    &\;\;\; + \underbrace{\cfrac{3L\eta_{g}}{2c\eta_{l}Q} \cfrac{1}{T}\sum_{t=0}^{T-1}{\frac{1}{\Bar{\alpha}_t}} {\left(\mathbb{E}\left[\left\lVert\frac{1}{|S||U|}\sum_{s \in S}{\Delta^{s}_{t} - \Tilde{\Delta}^{s}_{t}}\right\rVert^2\right] + \mathbb{E}\left[\left\lVert\frac{1}{|S||U|}\sum_{s \in S}{\Tilde{\Delta}^{s}_{t} - \Bar{\Delta}^{s}_{t}}\right\rVert^2\right]\right)}}_{B_2}.
\end{align}
Let $\smash{\displaystyle\max_{s \in S, u \in U, t \in [T]}}{\left(\frac{C}{\max{(C, \eta_{l}\lVert\mathbb{E}\left[\sum_{q\in [Q]}g^{s,u}_{t,q}\right]\rVert)}} \right)}$ be $\bar{C}$, and $\smash{\displaystyle\min_{s \in S, u \in U, t \in [T]}}{\left(\frac{C}{\max{(C, \eta_{l}\lVert\mathbb{E}\left[\sum_{q\in [Q]}g^{s,u}_{t,q}\right]\rVert)}} \right)}$ be $\underbar{C}$, $\frac{1}{|S|}\underbar{C} \le \Bar{\alpha}_t \le \frac{1}{|S|}\bar{C}$ since $\Bar{\alpha}_t$'s definition and $\frac{1}{|S|}\sum_{s\in S}w_{s,u} = \frac{1}{|S|}$.
Using this, (\ref{eq:last2}) is evaluated as follows:
\begin{align}
\label{eq:last3}
    \cfrac{1}{T}\sum_{t=0}^{T-1}{\mathbb{E}\left[ \left\lVert \nabla f(x_t) \right\rVert^2\right]}
    \le \; & \cfrac{1}{c T\eta_{g}\eta_{l}Q|S|} \left(\mathbb{E}\left[\frac{f(x_0)}{\underbar{C}} \right] - \mathbb{E}\left[\frac{f(x_T)}{\bar{C}} \right]\right) \notag \\
    &\;\;\; + \frac{5}{2c} L^2 Q \eta_{l}^2 (\sigma_{l}^2 + 6 Q \sigma_{g}^2) + \cfrac{3 \bar{C}L \eta_{g}\eta_{l} \sigma_l^2}{2c|S|^2|U|} + \cfrac{L \eta_g \sigma^2 C^2 d}{2c\underbar{C}\eta_l Q |S||U|^2} \notag \\
    &\;\;\; + \underbrace{\cfrac{|S|}{c\underbar{C}\eta_{l}Q} \cfrac{1}{T}\sum_{t=0}^{T-1}{\mathbb{E}\left[\langle\nabla f(x_{t}), \mathbb{E}\left[\frac{1}{|S||U|}\sum_{s \in S}{(\Delta^{s}_{t} - \Tilde{\Delta}^{s}_{t})}\right]\rangle + \langle\nabla f(x_{t}), \mathbb{E}\left[\frac{1}{|S||U|}\sum_{s \in S}{(\Tilde{\Delta}^{s}_{t} - \Bar{\Delta}^{s}_{t})}\right]\rangle\right]}}_{B_1} \notag \\
    &\;\;\; + \underbrace{\cfrac{3L\eta_{g}|S|}{2c\underbar{C}\eta_{l}Q} \cfrac{1}{T}\sum_{t=0}^{T-1}{\left(\mathbb{E}\left[\left\lVert\frac{1}{|S||U|}\sum_{s \in S}{\Delta^{s}_{t} - \Tilde{\Delta}^{s}_{t}}\right\rVert^2\right] + \mathbb{E}\left[\left\lVert\frac{1}{|S||U|}\sum_{s \in S}{\Tilde{\Delta}^{s}_{t} - \Bar{\Delta}^{s}_{t}}\right\rVert^2\right]\right)}}_{B_2}.
\end{align}

\noindent
$B_1$ and $B_2$ is upper-bounded as follows:
\begin{align}
\label{eq:last_B1}
    B_1 &\le \cfrac{|S|}{c\underbar{C}\eta_{l}Q}\cfrac{1}{T}\sum_{t=0}^{T-1}{\mathbb{E}\left[\langle\nabla f(x_{t}), \mathbb{E}\left[\frac{1}{|S||U|}\sum_{s \in S}\sum_{u \in U}{\left|\Delta^{s,u}_{t} - \Tilde{\Delta}^{s,u}_{t}\right|}\right]\rangle + \langle\nabla f(x_{t}), \mathbb{E}\left[\frac{1}{|S||U|}\sum_{s \in S}\sum_{u \in U}{\left|\Tilde{\Delta}^{s,u}_{t} - \Bar{\Delta}^{s,u}_{t}\right|}\right]\rangle\right]} \notag \\
    &\le \cfrac{G^2}{c\underbar{C}|U|}\cfrac{1}{T}\sum_{t=0}^{T-1}{\mathbb{E}\left[\sum_{s \in S}\sum_{u \in U}{\left(\left|\alpha^{s,u}_{t} - \Tilde{\alpha}^{s,u}_{t}\right| + \left|\Tilde{\alpha}^{s,u}_{t} - \Bar{\alpha}_{t}\right|\right)}\right]}, \notag \\
    B_2 &\stackrel{(a8)}{\le} \cfrac{3L\eta_{g}|S|}{2c\underbar{C}\eta_{l}Q}\cfrac{1}{T}\sum_{t=0}^{T-1}{\left(\mathbb{E}\left[\frac{1}{|S||U|}\sum_{s \in S}\sum_{u \in U}\left|{\Delta^{s,u}_{t} - \Tilde{\Delta}^{s,u}_{t}}\right|^2\right] + \mathbb{E}\left[\frac{1}{|S||U|}\sum_{s \in S}\sum_{u \in U}\left|{\Tilde{\Delta}^{s,u}_{t} - \Bar{\Delta}^{s,u}_{t}}\right|^2\right]\right)} \notag \\
    &\le \cfrac{3L\eta_{g}\eta_{l}QG^2}{2c\underbar{C}|U|}\cfrac{1}{T}\sum_{t=0}^{T-1}{\mathbb{E}\left[\sum_{s \in S}\sum_{u \in U}{\left(\left|\alpha^{s,u}_{t} - \Tilde{\alpha}^{s,u}_{t}\right|^2 + \left|\Tilde{\alpha}^{s,u}_{t} - \Bar{\alpha}_{t}\right|^2\right)}\right]}
\end{align}
with
\begin{align}
\label{eq:last_B2}
    \left|\Delta^{s,u}_{t} - \Tilde{\Delta}^{s,u}_{t}\right| &\le \eta_{l}QG\left|\alpha^{s,u}_{t} - \Tilde{\alpha}^{s,u}_{t}\right|, \notag \\
    \left|\Tilde{\Delta}^{s,u}_{t} - \Bar{\Delta}^{s,u}_{t}\right| &\le \eta_{l}QG\left|\Tilde{\alpha}^{s,u}_{t} - \Bar{\alpha}_{t}\right|,
\end{align}
by Assumption \ref{ass:global_bounded_grad} of globally bounded gradients, and we use the same technique at (a8) as (a1).

\end{document}